\documentclass{article} 
\usepackage{iclr2025_conference,times}

\usepackage{amsmath,amsfonts,bm}

\def\eqref#1{equation~\ref{#1}}

\def\1{\bm{1}}

\def\ra{{\textnormal{a}}}

\DeclareMathAlphabet{\mathsfit}{\encodingdefault}{\sfdefault}{m}{sl}
\SetMathAlphabet{\mathsfit}{bold}{\encodingdefault}{\sfdefault}{bx}{n}

\usepackage{algorithm}
\usepackage{algorithmic}

\usepackage{hyperref}
\usepackage{placeins}
\usepackage{enumitem}
\usepackage{subcaption}
\usepackage{url}
\usepackage[utf8]{inputenc} 
\usepackage[T1]{fontenc}    
\usepackage{hyperref}       
\usepackage{url}            
\usepackage{booktabs}       
\usepackage{amsfonts, amsmath, amssymb}       
\usepackage{placeins} 
\usepackage{nicefrac}       
\usepackage{microtype}      
\usepackage[dvipsnames]{xcolor}        
\usepackage{graphicx}
\usepackage{float}

\usepackage{%
  ./etc/paper_header}
\usepackage{%
  ./etc/kprema_newcommands}

\newcommand{\mr}{\mathrm}
\newcommand{\QCM}[2]{\ul{M}_{\ul{#2}}(\mc{#1})}
\newlength{\dhatheight}

\usepackage{makecell}
\def\QEDclosed{%
  \hfill\ensuremath{\mbox{\rule[0pt]{1.3ex}{1.3ex}}}}

\title{Semantic Uncertainty Quantification of Hallucinations in LLMs: A Quantum Tensor Network Based Method}

\author{
Pragatheeswaran Vipulanandan$^{1}$, Kamal Premaratne$^{1}$, Dilip Sarkar$^{2}$ \\
$^{1}$Department of Electrical and Computer Engineering \\
$^{2}$Department of Computer Science \\
University of Miami \\
Coral Gables, FL 33146, USA \\
\texttt{pxv245@miami.edu, kamal@miami.edu, sarkar@miami.edu}
}

\iclrfinalcopy 
\begin{document}

\maketitle


\begin{abstract}

Large language models (LLMs) exhibit strong generative capabilities but remain vulnerable to confabulations, fluent yet unreliable outputs that vary arbitrarily even under identical prompts. Leveraging a quantum tensor network–based pipeline, we propose a quantum physics-inspired uncertainty quantification framework that accounts for the aleatoric uncertainty in token sequence probability for semantic equivalence-based clustering of LLM generations. In turn, this offers a principled and interpretable scheme for hallucination detection. We further introduce an entropy-maximization strategy that prioritizes high-certainty, semantically coherent outputs and highlights entropy regions where LLM decisions are likely to be unreliable, offering practical guidelines for when human oversight is warranted. We evaluate the robustness of our scheme under different generation lengths and quantization levels, dimensions overlooked in prior studies, demonstrating that our approach remains reliable even in resource-constrained deployments. A total of 116 experiments on TriviaQA, NQ, SVAMP, and SQuAD across multiple architectures (Mistral-7B, Mistral-7B-instruct, Falcon-rw-1b, LLaMA-3.2-1b, LLaMA-2-13b-chat, LLaMA-2-7b-chat, LLaMA-2-13b and LLaMA-2-7b) show consistent improvements in AUROC and AURAC over state-of-the-art baselines. 
\end{abstract}


\section{Introduction}


Large language models (LLMs) have demonstrated remarkable capabilities in understanding, interpreting, and reasoning over human language and natural language generation (NLG) at a high level of fluency and coherence. LLMs, trained on massive text corpora, have led to the creation of powerful tools that brace state-of-the-art (SOTA) approaches in tasks such as summarization, question answering (Q\&A), dialogue generation, and are reshaping how humans interact with machines (e.g., ChatGPT, which provides conversational AI experiences (\cite{hosseini2023exploratory}), GitHub Copilot (\cite{chen2021evaluating}), DALL·E (\cite{lai2023mini}), and others).  

As LLMs are being integrated in new applications in nearly all aspects of daily life and in almost every scientific field, the reliability of their outputs has gained increasing attention. Errors and inconsistencies they introduce can have a direct and significant impact on downstream performance. This issue is particularly evident in interactive systems such as chatbots, where the responses are expected to be not only fluent but also factually grounded (\cite{lei2023chain}), especially when they are being relied upon in more sensitive application scenarios (e.g., healthcare, defense, law, journalism, autonomous driving) (\cite{kaddour2023challenges, zhao2023survey, lei2023chain}). One particular vulnerability of LLMs is \emph{hallucinations,} which refer to LLM generations that appear coherent and plausible but are in fact incorrect, unverifiable, or entirely fabricated (\cite{ji2023survey}). Hallucinated outputs can range from minor factual inaccuracies to completely spurious claims, and they pose a serious obstacle to the deployment of LLMs in high-stakes domains; use of hallucinatory content has significant ethical concerns as well.  


\subsection{Previous Work}


Most approaches for detecting and mitigating hallucinations lean heavily on supervised learning frameworks  (\cite{rateike2023weakly, quevedo2024detecting}) or in-context learning techniques grounded in complex syntactic/semantic analyses (\cite{wang2022self, manakul2023selfcheckgpt, zhang2023siren}). While showing promise, their computational demands often introduce latency making them unfit for real-time applications. Numerical features (e.g., statistical patterns and structural output properties) and correlations between them may also serve as effective indicators of hallucinations (\cite{azaria2023internal, lee2023mathematical, su2024unsupervised}), yielding lightweight, resource-efficient detection methods that complement or even outperform more computationally intensive techniques. Although Bayesian deep learning (BDL) methods have long been used to quantify uncertainty in standard classification tasks (\cite{schweighofer2024information}), they are far less effective for hallucination detection in LLMs: they are challenging to calibrate, they do not reliably reflect a model’s uncertainty about factual correctness, and next-token probabilities do not behave like true class-prediction probabilities. As a result, classical BDL tools struggle to capture the semantic and contextual instabilities that drive hallucinations, limiting their usefulness in this setting (\cite{kang2025uncertainty}).

Other works focus on specific types of hallucinations with particular attention paid to \emph{confabulations,} fluent but incorrect generations that appear arbitrary (\cite{berrios1998confabulations}). They are sensitive to irrelevant factors (e.g., random seeds and prompt phrasing) and are distinct from other types of hallucinations, such as, incorrect answers from models that have been trained on widespread misconceptions (\cite{lin2022teaching}), ``lies'' strategically generated in pursuit of reward signals in reinforcement learning setups (\cite{evans2021truthful}), and systematic failures triggered from flawed reasoning or generalization. 
To detect confabulations, \cite{quevedo2024detecting} employ token sequence (TS) probabilities and their induced entropy, a natural and interpretable measure of uncertainty (\cite{lindley1956measure, kadavath2022language}), but naive estimates of entropy may not serve as a good indicator of hallucinations (\cite{xiao2020wat}) because lexical/syntactic differences in LLM generations often do not imply a semantic difference. Therefore, \cite{farquhar2024detecting} and \cite{kossen2024semantic} employ \emph{semantic entropy (SE),} not lexical/syntactic variations, of generated outputs to quantify hallucinatory behavior. The output (TSs) generated when an LLM confronts the same input (context + prompt) repeatedly are clustered based on their bidirectional entailment (as assessed by DeBERTa). Higher semantic entropy indicates higher likelihood of hallucination. In a similar vein, Kernelized Likelihood Entropy (KLE) (\cite{nikitin2024kernel}) employs a graded similarity measure to account for semantic differences; Semantic Nearest Neighbour Entropy (SNNE) (\cite{nguyen2025beyond}) aggregates pairwise similarities with LogSumExp smoothing, avoiding explicit clustering and improving robustness on summarization and translation; Semantic Density (SD) (\cite{qiu2024semantic}) weights semantic distances by generation probabilities. These approaches improve confabulation detection, but are sensitive to clustering quality, similarity functions, and TS probability fluctuations.

Structure-aware and perturbation-based methods have also been explored for detecting hallucinations. Graph Uncertainty (\cite{jiang2024graph}) and KEA (Kernel-Enriched AI) explain  (\cite{haskins2025kea}) model relationships between knowledge graphs (KGs) constructed from LLM outputs and ground truth claims for detection of hallucinations. But they incur heavy computational overhead and rely on an external knowledge base and accurate KG construction. Sampling with Perturbation for Uncertainty Quantification (SPUQ) (\cite{gao2024spuq}) and Semantic-Invariant Perturbation Sampling (\cite{cox2025mapping}) estimate uncertainty by probing model sensitivity to paraphrased or perturbed inputs, exposing prompt sensitivity but at the cost of repeated queries.

In spite of the advances they have made, SOTA methods still struggle with reliably detecting fabricated content across diverse tasks and domains (\cite{li2023halueval}). A possible reason is that they leave unaddressed the aleatoric uncertainty of the TS probabilities. Hallucination risk should depend on how sensitive these TS probabilities, and hence the semantic entropies they induce, are to model perturbations. This highlights the need for \emph{local,} rather than \emph{global,} sensitivity metrics to assess TS probability  uncertainty. Prior studies have also not examined the robustness of their approaches with different quantization levels despite the latter playing a major role in real-world deployment of LLMs. 


\subsection{Our Contributions}
\label{contributions}


``White box'' inherently interpretable approaches (e.g., random sampling-based variational inference methods (\cite{Ghahramani2015N_ProbabilisticML})) can be computationally prohibitive for UQ of massively scaled LLMs and physics-inspired methods have emerged as a compelling alternative. Our work employs the physics-inspired approach, advocated in \cite{Principe2010_Book}, and adopted in \cite{Singh2020UAI_TimeSeriesAnalysis} and \cite{Vipulanandan2024}, that leverages perturbation theory of an appropriate quantum wave function to offer a deterministic, one-shot, interpretable (\cite{Rudin2019NatureMI_StopExplainingBlackBoxMLModels, Linardatos2021Entropy_ExplainableAI}) method for quantifying uncertainty locally and at different resolutions. 

\begin{figure}[htbp]
  \centering
  \includegraphics[width = 0.80\textwidth]{%
    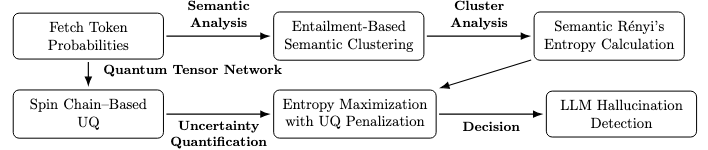}
    \caption{Overview of our hallucination detection pipeline. Sequences are clustered via directional entailment, and UQ obtained through QTN is used for entropy maximization to enable reliable LLM hallucination detection.}
  \color{black}
  \label{fig:semantic_pipeline}
\end{figure}

The main contributions and key advantages of our approach are as follows: 
    \tb{(a)}~We introduce \emph{TS probability uncertainty} as a novel indicator of confabulation detection. For this purpose, we view the TS probabilities as the wave function of a quantum tensor network (QTN) (\cite{ Qi2019Quantum_DeterminingLocalHamiltonian, Vipulanandan2024, vipulanandan2026sensitivity}) and leverage perturbation theory for UQ. 
    \tb{(b)}~We offer a method, based on entropy maximization, to calibrate the TS probabilities to account for this uncertainty associated with TS probabilities. 
    \tb{(c)}~This leads us to cluster LLM generations based on not only SE but uncertainty of TS probabilities. So, instead of simply flagging high entropy generations as hallucinations (\cite{farquhar2024detecting}), we offer a more meaningful scheme that flags outputs associated with higher uncertainty so that one may reduce the likelihood of false negatives and false positives. 
    \tb{(d)}~We evaluate the robustness of our framework under different LLM quantization levels as well as under different generation lengths, ranging from short phrase answers to sentence-level outputs. Our results show that the method is robust and remains reliable even in resource-constrained or efficiency-optimized deployments, thus extending the utility of hallucination detection to real-world settings. 
The overall flow of our approach is illustrated in Fig.~\ref{fig:semantic_pipeline}.



\section{Proposed Method for UQ of LLM Generations}
\label{sec:proposedMethod}


\subsection{Preliminaries: Semantic Clustering}
\label{sec:Preliminaries}


Consider the output TS $\ul{s}$ that the LLM generates in response to input $\ul{y}$. Let $s_i$ denote the $i$-th token and $\ul{s}_{<i}$ the sequence of previous tokens. Then, the probability of $\ul{s}$, conditioned on $\ul{y}$, is 
\begin{equation}
  P(\ul{s} \mid \ul{y})
    = \prod_i
      P(s_i \mid \ul{s}_{<i}, \ul{y}).
  \label{eq:1}
\end{equation}
Clustering is determined by the semantic dissonance between output TSs as determined by bidirectional entailment (for which we employ DeBERTa). Let $C$ denote the set of semantically dissonant classes and $|C|$ its cardinality. Treating cluster label $c_j$ as a random variable (on the space of $C$), the \emph{cluster probability} that the TS $\ul{s}$ belongs in $c_j \in C$ and the induced \emph{semantic (Shannon) entropy} are
\begin{equation}
  p_c^{(j)}
    \overset{\Delta}{=}
      P(c_j \mid \ul{y})
    = \frac{\sum_{\ul{s} \in c_j} 
      P(\ul{s} \mid \ul{y})}
      {
      \sum_{j = 1}^{|C|} \sum_{\ul{s} \in c_j}
        P(\ul{s} \mid \ul{y})};\;\;
  \mr{SE}_S(\ul{y}) 
    = -\sum_{c_j \in C} 
      p_c^{(j)}\, \log\,p_c^{(j)},        
  \label{eq:2}
\end{equation}
respectively. When the TS probabilities $P(s_i \mid \ul{s}_{<i}, \ul{y})$ are available, $p_c^{(j)}$ are computed from equations~\ref{eq:1} and \ref{eq:2}; otherwise, we use $p_c^{(j)} \approx \mbb{I}(c = c_j)/|C|$, the membership count in each cluster. The latter is referred to as \emph{discrete semantic (Shannon) entropy} in \cite{farquhar2024detecting}. 


\subsection{Semantic R{\'e}nyi Entropy} 



The physics-inspired framework for UQ in \cite{Principe2010_Book} requires a non-parametric feature mapping of the TS probabilities in the form of its kernel mean embedding (KME) in the reproducing kernel Hilbert space (RKHS) determined by a suitable positive semi-definite kernel (\cite{Aronszajn1950ToAMS, Scholkopf2015SC_RandomVariablesViaRKHS}). We employ the Gaussian kernel $\kappa_{\sigma}(p_s; x)$, where $p_s \overset{\Delta}{=} P(\ul{s} \mid \ul{y})$ and $x \in \mbb{R}$. The resulting KME and its empirical estimate are, respectively,
\begin{equation}
  \psi_{\ul{y}}(x)
    = \int_{\ul{s}}  
      \kappa_{\sigma}(p_s, x)\, p_s\, dp_s;\quad
  \wh{\psi}_{\ul{y}}(x) 
    = \frac{1}{R} 
      \sum_{r = 0}^{R-1} 
      \kappa_{\sigma}(p_s^{(r)}; x)\, p_s^{(r)},
  \label{eq:psr}
\end{equation}
where $\{\ul{s}^{(r)},\, r = 0, \ldots, R-1\}$ denotes the LLM generations when the input $\ul{y}$ is repeated $R$ times and $p_s^{(r)} \overset{\Delta}{=} P( \ul{s}^{(r)} \mid \ul{y})$. As \cite{Principe2010_Book} establishes, $\wh{\psi}_{\ul{y}}(\cd)$ serves as a kernel-based empirical estimator of the more general notion of R{\'e}nyi (quadratic) entropy
\begin{equation}
  \mr{SE}_R(\ul{y}) 
    = -\log
      \sum_{j = 1}^{|C|}
      p_c^{(j)^2}
    = -\log\,(\mbb{E}[p_c]),
    \label{eq:SRE}
\end{equation}
which in turn is a direct measure of uncertainty in a distribution. In this work, we exclusively focus on the quadratic form of R{\'e}nyi entropy and, for brevity, refer to it throughout the paper as semantic R{'e}nyi entropy. Here, $\mbb{E}[\cd]$ denotes expectation (over the space of $C$). Indeed, $\wh{\psi}_{\ul{y}}(x)$ can be viewed as how uncertainty varies with $x$. For purposes of UQ, we view this KME --- to be precise, a sampled version $\{\wh{\ul{\psi}}_{\ul{y}}\}$ of $\wh{\psi}_{\ul{y}}(x)$ --- as a `data' wave function associated with the time-independent Schrödinger equation of a QTN (\cite{Qi2019Quantum_DeterminingLocalHamiltonian, Vipulanandan2024, vipulanandan2026sensitivity}).

Given that it is R{\'e}nyi entropy which underpins our physics-inspired framework for UQ of TS probabilities, we opt to use the same (instead of Shannon entropy as in SOTA methods) for assessing semantic entropy of output clusters associated with LLM generations too. 


\subsection{Accounting for Uncertainty in TS Probabilities} 


\paragraph{UQ of TS Probabilities}
For purposes of UQ of TS probabiltiies, we identify a QTN whose Hamiltonian $\wh{\ul{H}}$ has $\{\wh{\ul{\psi}}_{\ul{y}}\}$ as one of its eigen-modes, and apply perturbation theory to compute the corrections (we only look at the first-order corrections) to all the eigen-modes/energies of $\wh{\ul{H}}$. Then, we construct the first-order uncertainty `feature' vectors
\begin{equation}
  \ul{V}_m^{(1)}(x)
    = E_m^{(1)} + \frac{\sigma^2}{2}\, \frac{\nabla_m^2 |\ul{\psi}_m^{(1)}(x)|}{|\ul{\psi}_m^{(1)}(x)|}, 
  \tx{ where }
  E_m^{(1)}
    = -
      \min_{p_x}
      \frac{\sigma^2}{2}\, \frac{\nabla_m^2 |\ul{\psi}_m^{(1)}(x)|}{|\ul{\psi}_m^{(1)}(x)|}.
  \label{eq:UQ_equation}
\end{equation}
Here, $\{\ul{\psi}_m,\, E_m\}$ and $\{\ul{\psi}_m^{(1)},\, E_m^{(1)}\}$, for $m \geq 0$, are the eigen-pairs of the Hamiltonian $\wh{\ul{H}}$ ordered as $E_0 \leq E_1 \leq \cdots$, and their first-order correction terms, respectively. For each mode $m \geq 0$, the Laplacian $\nabla_m^2 |\ul{\psi}_m^{(1)}(x)|$ measures the change in the first-order correction from its average in the local neighborhood across the modes and the vector $\ul{V}_m^{(1)}(x)$ can be viewed as a `spectrogram' of uncertainties across TS probability amplitudes. If $p_s^{(r)}$ is mapped to $x^{(r)}$ in the RKHS, the uncertainty $\mr{UQ}\,(p_s^{(r)})$ associated with the TS probability $p_s^{(r)}$ is taken as
\begin{equation}
  \mr{UQ}\,(p_s^{(r)})
    = \frac{1}{M} 
      \sum_m
      \left.
        \ul{V}_m^{(1)}(x)
      \right|_{x = x^{(r)}},
  \label{eq:UQ_equation2}
\end{equation}
where the summation is carried over $M$ (we use $M = 8$) modes that are adjacent to $\{\wh{\ul{\psi}}_{\ul{y}}\}$. Additional details on QTN based UQ are provided in Appendix~\ref{app:UQ_QTNappendix}.


\paragraph{Calibrated Adjustment of TS Probabilities} 
We now adjust the TS probabilities as  
\begin{equation}
  p_s^{(r)^*}
    = \arg\max_{\wh{p}_s^{(r)}} 
      \left\{ 
         -\log\,(\wh{p}_s^{(r)^2} + (1 - \wh{p}_s^{(r)})^2) - \lambda \cdot \frac{1}{\mr{UQ}\,(p_s^{(r)})} \cdot \mathrm{KL}\,(\wh{p}_s^{(r)} \Vert p_s^{(r)}) 
      \right\}.
  \label{eq:entropy_max}
\end{equation}
The $\log(\cd)$ term in the right-hand side is the R{\'e}nyi entropy; the second term penalizes the Kullback–Leibler (KL) divergence between the adjusted probability $\wh{p}_s^{(r)}$ and $p_s^{(r)}$ scaled inversely with the uncertainty $\mr{UQ}\,(p_s^{(r)})$; the hyperparameter $\lambda > 0$ controls the trade-off between entropy maximization and the adjusting the TS probability. In essence, we are maximizing entropy of the TS probabilities while penalizing the error between $p_s^{(r)}$ and its adjusted value $\wh{p}_s^{(r)}$.  This scheme which appears in  Fig.~\ref{fig:semantic_pipeline} enables one to account for uncertainty associated with TS probabilities so that a more informed decision regarding existence/absence of hallucinatory behavior could be made. Appendix~\ref{sec:cost_comp_and_pseudocode} contains additional discussion of computational overheads together with the pseudocode implementations of key components.
 

\subsection{Intuition Behind the Proposed Approach} 

Our method integrates perturbation-based uncertainty quantification and maximum entropy inference into a unified framework for correcting TS probabilities in a principled manner.

\paragraph{Perturbation-based uncertainty quantification.}
To estimate uncertainty in the TS probabilities, we map the KME of the TS probability distribution into the eigen structure of a QTN. The empirical KME is embedded as an eigen-mode of an admissible Hamiltonian $\widehat{\underline{H}}$ associated with the QTN. Therefore, perturbing $\widehat{\underline{H}}$ corresponds to perturbing the underlying TS probabilities; the resulting first-order corrections to its eigen-modes/energies quantify how sensitive the modeled distribution is to infinitesimal changes in its inputs. This mirrors standard quantum mechanical perturbation theory where the instability of an eigen-state under perturbation reveals its local variability.

Large first-order corrections, therefore, indicate highly unstable TS probabilities; small corrections indicate locally stable regions. This produces an interpretable, physically grounded uncertainty measure, formalized in eq.~(\ref{eq:UQ_equation2}), that captures the \emph{local variability} of the TS probabilities in the RKHS amplitude domain.

\paragraph{Maximum entropy inference under partial knowledge.}
When only partial information is available, the maximum entropy principle provides the least biased estimate of an unknown probability distribution. Among all distributions consistent with the known constraints, the one with the highest entropy is maximally non-committal with respect to the missing information (\cite{Jaynes1957TPR_InformationTheory,Jaynes2003_Book}). Equivalently, the maximum entropy distribution minimizes its KL divergence from the uniform distribution. This idea has deep roots in robust statistics and variational inference, where uncertainty in the likelihood leads naturally to regularization toward higher-entropy, less overconfident solutions.

Our formulation implements precisely this. The R\'{e}nyi entropy term in eq.~(\ref{eq:entropy_max}) specifies how much the probabilities \emph{should} be lifted to remain maximally non-committal, whereas the KL term ensures that the adjusted probabilities stay close to the model’s empirical TS estimates. When uncertainty is high, the KL penalty relaxes and the entropy term dominates; when uncertainty is low, the KL term keeps the adjusted probabilities anchored near their empirical values. 

\paragraph{Bringing the two components together.}
The overall procedure thus follows a coherent logic: (i) infer local uncertainty in the TS probabilities by perturbing the Hamiltonian and reading out the first-order eigen-mode corrections; (ii) use these corrections to scale the KL penalty in the entropy maximization step; and (iii) adjust the TS probabilities to the maximum entropy distribution allowed by the empirical TS estimates. High-uncertainty regions are pushed toward higher entropy, while low-uncertainty regions remain close to the model’s original predictions.

This yields a deterministic, single-shot, principled method for correcting TS probabilities in accordance with their underlying uncertainty, unifying perturbation-based UQ with maximum-entropy reasoning in a single framework.

\section{Experiments}
\label{sec:experimental_setup}


\paragraph{Datasets} 
Our evaluations cover Q\&A in trivia knowledge (TriviaQA (\cite{joshi2017triviaqa})), general knowledge (SQuAD 1.1 (\cite{SQuAD_URL})), and open-domain natural questions (NQ-Open (\cite{lee2019latent})) derived from actual queries to Google Search (\cite{kwiatkowski2019natural}), and mathematical word problems (SVAMP (\cite{patel2021nlp}). Only the results summary is reported here; while detailed results are deferred to Appendix~\ref{supp:further_details}.


\paragraph{Models} 
We use a diverse set of LLMs obtained via Huggingface (\ttt{huggingface.com}), including Falcon-RW 1B (\cite{penedo2023refinedweb}), LLaMA-3.2 1B  (\cite{grattafiori2024llama}), LLaMA-2-7B-chat, LLaMA-2 7B, LLaMA-2-13B-chat, LLaMA-2 13B (\cite{touvron2023llama}), Mistral-7B-instruct-v0.3, and Mistral-v0.1 7B (\cite{jiang2024identifying}). The inclusion of smaller-scale models and quantized versions of larger models serves two purposes. From a practical standpoint, resource limitations necessitate the use of lighter models to enable extensive sampling and repeated querying. From a scientific viewpoint, evaluating smaller and compressed models allows us to explore hallucinations and uncertainty under constrained model capacity and precision, providing insights into how semantic entropy behaves not only in ideal settings but also in more realistic, resource-constrained deployments. This is particularly relevant as smaller and quantized models are increasingly being deployed in edge, mobile, and enterprise environments. All experiments \footnote{code available: https://github.com/pragasv/semantic-entropy-UQ-.git} were run on a workstation with 64 GB memory and NVIDIA A6000 GPU.


\paragraph{Model Prompting, Entailment, and Answer Selection}  
Appendix~\ref{sec:Templates} provides the prompt template used for all datasets to generate LLM responses and the prompt we used to detect entailment. While other classifiers could have been used, we employ a DeBERTa-large model fine-tuned on the MNLI dataset (\cite{williams2017broad}) for entailment prediction. This method builds upon previous research in paraphrase detection based on embedding similarity (\cite{socher2011dynamic, yu2014deep}) and BERT-style models (\cite{he2020realformer, tay2021charformer}). For simplicity, entailment is checked by concatenating the question with one answer and comparing it to the concatenation of the question with another answer. Instruction-tuned LLMs, such as LLaMA 2, GPT-3.5 Turbo or GPT-4, could also have been used to predict entailment between generated outputs.


\paragraph{Entropy Maximization} 
Following UQ of the LLM generations, entropy maximization (see \eqref{eq:entropy_max}) was applied to identify the `optimum' answer based on the associated uncertainty estimates. The effectiveness of this approach (in the sense of whether incorporating uncertainty information improves hallucination detection) was assessed using the evaluation metrics described below. We also investigated the optimal choice of the hyperparameter $\lambda$. Implementation details and extended analyses are deferred to Appendix~\ref{sec:lambda_select_test}.


\paragraph{Comparison Methods} 
For evaluation, we use the following methods for comparison against the proposed Rényi semantic entropy pre- and post-TS uncertainty integration ($\mr{SE}_R$ and $\mr{SE}_R^+$, respectively): naive entropy (NE), semantic entropy ($\mr{SE}_S$), and discrete semantic entropy ($\mr{DSE}_S$), all employed in \cite{farquhar2024detecting}, and two strong baseline methods, embedding regression (ER) and verifier $p\,(\mr{True})$. ER is a supervised method inspired by the P(IK) approach (\cite{kadavath2022language}). Rather than fine-tuning the entire LLM as in the original work, this method trains a logistic regression classifier on the final hidden representations to predict answer correctness, improving both simplicity and reproducibility. This baseline performs well with in-distribution data but poorly with out-of-distribution data. In $p\,(\mr{True})$ (\cite{kadavath2022language}), the model samples multiple candidate answers and is then asked, via a multiple-choice prompt, whether the top answer is ``True'' or ``False''. Confidence is measured by the probability assigned to the ``True'' response. A few-shot strategy, using up to 20 labeled examples, further enhances performance, although context window limitations sometimes require reducing the number of few-shot examples.


\paragraph{Evaluation Metrics} 
We use three primary metrics --- each based on an automated assessment of factual consistency with reference answers from the datasets employed --- for evaluation purposes: 
    \tb{(a)}~\emph{AUROC (Area Under the Receiver Operating Characteristics)} measures the ability of a classifier to distinguish between correct and incorrect answers while accounting for both precision and recall. Intuitively, it represents the probability that a randomly selected correct answer receives a higher confidence score than a randomly selected incorrect one. A perfect model achieves an AUROC of 1. 
    \tb{(b)}~\emph{RAC (Rejection Accuracy Curve)} measures the question-answering accuracy that a specified percent of the most confident model predictions. An effective uncertainty estimation approach should ensure that the high-confidence predictions being retained are more accurate, with RAC improving as less confident examples are progressively excluded. 
    \tb{(c)}~\emph{AURAC Area Under the Rejection Accuracy Curve)} quantifies how accuracy changes as increasingly uncertain predictions are rejected. Higher AURAC values indicate that the uncertainty method more effectively distinguishes accurate from inaccurate responses. Unlike AUROC, AURAC is directly sensitive to the model's overall accuracy, providing a complementary perspective on the quality of uncertainty estimation methodology.


\section{Results}
\label{sec:results}


\subsection{Uncertainty-Aware Semantic Entropy: A Worked Example}


This section presents a worked example illustrating the computation of Semantic R{\'e}nyi entropy with UQ maximization (SRE-UQ), denoted as $\mr{SE}_R^+$, together with NE and SE. The query used for this illustration is, \textit{``Which oil producer is a close ally of the United States?''} repeated ten times. The corresponding TS probabilities are in Table ~\ref{tab:worked_example}.

The results demonstrate that incorporating uncertainty quantification into semantic entropy (i.e., $\mr{SE}_R^+$) yields systematically lower entropy estimates relative to both $\mr{NE}$ and `standard' Shannon entropy-based $\mr{SE}_S$. This reduction arises because semantically equivalent generations are clustered together for this question, while their TS probabilities exhibit reduced aleatoric variability. Importantly, accounting for such uncertainty mitigates overestimation biases in entropy, providing a more faithful measure of confabulation in LLM outputs.

\begin{table}[H]
  \centering
  \resizebox{\textwidth}{!}{%
  \begin{tabular}{l c c c cc cccc}
  \toprule
  \tb{LLM Generation} 
    & \tb{Cluster ID} 
    & \tb{$\wt{p}_s^{(r)}$} 
    & \tb{$p_s^{(r)}$}   
    & \tb{$p_c^{(j)}$} & \tb{$p_c^{(j)^*}$} 
    & \tb{$\mr{NE}$} & \tb{$\mr{SE}_S$} & \tb{$\mr{SE}_R^+$} \\
  \midrule
  Russia 
    & 1
    & 0.05899
    & 0.01814  
    & 0.01814 & 0.02223 
    & -0.03159 & -0.03159 & 0.00049 \\
  Saudi Arabia  
    & 2
    & 0.57761
    & 0.17765  
    & 0.88824 & 0.85880 
    & -0.13331 & -0.04572 & 0.73754 \\
  Saudi Arabia  
    & 2
    & 0.57761
    & 0.17765  
    & 0 & 0       
    & -0.13331 & --       & -- \\
  Iran          
    & 3
    & 0.07227
    & 0.02223  
    & 0.02223 & 0.02697 
    & -0.03674 & -0.03674 & 0.00073 \\
  Saudi Arabia  
    & 2
    & 0.57761
    & 0.17765  
    & -- & --       
    & -0.13331 & --       & -- \\
  Kuwait        
    & 4
    & 0.08940
    & 0.02749  
    & 0.02749 & 0.03488 
    & -0.04291 & -0.04291 & 0.00122 \\
  Qatar         
    & 5
    & 0.02185
    & 0.00672  
    & 0.00672 & 0.01214 
    & -0.01460 & -0.01460 & 0.00015 \\
  Saudi Arabia  
    & 2
    & 0.57761
    & 0.17765  
    & -- & --       
    & -0.13331 & --       & -- \\
  Iraq          
    & 6
    & 0.12086
    & 0.03717  
    & 0.03717 & 0.04498 
    & -0.05315 & -0.05315 & 0.00202 \\
  Saudi Arabia  
    & 2
    & 0.57761
    & 0.17765  
    & 0 & 0       
    & -0.13331 & --       & -- \\
  \midrule
  \tb{Total} 
    &
    & 3.25144
    & 1.00000      
    & 1.00000 & 1.00000   
    & \tb{0.84557} & \tb{0.22471} & \tb{0.12951} \\
  \bottomrule
  \end{tabular}
  }
  \caption{An example calculation of TS probability and uncertainty metrics corresponding to the query above.   
  Columns: $\wt{p}_s^{(r)}$ are the raw (unnormalized) sequence probabilities;  
  $p_s^{(r)}$ are the normalized sequence probabilities (\eqref{eq:psr});  
  $p_c^{(j)} = P(c_j \mid \ul{y})$ are the cluster probabilities (\eqref{eq:2}) post-normalized;  
  $p_c^{(j)^*}$ are the cluster probabilities post-TS uncertainty integration (\ref{eq:entropy_max}). 
  Methods: $\mr{NE}$ and $\mr{SE}_S$ denote naive entropy and semantic entropy, respectively (\cite{farquhar2024detecting}); 
  $\text{SE}_R^+$ denotes semantic Rényi entropy post-TS uncertainty integration. 
  Values for only one representative LLM generation per cluster are shown.}
  \label{tab:worked_example}
\end{table}


Table~\ref{tab:worked_example} illustrates this effect for the example query. It compares the semantic response distributions produced by `standard' R{\'e}nyi entropy-based $\mr{SE}_R$ and its uncertainty-adjusted variant $\mr{SE}_R^+$. While several hallucinated outputs (e.g., Qatar, Iraq, Iran) receive non-trivial probability mass, $\mr{SE}_S$ underestimates the associated confabulation risk. In contrast, the proposed uncertainty-aware adjustment penalizes responses with higher uncertainty, redistributing probability mass toward more semantically coherent answers. For instance, Saudi Arabia—a contextually appropriate response—receives a higher probability assignment after adjustment. This highlights a promising direction: even when hallucinations are present, localized uncertainty quantification enables prioritization of high-certainty, semantically meaningful answers, a factor not considered in prior work.



\subsection{Detecting Confabulations}
\label{sec:results_benchmark}


\textbf{Benchmark Datasets.} 
Building on the worked example above, we now present the results obtained from TriviaQA, NQ, SVAMP, and SQuAD datasets, across multiple LLM models such as Mistral 7B, LLaMA-2 7B, LLaMA-3.2 1B, Falcon-rw-1B and LLaMA-2 13B as well as uncertainty estimation methods; detailed validation results appear in Appendix~\ref{supp:further_details}.

\begin{figure}[htpb]
  \begin{subfigure}[b]{0.48\columnwidth}
    \centering
    \includegraphics[height=4.5cm, keepaspectratio]{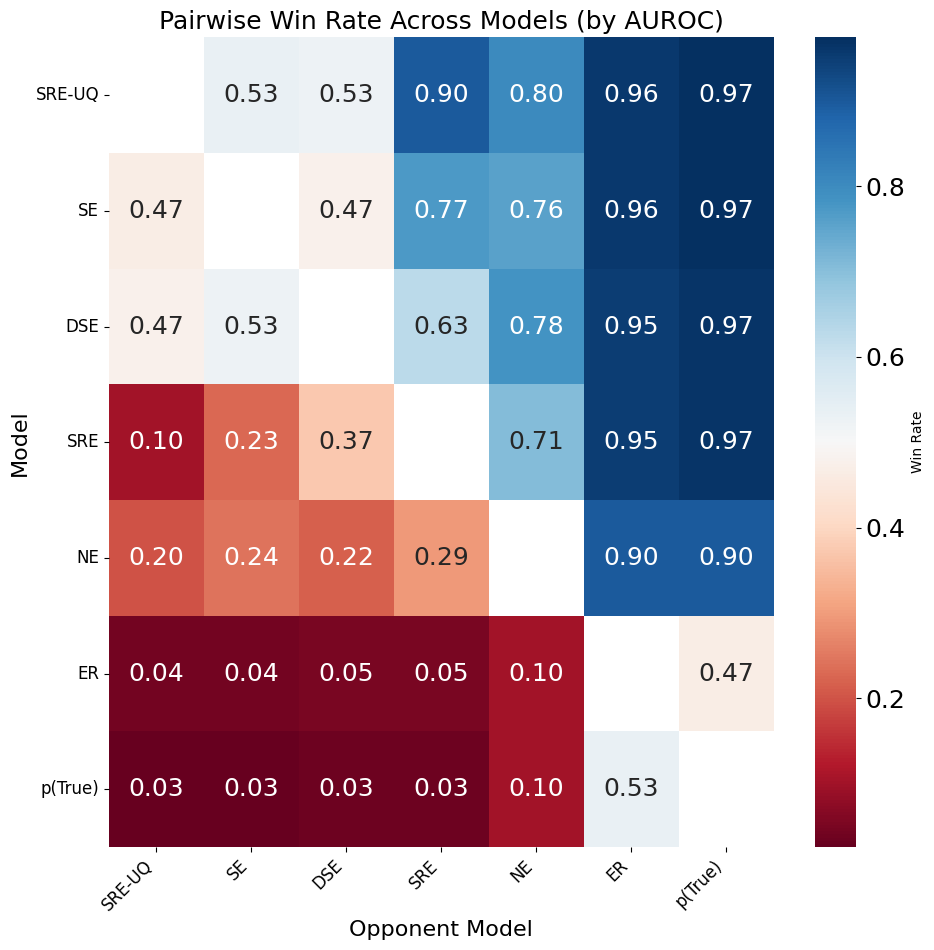}
    \caption{AUROC-based win rate.}
    \label{fig:winrate_AUROC}
  \end{subfigure}
  \hfill
  \begin{subfigure}[b]{0.48\columnwidth}
    \centering
    \includegraphics[height=4.5cm, keepaspectratio]{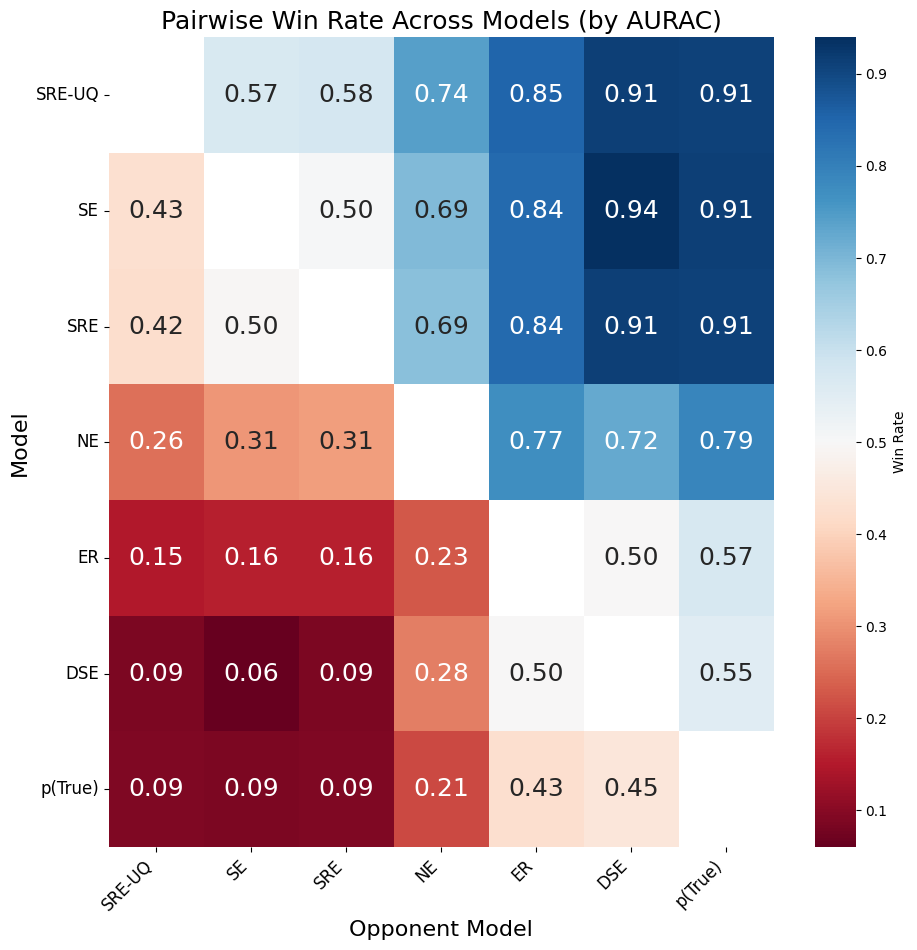}
    \caption{AURAC-based win rate.}
    \label{fig:winrate_AURAC}
  \end{subfigure}
  \caption{Pairwise win rate matrices across SOTA hallucination detection methods on diverse datasets and LLM models. This summarizes 116 experimental scenarios. Each cell indicates the probability that the row model outperforms the column model. Semantic R{\'e}nyi entropy with UQ maximization consistently outperforms baselines, even surpassing methods reliant on supervised learning.}
  \label{fig:winrate_combined}
\end{figure}

Fig.~\ref{fig:winrate_AUROC} reports the SRE-UQ's performance against SOTA methods evaluated via AUROC performance. Higher AUROC scores indicate better separability between factually correct and hallucinated outputs. Across all evaluated models, SRE-UQ is competitive with SOTA baseline methods. Notably, SRE-UQ achieves better AUROC despite not relying on ground truth labels for training. This is evidence of its effectiveness in hallucination detection with no reliance on supervised fine-tuning.

While AUROC reflects a model’s global ranking ability, practical deployments often involve thresholding on confidence scores to reject outputs associated with higher uncertainty. To assess this, we compute the RAC over rejection thresholds ranging from 80\% to 100\% confidence. Fig.~\ref{fig:winrate_AURAC} presents the SRE-UQ's performance against SOTA methods evaluated via AURAC performance. Consistently, SRE-UQ maintains strong AURAC as more uncertain outputs are filtered out. Particularly it offers superior robustness compared to discrete semantic entropy, naive entropy, and supervised p(True) baseline. This is evidence of its effectiveness in pruning predictions that are associated with higher uncertainty. Appendix~\ref{supp:further_details} provides in-depth results at the dataset level for Trivia-QA, SQuAD, NQ-Open, and SVAMP.

\begin{figure}[htpb]
  \centering
  \begin{subfigure}[b]{0.8\columnwidth}
    \centering
    \includegraphics[width=\linewidth]{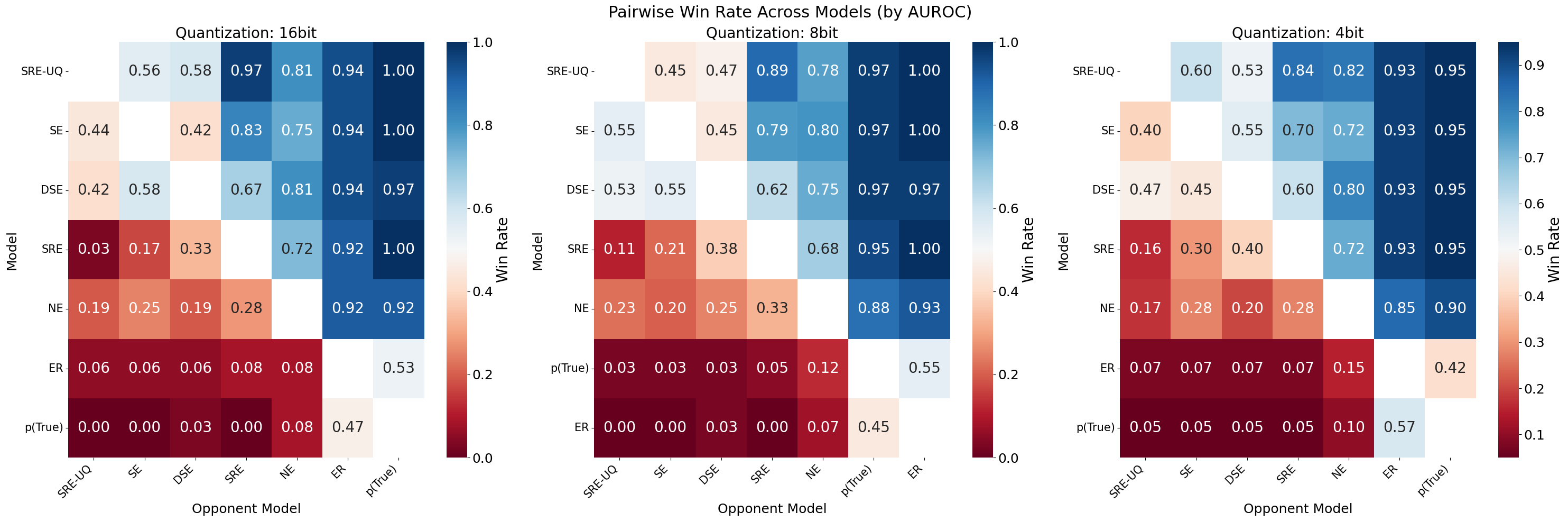}
    \caption{Pairwise AUROC-based win rate matrices across quantization levels.}
    \label{fig:quantwise_auroc}
  \end{subfigure}
  \begin{subfigure}[b]{0.8\columnwidth}
    \centering
    \includegraphics[width=\linewidth]{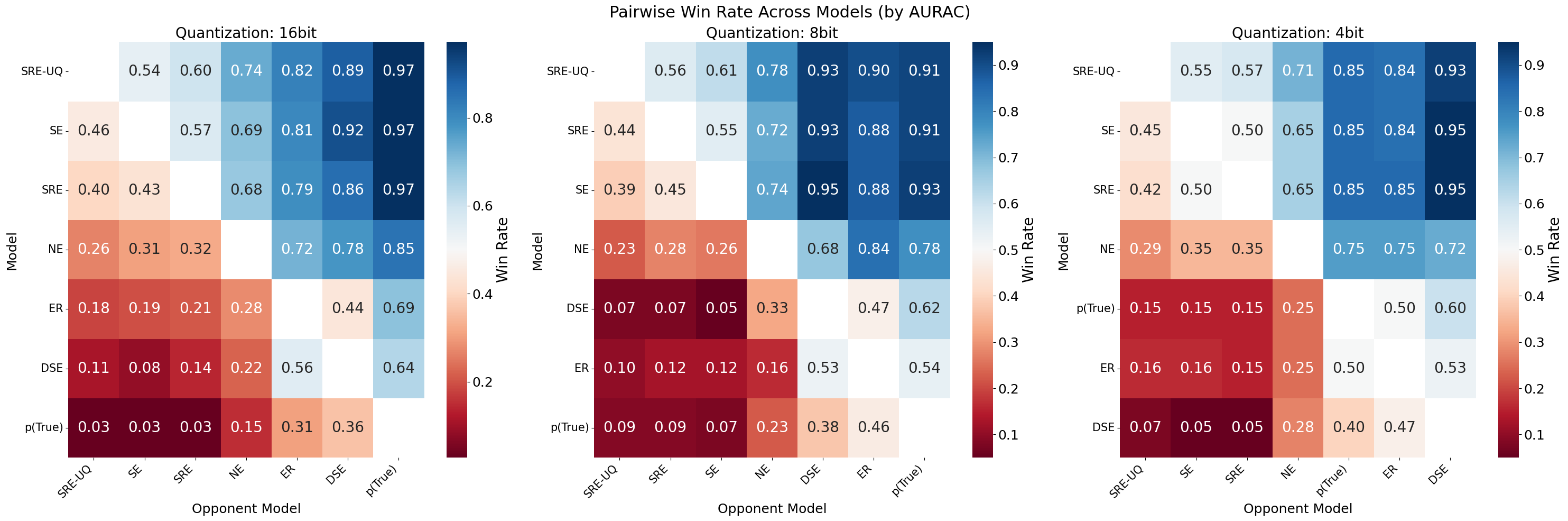}
    \caption{Pairwise AURAC-based win rate matrices across quantization levels.}
    \label{fig:quantwise_aurac}
  \end{subfigure}
  \caption{Evaluation of SOTA hallucination detection methods across different quantization levels. This summarizes 116 experimental scenarios. Each subfigure shows pairwise win rate matrices under a different quantization precision (16-bit, 8-bit, 4-bit). SRE-UQ maximization consistently outperforms baselines across both AUROC and AURAC, demonstrating robustness in hallucination detection against reduced precision.}
  \label{fig:quantwise_results}
\end{figure}

\paragraph{Quantized LLMs} 
While most prior work on hallucination detection benchmarks full-precision models, real-world deployments almost always rely on quantization (e.g., 16-bit, 8-bit, or 4-bit) to reduce inference latency and memory consumption. However, quantization does not merely compress model weights — it also perturbs probability distributions and modifies token-level uncertainty. These shifts can affect semantic entropy measurements and alter the relative competitiveness of different detection methods. Thus, a rigorous study of hallucination detection must examine the stability of detection methods across quantization levels, ensuring that performance gains are not an artifact of precision settings but persist in practical deployment regimes. Figure~\ref{fig:quantwise_results} presents pairwise win rate matrices of hallucination detection methods under different quantization settings (16-bit, 8-bit, and 4-bit), evaluated using both AUROC and AURAC metrics. The results show that SRE-UQ maximization remains robust across quantization levels, consistently outperforming or matching SOTA baselines even under aggressive compression.

The quantization-wise analysis reveals that SRE -UQ maximization consistently outperforms alternative detection strategies across 16-bit, 8-bit, and 4-bit models. While quantization slightly reduces absolute AUROC values due to coarser uncertainty calibration, the relative ordering of methods remains largely stable, demonstrating that the proposed approach is resilient to reduced precision. This robustness is critical for deployment at scale, where low-bit quantization is increasingly adopted to balance efficiency with reliability in safety-sensitive applications.


\subsection{Entropy Uncertainty Analysis}

\begin{figure}[htpb]
  \centering
  \includegraphics[width = 0.8\columnwidth, keepaspectratio]{%
    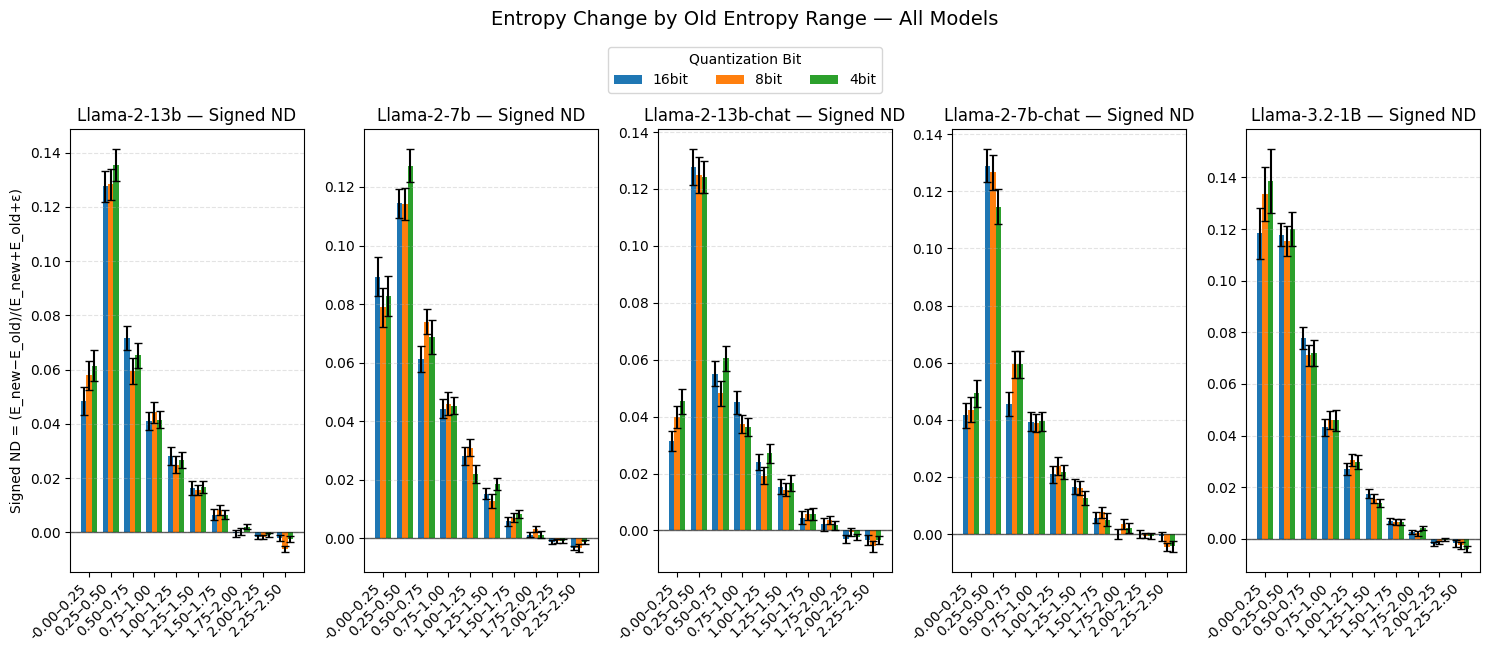}
  \caption{Signed normalized difference (ND) in entropy across old entropy bins for Llama-2 variants under different quantization settings (16-bit, 8-bit, 4-bit). Bars represent the mean signed ND with error bars indicating standard error. Notably, all models exhibit the largest variability in the 0.25–0.50 entropy range, suggesting this is a high-risk region where models frequently oscillate between multiple confident yet semantically divergent answers.}
  \label{fig:SND}
\end{figure}

To further assess the effect of uncertainty-aware adjustments proposed in \eqref{eq:entropy_max}, we analyze how SRE changes across different levels of generation diversity. Here, we measure the number of distinct semantic clusters produced by the model when answering the same question multiple times (10 times). As shown in Fig.~\ref{fig:SND}, the x-axis represents the observed entropy for a given question. A value of 0 on the x-axis indicates that all sampled generations were semantically identical - hence entropy is zero. This analysis directly probes how model uncertainty and semantic variability interact under the proposed adjustment mechanism.

While much of the prior work in hallucination detection relies on applying a global threshold to entropy-based scores, such approaches ignore how model behavior varies across different ranges of historical entropy. To address this, we evaluate how the change in entropy (normalized difference) distributes across old entropy bins for each LLM. This perspective provides a more fine-grained decision-making criterion: instead of arbitrarily choosing a fixed threshold, one can calibrate decisions depending on the region of the entropy spectrum.

The results demonstrate that entropy change is not uniform across the input space: low-entropy regimes (close to zero) remain relatively stable, while intermediate ranges (particularly 0.25–0.75) are highly volatile. This implies that decision rules based solely on absolute entropy thresholds are brittle and may fail to capture these nuanced behaviors. Instead, practitioners should treat certain entropy regions with caution and, in some cases, defer to auxiliary signals before committing to a prediction. 


\section{Conclusion}
\label{sec:conclusion}


This work presents a novel framework for hallucination detection in LLMs which leverages semantic R{\'e}nyi entropy and a quantum TN-based UQ method that explicitly models and accounts for aleatoric uncertainty of sequence probabilities. Through extensive evaluations on several datasets, diverse LLM architectures, and various quantization precision (16-bit, 8-bit, and 4-bit),  we demonstrate that semantic entropy measures, especially when combined with principled uncertainty penalization, significantly improve the detection of hallucinated outputs. By adjusting output probabilities to maximize entropy while penalizing deviations weighted by uncertainty, the proposed method yields consistent gains in AUROC, AURAC and RAC across 116 experiments, outperforming strong baselines, without requiring supervised fine-tuning. Additionally, this provides a principled approach to select answers from an LLM even when it is hallucinating.  

Incorporating uncertainty not only sharpens the sensitivity of entropy-based metrics but also enables principled answer selection, even under confabulation. By adjusting TS probabilities to balance entropy maximization with uncertainty-aware penalization, our method promotes more coherent and trustworthy outputs. Furthermore, the uncertainty quantification method operates utilizing QTNs and perturbation theory, making it suitable for scalable, explainable real-world deployments.

Overall, this study highlights the importance of uncertainty-aware semantic evaluation in achieving reliable language generation, and opens up new avenues for integrating information-theoretic principles with uncertainty quantification to advance the robustness of LLMs.


\paragraph{Limitations} 
While the proposed method combining quadratic semantic R{\'e}nyi entropy and quantum TN-based UQ shows great promise in detecting hallucinations, a few limitations remain. Most notably, our evaluations are conducted using relatively small language models due to computational resource constraints. While these models enable efficient experimentation, they may not fully capture the behavior of larger frontier models such as GPT-4 or Claude. Consequently, both the absolute detection accuracy and the complexity of hallucinations observed could differ at larger scales. Furthermore, our semantic clustering relies on an external entailment predictor (DeBERTa-large fine-tuned on MNLI), and any inaccuracies in this entailment model may propagate into the entropy estimates. Finally, the proposed approach requires access to token-level probabilities, which restricts its applicability to open-weight models; black-box LLMs, where such information is not exposed, remain outside the scope of this method.

The uncertainty of TS probabilities in \eqref{eq:UQ_equation2} may very well serve to arrive at a meaningful definition of an uncertainty associated with each semantic cluster. For instance, for the cluster $C_j$, one could employ $\mr{UQ}\,(C_j \mid \ul{y}) = (1/|C_j|)  \sum_{s^{(r)} \in C_j} \mr{UQ}\,(p_s^{(r)})$. This may facilitate one to identify semantic clusters whose membership is more sensitive to changes in TS probabilities which in turn may lead to practical guidelines for identifying situations where automated outputs should be supplemented with careful, perhaps human-assisted, review. We hope to undertake this direction of research in the future. 
    

\subsubsection*{Acknowledgments}


This work is based on research supported by National Science Foundation (NSF) award xxxxxx.

\vfill
\clearpage
\bibliographystyle{iclr2025_conference}
\bibliography{references}

@article{Harremoes2009Kybernetika_JointRangeRenyiEntropies,
  author    = "P. Harrem{\"{o}}es",
  title     = "Joint Range of {R}{\'{e}}nyi Entropies",
  journal   = "Kybernetika",
  volume    = "45",
  number    = "6",
  pages     = "901--911",
  month     = "",
  year      = "2009",
}

@article{Jaynes1957TPR_InformationTheory, 
  author    = "E. T. Jaynes", 
  title     = "Information Theory and Statistical Mechanics", 
  journal   = "The Physical Review", 
  volume    = "106", 
  number    = "4", 
  pages     = "620--630", 
  month      = may,
  year      = "1957", 
}

@book{Jaynes2003_Book,
    title       = "Probability Theory: The Logic of Science", 
    author      = "E. T. Jaynes", 
    publisher   = "Cambridge University Press", 
    year        = "2003", 
}

@article{Ghahramani2015N_ProbabilisticML, 
  author    = "Z. Ghahramani", 
  title     = "Probabilistic Machine Learning and Artificial Intelligence", 
  journal   = "Nature", 
  volume    = "521", 
  number    = "7553", 
  pages     = "452--459", 
  moth      = may,
  year      = "2015", 
}

@book{CohenTannoudji1977_Book,
  author       = "C. Cohen-Tannoudji and B. Diu and F. Lalo{\"e}",
  title        = "Quantum Mechanics, Volume 2: Angular Momentum, Spin, and Approximation Methods",
  publisher    = "Wiley",
  address      = "New York, {NY}", 
  year         = "1977",
}

@article{Qi2019Quantum_DeterminingLocalHamiltonian, 
  author	= "X.-L. Qi and D. Ranard", 
  title		= "Determining a Local {H}amiltonian From a Single Eigenstate", 
  journal	= "Quantum", 
  volume	= "3", 
  number	= "159", 
  pages		= "5995-6000", 
  month		= jul, 
  year 		= "2019", 
  doi		= "{10.22331/q-2019-07-08-159}",   
}

@article{Scholkopf2015SC_RandomVariablesViaRKHS, 
  author 	= "B. Sch{\"o}lkopf and K. Muandet and K. Fukumizu and S. Harmeling and J. Peters",
  title 	= "Computing Functions of Random Variables Via Reproducing Kernel {H}ilbert Space Representations",
  journal	= "Statistics and Computing", 
  volume	= "25", 
  number	= "", 
  pages		= "755--766",
  month		= jun, 
  year		= "2015",
}

@article{Aronszajn1950ToAMS,
  author	= "N. Aronszajn",
  title		= "Theory of Reproducing Kernels",
  journal	= "Transactions of the American Mathematical Society",
  volume	= "68",
  number	= "3",
  pages		= "337-404",
  month		= may, 
  year		= "1950",
}

@article{Parzen1962AMS, 
  author	= "E. Parzen", 
  title		= "On Estimation of a Probability Density Function and Mode", 
  journal	= "Annals of Mathematical Statistics", 
  volume	= "33",
  number	= "3",
  pages		= "1065--1076",
  month		= "",
  year		= "1962",
}

@incollection{Principe2010_Book, 
  author		= "J. C. Principe",
  title			= "Information Theoretic Learning: Renyi's Entropy and Kernel Perspectives",
  booktitle		= "Information Science and Statistics", 
  editor		= "M. Jordan and R. Novak and B. Sch{\"o}lkopf",
  publisher		= "Springer", 
  address		= "New York, {NY}",
  year			= "2010",
}

@article{Rudin2019NatureMI_StopExplainingBlackBoxMLModels, 
  author	= "C. Rudin", 
  title		= "Stop Explaining Black Box Machine Learning Models for High Stakes Decisions and Use Interpretable Models Instead",
  journal	= "Nature Machine Intelligence",
  volume	= "1",
  number	= "",
  pages		= "206--215",
  month		= may,
  year		= "2019",
}

@article{Linardatos2021Entropy_ExplainableAI, 
  author 	= "P. Linardatos and V. Papastefanopoulos and S. Kotsiantis",
  title 	= "Explainable {AI}: A Review of Machine Learning Interpretability Methods", 
  journal	= "Entropy", 
  volume	= "23", 
  number	= "18", 
  pages		= "",
  month		= "", 
  year		= "2021",
  note		= "doi: 10.3390/e23010018", 
}

@misc{SQuAD_URL,
 author			= "{Stanford {NLP} Group}",
 title 			= "{SQuAD}: The Stanford Question Answering Dataset",
 howpublished	= "\url{https://rajpurkar.github.io/SQuAD-explorer/}",
 month			= "", 
 year			= "",
 note			= "[Accessed 28 Apr 2025]",
}

@article{kossen2024semantic,
  title={Semantic entropy probes: Robust and cheap hallucination detection in llms},
  author={Kossen, Jannik and Han, Jiatong and Razzak, Muhammed and Schut, Lisa and Malik, Shreshth and Gal, Yarin},
  journal={arXiv preprint arXiv:2406.15927},
  year={2024}
}

@article{farquhar2024detecting,
  title={Detecting hallucinations in large language models using semantic entropy},
  author={Farquhar, Sebastian and Kossen, Jannik and Kuhn, Lorenz and Gal, Yarin},
  journal={Nature},
  volume={630},
  number={8017},
  pages={625--630},
  year={2024},
  publisher={Nature Publishing Group UK London}
}

@article{zhao2023survey,
  title={A survey of large language models},
  author={Zhao, Wayne Xin and Zhou, Kun and Li, Junyi and Tang, Tianyi and Wang, Xiaolei and Hou, Yupeng and Min, Yingqian and Zhang, Beichen and Zhang, Junjie and Dong, Zican and others},
  journal={arXiv preprint arXiv:2303.18223},
  volume={1},
  number={2},
  year={2023}
}

@article{kaddour2023challenges,
  title={Challenges and applications of large language models},
  author={Kaddour, Jean and Harris, Joshua and Mozes, Maximilian and Bradley, Herbie and Raileanu, Roberta and McHardy, Robert},
  journal={arXiv preprint arXiv:2307.10169},
  year={2023}
}

@article{hosseini2023exploratory,
  title={An exploratory survey about using ChatGPT in education, healthcare, and research},
  author={Hosseini, Mohammad and Gao, Catherine A and Liebovitz, David M and Carvalho, Alexandre M and Ahmad, Faraz S and Luo, Yuan and MacDonald, Ngan and Holmes, Kristi L and Kho, Abel},
  journal={Plos one},
  volume={18},
  number={10},
  pages={e0292216},
  year={2023},
  publisher={Public Library of Science San Francisco, CA USA}
}

@article{chen2021evaluating,
  title={Evaluating large language models trained on code},
  author={Chen, Mark and Tworek, Jerry and Jun, Heewoo and Yuan, Qiming and Pinto, Henrique Ponde De Oliveira and Kaplan, Jared and Edwards, Harri and Burda, Yuri and Joseph, Nicholas and Brockman, Greg and others},
  journal={arXiv preprint arXiv:2107.03374},
  year={2021}
}

@article{lai2023mini,
  title={Mini-dalle3: Interactive text to image by prompting large language models},
  author={Lai, Zeqiang and Zhu, Xizhou and Dai, Jifeng and Qiao, Yu and Wang, Wenhai},
  journal={arXiv preprint arXiv:2310.07653},
  year={2023}
}

@article{lei2023chain,
  title={Chain of natural language inference for reducing large language model ungrounded hallucinations},
  author={Lei, Deren and Li, Yaxi and Hu, Mengya and Wang, Mingyu and Yun, Vincent and Ching, Emily and Kamal, Eslam},
  journal={arXiv preprint arXiv:2310.03951},
  year={2023}
}

@article{ji2023survey,
  title={Survey of hallucination in natural language generation},
  author={Ji, Ziwei and Lee, Nayeon and Frieske, Rita and Yu, Tiezheng and Su, Dan and Xu, Yan and Ishii, Etsuko and Bang, Ye Jin and Madotto, Andrea and Fung, Pascale},
  journal={ACM computing surveys},
  volume={55},
  number={12},
  pages={1--38},
  year={2023},
  publisher={ACM New York, NY}
}

@article{berrios1998confabulations,
  title={Confabulations: a conceptual history},
  author={Berrios, German E},
  journal={Journal of the History of the Neurosciences},
  volume={7},
  number={3},
  pages={225--241},
  year={1998},
  publisher={Taylor \& Francis}
}

@article{lin2022teaching,
  title={Teaching models to express their uncertainty in words},
  author={Lin, Stephanie and Hilton, Jacob and Evans, Owain},
  journal={arXiv preprint arXiv:2205.14334},
  year={2022}
}

@article{evans2021truthful,
  title={Truthful AI: Developing and governing AI that does not lie},
  author={Evans, Owain and Cotton-Barratt, Owen and Finnveden, Lukas and Bales, Adam and Balwit, Avital and Wills, Peter and Righetti, Luca and Saunders, William},
  journal={arXiv preprint arXiv:2110.06674},
  year={2021}
}

@article{zhang2023siren,
  title={Siren's song in the AI ocean: a survey on hallucination in large language models},
  author={Zhang, Yue and Li, Yafu and Cui, Leyang and Cai, Deng and Liu, Lemao and Fu, Tingchen and Huang, Xinting and Zhao, Enbo and Zhang, Yu and Chen, Yulong and others},
  journal={arXiv preprint arXiv:2309.01219},
  year={2023}
}

@article{manakul2023selfcheckgpt,
  title={Selfcheckgpt: Zero-resource black-box hallucination detection for generative large language models},
  author={Manakul, Potsawee and Liusie, Adian and Gales, Mark JF},
  journal={arXiv preprint arXiv:2303.08896},
  year={2023}
}

@article{wang2022self,
  title={Self-consistency improves chain of thought reasoning in language models},
  author={Wang, Xuezhi and Wei, Jason and Schuurmans, Dale and Le, Quoc and Chi, Ed and Narang, Sharan and Chowdhery, Aakanksha and Zhou, Denny},
  journal={arXiv preprint arXiv:2203.11171},
  year={2022}
}

@article{li2023halueval,
  title={Halueval: A large-scale hallucination evaluation benchmark for large language models},
  author={Li, Junyi and Cheng, Xiaoxue and Zhao, Wayne Xin and Nie, Jian-Yun and Wen, Ji-Rong},
  journal={arXiv preprint arXiv:2305.11747},
  year={2023}
}

@article{lee2023mathematical,
  title={A mathematical investigation of hallucination and creativity in GPT models},
  author={Lee, Minhyeok},
  journal={Mathematics},
  volume={11},
  number={10},
  pages={2320},
  year={2023},
  publisher={MDPI}
}

@article{azaria2023internal,
  title={The internal state of an LLM knows when it's lying},
  author={Azaria, Amos and Mitchell, Tom},
  journal={arXiv preprint arXiv:2304.13734},
  year={2023}
}

@article{su2024unsupervised,
  title={Unsupervised real-time hallucination detection based on the internal states of large language models},
  author={Su, Weihang and Wang, Changyue and Ai, Qingyao and Hu, Yiran and Wu, Zhijing and Zhou, Yujia and Liu, Yiqun},
  journal={arXiv preprint arXiv:2403.06448},
  year={2024}
}

@article{kadavath2022language,
  title={Language models (mostly) know what they know},
  author={Kadavath, Saurav and Conerly, Tom and Askell, Amanda and Henighan, Tom and Drain, Dawn and Perez, Ethan and Schiefer, Nicholas and Hatfield-Dodds, Zac and DasSarma, Nova and Tran-Johnson, Eli and others},
  journal={arXiv preprint arXiv:2207.05221},
  year={2022}
}

@article{lindley1956measure,
  title={On a measure of the information provided by an experiment},
  author={Lindley, Dennis V},
  journal={The Annals of Mathematical Statistics},
  volume={27},
  number={4},
  pages={986--1005},
  year={1956},
  publisher={Institute of Mathematical Statistics}
}

@article{xiao2020wat,
  title={Wat zei je? detecting out-of-distribution translations with variational transformers},
  author={Xiao, Tim Z and Gomez, Aidan N and Gal, Yarin},
  journal={arXiv preprint arXiv:2006.08344},
  year={2020}
}

@inproceedings{quevedo2024detecting,
  title={Detecting hallucinations in large language model generation: A token probability approach},
  author={Quevedo, Ernesto and Salazar, Jorge Yero and Koerner, Rachel and Rivas, Pablo and Cerny, Tomas},
  booktitle={World Congress in Computer Science, Computer Engineering \& Applied Computing},
  pages={154--173},
  year={2024},
  organization={Springer}
}

@article{joshi2017triviaqa,
  title={Triviaqa: A large scale distantly supervised challenge dataset for reading comprehension},
  author={Joshi, Mandar and Choi, Eunsol and Weld, Daniel S and Zettlemoyer, Luke},
  journal={arXiv preprint arXiv:1705.03551},
  year={2017}
}

@article{lee2019latent,
  title={Latent retrieval for weakly supervised open domain question answering},
  author={Lee, Kenton and Chang, Ming-Wei and Toutanova, Kristina},
  journal={arXiv preprint arXiv:1906.00300},
  year={2019}
}

@article{patel2021nlp,
  title={Are NLP models really able to solve simple math word problems?},
  author={Patel, Arkil and Bhattamishra, Satwik and Goyal, Navin},
  journal={arXiv preprint arXiv:2103.07191},
  year={2021}
}

@article{kwiatkowski2019natural,
  title={Natural questions: a benchmark for question answering research},
  author={Kwiatkowski, Tom and Palomaki, Jennimaria and Redfield, Olivia and Collins, Michael and Parikh, Ankur and Alberti, Chris and Epstein, Danielle and Polosukhin, Illia and Devlin, Jacob and Lee, Kenton and others},
  journal={Transactions of the Association for Computational Linguistics},
  volume={7},
  pages={453--466},
  year={2019},
  publisher={MIT Press One Rogers Street, Cambridge, MA 02142-1209, USA journals-info~…}
}

@INPROCEEDINGS{Vipulanandan2024,
  author={Vipulananthan, Pragatheeswaran and Premaratne, Kamal and Sarkar, Dilip and Murthi, Manohar N.},
  booktitle={2024 IEEE International Conference on Knowledge Graphs (ICKG)}, 
  title={A Quantum Tensor Network-Based Viewpoint for Modeling and Analysis of Time Series Data}, 
  address   = "Abu Dhabi, {UAE}", 
  month     = dec, 
  year={2024},
  volume={},
  number={},
  pages={378-387},
  doi={10.1109/ICKG63256.2024.00054}}

@inproceedings{Singh2024AIJCNN_FindingLocalDependentRegions,
  author		= "R. Singh and Y. Ma and J. C. Principe", 
  title			= "Finding Local Dependent Regions in PDFs using RKHS Uncertainty Moments and Optimal Transport",
  booktitle	= "Proc. International Joint Conference on Neural Networks ({IJCNN})",
  address		= "Yokohama, Japan",
  volume		= "",
  pages			= "",
  month			= jun # "/" # jul,
  year			= "2024", 
  doi			= "10.1109/IJCNN60899.2024.10651130", 
}

@article{Singh2021NC_TowardKernelBasedUncertainty,
  author 	= "R. Singh and J. C. Principe", 
  title 	= "Toward a Kernel-Based Uncertainty Decomposition Framework for Data and Models", 
  journal	= "Neural Computation", 
  volume	= "33", 
  number	= "", 
  pages		= "1164-1198",
  month		= "", 
  year		= "2021",
  doi    = "{10.1162/neco_{a_01372}}",
}

@inproceedings{Singh2020UAI_TimeSeriesAnalysis,
  author    = "R. Singh and J. Principe",
  title     = "Time Series Analysis Using a Kernel Based Multi-Modal Uncertainty Decomposition Framework",
  booktitle = "Conference on Uncertainty in Artificial Intelligence ({UAI})",
  pages     = "1368--1377",
  year      = "2020",
}

@article{rateike2023weakly,
  title={Weakly supervised detection of hallucinations in llm activations},
  author={Rateike, Miriam and Cintas, Celia and Wamburu, John and Akumu, Tanya and Speakman, Skyler},
  journal={arXiv preprint arXiv:2312.02798},
  year={2023}
}

@article{touvron2023llama,
  title={Llama 2: Open foundation and fine-tuned chat models},
  author={Touvron, Hugo and Martin, Louis and Stone, Kevin and Albert, Peter and Almahairi, Amjad and Babaei, Yasmine and Bashlykov, Nikolay and Batra, Soumya and Bhargava, Prajjwal and Bhosale, Shruti and others},
  journal={arXiv preprint arXiv:2307.09288},
  year={2023}
}

@article{grattafiori2024llama,
  title={The llama 3 herd of models},
  author={Grattafiori, Aaron and Dubey, Abhimanyu and Jauhri, Abhinav and Pandey, Abhinav and Kadian, Abhishek and Al-Dahle, Ahmad and Letman, Aiesha and Mathur, Akhil and Schelten, Alan and Vaughan, Alex and others},
  journal={arXiv preprint arXiv:2407.21783},
  year={2024}
}

@article{penedo2023refinedweb,
  title={The RefinedWeb dataset for Falcon LLM: outperforming curated corpora with web data, and web data only},
  author={Penedo, Guilherme and Malartic, Quentin and Hesslow, Daniel and Cojocaru, Ruxandra and Cappelli, Alessandro and Alobeidli, Hamza and Pannier, Baptiste and Almazrouei, Ebtesam and Launay, Julien},
  journal={arXiv preprint arXiv:2306.01116},
  year={2023}
}

@mastersthesis{jiang2024identifying,
  title={Identifying and mitigating vulnerabilities in llm-integrated applications},
  author={Jiang, Fengqing},
  year={2024},
  school={University of Washington}
}

@article{williams2017broad,
  title={A broad-coverage challenge corpus for sentence understanding through inference},
  author={Williams, Adina and Nangia, Nikita and Bowman, Samuel R},
  journal={arXiv preprint arXiv:1704.05426},
  year={2017}
}

@article{yu2014deep,
  title={Deep learning for answer sentence selection},
  author={Yu, Lei and Hermann, Karl Moritz and Blunsom, Phil and Pulman, Stephen},
  journal={arXiv preprint arXiv:1412.1632},
  year={2014}
}

@article{socher2011dynamic,
  title={Dynamic pooling and unfolding recursive autoencoders for paraphrase detection},
  author={Socher, Richard and Huang, Eric and Pennin, Jeffrey and Manning, Christopher D and Ng, Andrew},
  journal={Advances in neural information processing systems},
  volume={24},
  year={2011}
}

@article{he2020realformer,
  title={Realformer: Transformer likes residual attention},
  author={He, Ruining and Ravula, Anirudh and Kanagal, Bhargav and Ainslie, Joshua},
  journal={arXiv preprint arXiv:2012.11747},
  year={2020}
}

@article{tay2021charformer,
  title={Charformer: Fast character transformers via gradient-based subword tokenization},
  author={Tay, Yi and Tran, Vinh Q and Ruder, Sebastian and Gupta, Jai and Chung, Hyung Won and Bahri, Dara and Qin, Zhen and Baumgartner, Simon and Yu, Cong and Metzler, Donald},
  journal={arXiv preprint arXiv:2106.12672},
  year={2021}
}

@article{nikitin2024kernel,
  title={Kernel language entropy: Fine-grained uncertainty quantification for llms from semantic similarities},
  author={Nikitin, Alexander and Kossen, Jannik and Gal, Yarin and Marttinen, Pekka},
  journal={Advances in Neural Information Processing Systems},
  volume={37},
  pages={8901--8929},
  year={2024}
}

@article{nguyen2025beyond,
  title={Beyond Semantic Entropy: Boosting LLM Uncertainty Quantification with Pairwise Semantic Similarity},
  author={Nguyen, Dang and Payani, Ali and Mirzasoleiman, Baharan},
  journal={arXiv preprint arXiv:2506.00245},
  year={2025}
}

@article{qiu2024semantic,
  title={Semantic density: Uncertainty quantification for large language models through confidence measurement in semantic space},
  author={Qiu, Xin and Miikkulainen, Risto},
  journal={Advances in neural information processing systems},
  volume={37},
  pages={134507--134533},
  year={2024}
}

@article{jiang2024graph,
  title={Graph-based Uncertainty Metrics for Long-form Language Model Generations},
  author={Jiang, Mingjian and Ruan, Yangjun and Sattigeri, Prasanna and Roukos, Salim and Hashimoto, Tatsunori B},
  journal={Advances in Neural Information Processing Systems},
  volume={37},
  pages={32980--33006},
  year={2024}
}

@INPROCEEDINGS{haskins2025kea,
  title={KEA Explain: Explanations of Hallucinations using Graph Kernel Analysis},
  author={Haskins, Reilly and Adams, Benjamin},
  booktitle={International Conference on Neurosymbolic Learning and Reasoning.},
  volume={37},
  year={2025}
}

@article{gao2024spuq,
  title={Spuq: Perturbation-based uncertainty quantification for large language models},
  author={Gao, Xiang and Zhang, Jiaxin and Mouatadid, Lalla and Das, Kamalika},
  journal={arXiv preprint arXiv:2403.02509},
  year={2024}
}

@inproceedings{cox2025mapping,
  title={Mapping from Meaning: Addressing the Miscalibration of Prompt-Sensitive Language Models},
  author={Cox, Kyle and Xu, Jiawei and Han, Yikun and Xu, Rong and Li, Tianhao and Hsu, Chi-Yang and Chen, Tianlong and Gerych, Walter and Ding, Ying},
  booktitle={Proceedings of the AAAI Conference on Artificial Intelligence},
  volume={39},
  number={22},
  pages={23696--23703},
  year={2025}
}

@article{kang2025uncertainty,
  title={Uncertainty Quantification for Hallucination Detection in Large Language Models: Foundations, Methodology, and Future Directions},
  author={Kang, Sungmin and Bakman, Yavuz Faruk and Yaldiz, Duygu Nur and Buyukates, Baturalp and Avestimehr, Salman},
  journal={arXiv preprint arXiv:2510.12040},
  year={2025}
}

@article{schweighofer2024information,
  title={On information-theoretic measures of predictive uncertainty},
  author={Schweighofer, Kajetan and Aichberger, Lukas and Ielanskyi, Mykyta and Hochreiter, Sepp},
  journal={arXiv preprint arXiv:2410.10786},
  year={2024}
}

@article{hand2001simple,
  title={A simple generalisation of the area under the ROC curve for multiple class classification problems},
  author={Hand, David J and Till, Robert J},
  journal={Machine learning},
  volume={45},
  number={2},
  pages={171--186},
  year={2001},
  publisher={Springer}
}

@inproceedings{vipulanandan2026sensitivity,
  title        = {Sensitivity and Uncertainty Quantification in Regression Neural Networks: A Quantum Tensor Network Based Approach},
  author       = {Vipulanandan, Pragatheeswaran and Premaratne, Kamal and Sarkar, Dilip},
  booktitle    = {Proceedings of the International Symposium on Artificial Intelligence and Mathematics (ISAIM)},
  year         = {2026},
  address      = {Fort Lauderdale, FL, USA},
}

\vfill
\clearpage
\appendix
\section{Expanded Version of Section~\ref{sec:proposedMethod}}
\label{sec:ExpandedVersionOfSection2}

\subsection{Semantic Equivalence-Based Clustering}

Computation of semantic entropy requires clustering LLM-generated output TSs, such that the TSs in each cluster are \emph{semantically equivalent}, while those in different clusters are \emph{semantically dissonant.} To identify such semantically equivalent TSs, \cite{farquhar2024detecting} employ \emph{bidirectional entailment}, a notion that has attracted considerable attention in natural language inference (NLI). We, use  pretrained models such as DeBERTa-Large-MNLI or general-purpose language models like GPT-3.5 to assess entailment between sentence pairs. When confronted with a new TS, we check for bidirectional entailment with a representative TS from each cluster. If no entailment is found with any of these, a new cluster is created. This process of semantic clustering allows a semantic entropy to be computed for the clusters associated with each question. Higher entropy indicates higher likelihood that the LLM is hallucinating when responding to that \emph{question.} The top rung of Fig.~\ref{fig:semantic_pipeline} summarizes this process in \cite{farquhar2024detecting}. 

However, semantic entropy itself is not a reliable measure for identifying hallucinations because TS probabilities can be highly sensitive to model perturbations. This warrants adjusting the confidence with which hallucinatory behavior is deemed to be present or absent. We therefore need \emph{local} information (as opposed to \emph{global} metrics) that allows us to assess sensitivity of the TS probability of each LLM generation.


\subsection{Quantum TN-Based UQ Method for Deep Learning  Models}
\label{app:UQ_QTNappendix}


Physics-inspired methods offer such local uncertainty information and have emerged as a compelling way for UQ of deep learning models (\cite{Singh2020UAI_TimeSeriesAnalysis, Singh2021NC_TowardKernelBasedUncertainty, Singh2024AIJCNN_FindingLocalDependentRegions}). However, these methods are tethered to the Hamiltonian of the quantum harmonic oscillator (QHO) and are hence severely constrained in their application. In contrast, the recent work in \cite{Vipulanandan2024} introduces a deterministic, interpretable, and single-shot UQ method built on a QTN-based Hamiltonian. Moreover, it leverages the well-established method of perturbation theory to ferret out multi-resolution, local uncertainty estimates that are ideally suited for our current purpose.

\begin{figure}[htbp]
  \centering
  \includegraphics[width = 0.80\textwidth]{%
    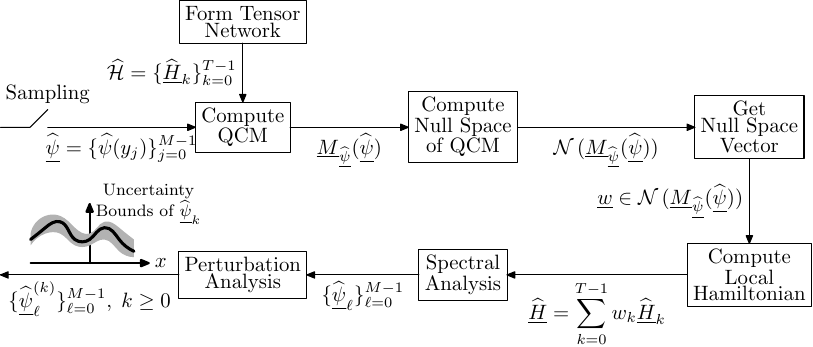}
  \caption{Overview of the QTN-based UQ pipeline used in our work. 
The procedure begins with computing the quantum correlation matrix (QCM) from TS probabilities and extracting its null-space vectors. These are used to construct the local Hamiltonian $\wh{\ul{H}}$, whose eigen-modes provide the foundation for spectral and perturbation analysis. 
First-order perturbation corrections yield uncertainty feature vectors, which are organized into a tensor network representation. Finally, sampling across modes produces uncertainty bounds on probability amplitudes, enabling uncertainty quantification of semantic entropy. }
  \label{fig:pipeline}
\end{figure}

In our work, we expand this QTN-based framework—originally developed for time-series signals—to the domain of semantic datasets, where TS probability distributions can be naturally viewed as signal-like processes. This extension allows us to capture semantic uncertainty in natural language through the same principled, physics-inspired lens. The importance of this contribution is twofold: first, it bridges the gap between temporal signal analysis and semantic modeling, demonstrating that techniques rooted in quantum physics can generalize beyond time-series to complex language domains. Second, by bringing interpretability and multi-resolution uncertainty estimates into the study of semantic entropy, our approach lays the groundwork for more trustworthy and explainable AI systems, especially in high-stakes applications where hallucination detection and careful decision-making are critical. Fig.~\ref{fig:pipeline} shows the QTN-based UQ method that we leverage in our work.

\subsection{Rationale for Adopting Semantic R{\'e}nyi Entropy}
\label{app:shannon_vs_SRE}

While SE is widely used for characterizing uncertainty in LLM outputs, its inherent properties introduce limitations in the present setting. For any fixed predictive distribution, SE is always greater than or equal to quadratic SRE (\cite{Harremoes2009Kybernetika_JointRangeRenyiEntropies}), but this pointwise ordering does not in general determine the relative spread of the two measures across a corpus of examples. Empirically, however, SE exhibits a larger variation in most cases (between hallucinatory and non-hallucinatory behavior) in our experiments, which correlates with slightly better AUROC performance of SE (see Fig.~\ref{fig:winrate_AUROC}) (\cite{hand2001simple}). SE also emphasizes rarer outcomes. But the tail probabilities do not play a role in AURAC (see Fig.~\ref{fig:winrate_AURAC}) which measures how well increasing entropy corresponds to decreasing correctness.

A more decisive consideration arises from the structural compatibility between SRE and the proposed UQ framework. Quadratic Rényi entropy is a second-order functional of the underlying probability distribution and is directly estimated by the KME. This correspondence enables the KME to be encoded as an eigen-mode of a QTN Hamiltonian. Perturbations of this Hamiltonian, in turn, yield first-order and higher-order spectral corrections that quantify local sensitivity in the amplitude domain. Such a perturbation-based formulation of uncertainty cannot be obtained from SE because it lacks a corresponding RKHS-based representation.

Although SE achieves marginally better separability in AUROC due to its wider dynamic range, the empirical performance gap relative to SRE is small. In contrast, the methodological advantages facilitated by SRE—namely, its direct alignment with RKHS theory, its compatibility with QTN-based spectral analysis, and its ability to support principled perturbation-based uncertainty quantification—are substantially more consequential. For these reasons, SRE constitutes the most appropriate entropy measure for the proposed framework.


\subsection{R{\'e}nyi Entropy and Kernel Mean Embedding (KME) of Data PDF}


To provide the essentials of the work in \cite{Vipulanandan2024}, take the R{\'e}nyi (quadratic) entropy $H(X) = -\log\,(\psi(X))$ of the continuous random variable $X$ with PDF $p(x)$ \cite{Principe2010_Book}. Here, $\psi(X) = \int p^2(x)\, dx = \mbb{E}\, [p(x)]$, where $\mbb{E}[\cd]$ is the expectation operator. With $N$ samples $\{x_i\}_{i = 1}^N$ from $X$, the non-parametric Parzen Gaussian density estimator of the PDF $p(x)$ is $\wh{p}(x) = \sum_{i = 1}^N \kappa_{\sigma}(x; x_i)/N$, where $\kappa_{\sigma}(x; x_i) = (1/\sqrt{2\pi \sigma^2})\, \exp\,(-(x - x_i)^2/2\sigma^2)$ \cite{Parzen1962AMS}. This turns out to be the (empirical) kernel mean embedding (KME) $\wh{\psi}(x)$ of the data PDF in the reproducing kernel Hilbert space (RKHS) determined by $\kappa_{\sigma}(\cd; \cd)$ (\cite{Aronszajn1950ToAMS, Scholkopf2015SC_RandomVariablesViaRKHS}), i.e., $\wh{p}(x) = \wh{\psi}(x)$. $H(X) = \displaystyle -\log\, (\psi(X))$ being an entropy measure, the scalar $\wh{\psi}(X)$ is taken as a measure of uncertainty and the (empirical) KME $\wh{\psi}(x)$ viewed as how it varies with $x$, the data amplitude values. 

To get a more interpretable tool for UQ, \cite{Vipulanandan2024} seeks a finite QTN having local nearest neighbor random couplings whose Hamiltonian has this KME $\wh{\psi}(x)$ as \emph{one} of its eigen-modes. Suffice it to say that 
    \tb{(a)}~a spin chain having $L$ spin particles can represent a sampled version $\wh{\ul{\psi}} = \{\wh{\psi}_j\}_{j = 0}^{M - 1}$, where $M = 2^L$, of the KME $\wh{\psi}(x)$; 
    \tb{(b)}~the Hamiltonian of the TN takes the form of a linear combination of a finite set $\wh{\mc{H}} = \{\wh{\ul{H}}_k\}_{k = 0}^{T - 1}$ of pairwise orthonormal (w.r.t. the Hilbert-Schmidt inner product) Hermitian operators, where the extent of local nearest neighbor influence determines $T$; and 
    \tb{(c)}~$T$, in turn, determines whether the set $\wh{\mc{H}}$ is `rich' enough for a Hamiltonian of the type $\wh{\ul{H}} = \sum_{k = 0}^{T - 1} w_k \wh{\ul{H}}_k$ (i.e., $\wh{\ul{H}} \in \mr{span}\, (\wh{\mc{H}})$) to have $\wh{\ul{\psi}}$ as an eigen-mode.
    
To identify such a Hamiltonian, the quantum correlation matrix (QCM) $\ul{M}_{\wh{\ul{\psi}}}(\wh{\mc{H}})$ (\cite{Qi2019Quantum_DeterminingLocalHamiltonian}) because $\wh{\ul{\psi}}$ is an eigen-mode of $\wh{\ul{H}}$ iff $\ul{w}$ belongs in the null space of the QCM. Here, $\ul{w} = \{w_k\}_{k = 0}^{T - 1}$ is the linear combination weight vector that generates $\wh{\ul{H}} = \sum_{k = 0}^{T - 1} w_k \wh{\ul{H}}_k$. Additional theoretical results and supporting lemmas related to the QCM formulation are provided in Appendix~\ref{supp:QCM_supplimentary}.

With this Hamiltonian in hand, the well-established method of perturbation theory is applied to the associated time-independent Schr{\"o}dinger equation. This yields first- and higher order `corrections' to \emph{all} the eigen-modes/energies (including the KME) when the underlying Hamiltonian undergoes a perturbation (\cite{CohenTannoudji1977_Book}). The higher order modes provide better discriminative resolution of those data regions for which less information is available (these correspond to the tails of the PDF) (\cite{Singh2021NC_TowardKernelBasedUncertainty}); sensitivity analysis and UQ can be carried out via the corrections to the eigen-modes. This is the UQ method in \cite{Vipulanandan2024} that allows one to survey aleatoric uncertainties locally and at different levels of resolution using algebraic and spectral properties of QTNs.


\subsection{Some Properties of the QCM}
\label{supp:QCM_supplimentary}
\label{supp:Lemmas_and_corollaries}


In this section, we present a series of results related to the QCM that form the theoretical backbone of our work, discussed in Section~\ref{sec:Preliminaries}.

Note that $\wh{\mc{H}} = \{\wh{\ul{H}}_i\}$ denotes a finite set $\{\wh{\ul{H}}_i,\; i = 0, \ldots, T-1\}$ of pairwise orthonormal Hermitian operators. For convenience, we will use $(\wh{\mc{H}})_{\ul{w}}$ to denote a (real-valued) linear combination of the operators in $\mc{\wh{H}}$, i.e., 
\begin{equation}
  (\mc{\wh{H}})_{\ul{w}}
    = \sum_{i = 0}^{T-1}
      w_i \wh{\ul{H}}_i,
  \tx{ where }
  \ul{w}
    = \{w_i\}_{i = 0}^{T-1}
    \in \mbb{R}^T.
\end{equation}

\begin{lemma}
\label{lem:M_realsymmetric}
The QCM $\QCM{H}{v} = \{(\QCM{H}{v})_{ij}\}$ associated with $\mc{\wh{H}}$ is real and symmetric. 
\end{lemma}

\emph{Proof.} 
From Definition, we have
\begin{align*}
  (\QCM{H}{v})_{ji}
    &= 0.5\, 
       \la \{\wh{\ul{H}}_j, \wh{\ul{H}}_i\} \ra_{\ul{v}} 
         - \la \wh{\ul{H}}_j \ra_{\ul{v}}
           \cdot
           \la \wh{\ul{H}}_i \ra_{\ul{v}} \\
    &= 0.5\, 
       \la \{\wh{\ul{H}}_i, \wh{\ul{H}}_j\} \ra_{\ul{v}} 
         - \la \wh{\ul{H}}_i \ra_{\ul{v}}
           \cdot
           \la \wh{\ul{H}}_j \ra_{\ul{v}}
     = (\QCM{H}{v})_{ij},
\end{align*}
because $\{\wh{\ul{H}}_i, \wh{\ul{H}}_j\} = \{\wh{\ul{H}}_j, \wh{\ul{H}}_i\}$ and both $\la \wh{\ul{H}}_i \ra_{\ul{v}}$ and $\la \wh{\ul{H}}_j \ra_{\ul{v}}$ are scalars. So, $\ul{M}$ is symmetric.

In addition, 
\begin{align*}
  (\QCM{H}{v})_{ij}^*
    &= 0.5\, \la \{\wh{\ul{H}}_i, \wh{\ul{H}}_j\} \ra_{\ul{v}}^* 
         - \la \wh{\ul{H}}_i \ra_{\ul{v}}^*
           \cdot
           \la \wh{\ul{H}}_j \ra_{\ul{v}}^* \\
    &= 0.5\, \la \ul{v} \mid \{\wh{\ul{H}}_i, \wh{\ul{H}}_j\} \mid \ul{v} \ra^* 
         - \la \ul{v} \mid \wh{\ul{H}}_i \mid \ul{v} \ra^*
           \cdot
           \la \ul{v} \mid \wh{\ul{H}}_j \mid \ul{v} \ra^* \\
    &= 0.5\, \la \ul{v} \mid \wh{\ul{H}}_i \wh{\ul{H}}_j \mid \ul{v} \ra^* 
         + 0.5\, \la \ul{v} \mid \wh{\ul{H}}_j \wh{\ul{H}}_i \mid \ul{v} \ra^* 
         - \la \ul{v} \mid \wh{\ul{H}}_i \mid \ul{v} \ra^*
           \cdot
           \la \ul{v} \mid \wh{\ul{H}}_j \mid \ul{v} \ra^* 
\end{align*}
But we note that 
\begin{alignat*}{4}
  &\la \ul{v} \mid \wh{\ul{H}}_i \wh{\ul{H}}_j \mid \ul{v} \ra^*
    &
      &= \la \wh{\ul{H}}_i \ul{v} \mid \wh{\ul{H}}_j \ul{v} \ra^*
        &
          &= \la \wh{\ul{H}}_j \ul{v} \mid \wh{\ul{H}}_i \ul{v} \ra 
            &
              &= \la \ul{v} \mid \wh{\ul{H}}_j \wh{\ul{H}}_i \mid \ul{v} \ra; \\
  &\la \ul{v} \mid \wh{\ul{H}}_i \mid \ul{v} \ra^*
    &
      &= \la \ul{v} \mid \wh{\ul{H}}_i \ul{v} \ra^*
        &
          &= \la \wh{\ul{H}}_i \ul{v} \mid \ul{v} \ra 
            &
              &= \la \ul{v} \mid \wh{\ul{H}}_i \mid \ul{v} \ra.
\end{alignat*}
In a similar manner, we can also show that $\la \ul{v} \mid \wh{\ul{H}}_j \wh{\ul{H}}_i \mid \ul{v} \ra^* = \la \ul{v} \mid \wh{\ul{H}}_i \wh{\ul{H}}_j \mid \ul{v} \ra^*$ and $\la \ul{v} \mid \wh{\ul{H}}_j \mid \ul{v} \ra^* = \la \ul{v} \mid \wh{\ul{H}}_j \mid \ul{v} \ra$. When substituted into the expression above for $(\QCM{H}{v})_{ij}^*$, we see that $(\QCM{H}{v})_{ij}^* = (\QCM{H}{v})_{ij}$. So, $\ul{M}$ is real.
\QEDclosed

\begin{lemma}
\label{lem:Var_positive}
The variance $\mr{Var}\, (\wh{\ul{H}})_{\ul{\psi}} = \la \wh{\ul{H}}^2 \ra_{\ul{\psi}} - |\la \wh{\ul{H}} \ra_{\ul{\psi}}|^2$ of the Hermitian operator $\wh{\ul{H}}$ w.r.t. an arbitrary normalized vector $\ul{\psi},\; \la \ul{\psi} \mid \ul{\psi} \ra = 1$, satisfies
\[
  \mr{Var}\, (\wh{\ul{H}})_{\ul{\psi}}
    = \la \wh{\ul{H}}^2 \ra_{\ul{\psi}} - |\la \wh{\ul{H}} \ra_{\ul{\psi}}|^2 
    \geq 0;
\]
the equality holds iff $\ul{\psi}$ is an eigen-mode of $\wh{\ul{H}}$.
\end{lemma}

\emph{Proof.}
For arbitrary $\alpha \in \mbb{C}$ and $\ul{\psi}$, we know that
\[
  0
    \leq
      \la \wh{\ul{H}}\, \ul{\psi} - \alpha\, \ul{\psi},\, \wh{\ul{H}}\, \ul{\psi} - \alpha\, \ul{\psi} \ra.
\]
Furthermore, the equality holds true iff $\wh{\ul{H}}\, \ul{\psi} - \alpha\, \ul{\psi} = 0$, i.e., iff $\ul{\psi}$ is an eigen-mode of $\wh{\ul{H}}$. 

So, let us consider the case when $\ul{\psi}$ is not an eigen-mode of $\wh{\ul{H}}$ so that 
\begin{align*}
  0
    &< \la \wh{\ul{H}}\, \ul{\psi} - \alpha\, \ul{\psi},\, \wh{\ul{H}}\, \ul{\psi} - \alpha\, \ul{\psi} \ra 
     = \la \wh{\ul{H}}\, \ul{\psi} \mid \wh{\ul{H}}\, \ul{\psi} \ra - \la \wh{\ul{H}}\, \ul{\psi} \mid \alpha\, \ul{\psi} \ra - \la \alpha\, \ul{\psi} \mid \wh{\ul{H}}\, \ul{\psi} \ra 
         + \la \alpha\, \ul{\psi} \mid \alpha\, \ul{\psi} \ra \\
    &= \la \wh{\ul{H}}\, \ul{\psi} \mid \wh{\ul{H}}\, \ul{\psi} \ra - \alpha\, \la \wh{\ul{H}}\, \ul{\psi}  \mid \ul{\psi} \ra - \alpha^* \la \ul{\psi} \mid \wh{\ul{H}}\, \ul{\psi} \ra 
         + |\alpha|^2\, \la \ul{\psi} \mid \ul{\psi} \ra.
\end{align*}
Since $\alpha$ is arbitrary, select
\[ 
  \alpha
    = \la \ul{\psi} \mid \wh{\ul{H}}\, \ul{\psi} \ra
  \implies
  \alpha^*
    = \la \wh{\ul{H}}\, \ul{\psi} \mid \ul{\psi} \ra.
\]
Substitute into the above inequality:
\begin{align*}
  0
    &< \la \ul{\psi} \mid \wh{\ul{H}}^2 \mid \ul{\psi} \ra - |\la \ul{\psi} \mid \wh{\ul{H}}\, \ul{\psi} \ra|^2 - |\la \ul{\psi} \mid \wh{\ul{H}}\, \ul{\psi} \ra|^2 
         + |\la \ul{\psi} \mid \wh{\ul{H}}\, \ul{\psi} \ra|^2 \\
    &= \la \ul{\psi} \mid \wh{\ul{H}}^2 \mid \ul{\psi} \ra - |\la \ul{\psi} \mid \wh{\ul{H}} \mid \ul{\psi} \ra|^2
     = \la \wh{\ul{H}}^2 \ra_{\ul{\psi}} - |\la \wh{\ul{H}} \ra_{\ul{\psi}}|^2.
\end{align*}
This establishes the claim.
\QEDclosed

The following result follows immediately:

\begin{corollary}
\label{cor:H_eigen-mode}
Given the Hermitian operator $\wh{\ul{H}}$ and the normalized vector $\ul{\psi},\; \la \ul{\psi} \mid \ul{\psi} \ra = 1$, $\mr{Var}\, (\wh{\ul{H}})_{\ul{\psi}} = 0$ iff $\ul{\psi}$ is an eigen-mode of $\wh{\ul{H}}$.
\end{corollary}

\begin{corollary}
\label{cor:M_NullSpace}
Suppose, for some arbitrary real-valued vector $\ul{w} \in \{w_i\} \in \mbb{R}^T$, $\wh{\ul{H}} = (\mc{\wh{H}})_{\ul{w}}$. Then the following are true:
\begin{itemize}
  \item[(i)] For arbitrary $\ul{v} \in \mbb{R}^T$,  $\QCM{H}{v}$ is p.s.d. 
  \item[(ii)] $\ul{w} \in \NN{\QCM{H}{v}}$, the null space of the QCM $\QCM{H}{v}$ associated with $\mc{\wh{H}}$ w.r.t $\ul{v}$ iff $\ul{v},\; \la \ul{v} \mid \ul{v} \ra = 1$, is an eigen-mode of $\wh{\ul{H}}$. 
\end{itemize}
\end{corollary}

\emph{Proof.} 
\begin{itemize}
  \item[(i)] This follows directly from Lemma~\ref{lem:Var_positive} and \cite{Qi2019Quantum_DeterminingLocalHamiltonian}.
  \item[(ii)] First, suppose $\ul{w} \in \NN{\QCM{H}{v}}$ so that 
\[
  \QCM{H}{v}\, \ul{w} 
    = 0
  \implies
  \la \QCM{H}{v} \ra_{\ul{w}}
    = \ul{w}^T \QCM{H}{v}\, \ul{w}
    = 0.
\]
Then, $\mr{Var}\, (\wh{\ul{H}})_{\ul{v}} = \la \QCM{H}{v} \ra_{\ul{w}} = 0$ (\cite{Qi2019Quantum_DeterminingLocalHamiltonian}). Corollary~\ref{cor:H_eigen-mode} then implies that $\ul{v}$ is an eigen-mode of $\wh{\ul{H}}$.

Conversely, suppose $\la \QCM{H}{v} \ra_{\ul{w}} = \ul{w}^T \QCM{H}{v}\, \ul{w} = 0$. We know from Lemma~\ref{lem:M_realsymmetric} that $\QCM{H}{v}$ is real and symmetric. Let its SVD be
\[
  \QCM{H}{v}
    = \ul{U}\, \ul{\Sigma}\, \ul{U}^T,
\]
where $\ul{U} \in \mbb{R}^{T \times T}$ is unitary and $\ul{\Sigma} \in \mbb{R}^{T \times T}$ is diagonal with its diagonal entries being the non-negative singular values of $\QCM{H}{v}$. Let $\QCM{H}{v}^{1/2} = \ul{\Sigma}^{1/2} \ul{U}^T$. Then,  
\[
  \ul{w}^T \QCM{H}{v}\, \ul{w}
    = \ul{w}^T \QCM{H}{v}^{{1/2}^T}\, \QCM{H}{v}^{1/2} \ul{w} 
    = \la \QCM{H}{v}^{1/2} \ul{w}, \QCM{H}{v}^{1/2} \ul{w} \ra
    = 0,
\]
meaning that we must have 
\[
  \QCM{H}{v}^{1/2} \ul{w} 
    = 0 
  \implies 
  \QCM{H}{v}\, \ul{w} 
    = 0,
\]
i.e., $\ul{w} \in \NN{\QCM{H}{v}}$.
\QEDclosed
\end{itemize}

\subsection{On the Null Space of the QCM: The Zero Dimension Case}


We now study a special case where the QCM has full rank (zero-dimensional null space) and investigate how small perturbations can introduce a one-dimensional null space. Understanding this transition is important because the emergence of a nontrivial null space corresponds to identifying new uncertainty directions, which are critical for robust decision-making in our framework.

Consider the case in the QCM, when the dimension of the null space of the QCM $\QCM{H}{\psi^{\sharp}}$ is zero and we `perturb' $\QCM{H}{\psi^{\sharp}}$ to $\QCM{H'}{\psi^{\sharp}}$ as
\begin{equation}
  \QCM{H'}{\psi^{\sharp}}
    = \QCM{H}{\psi^{\sharp}} - \delta\ul{M}
\end{equation} 
s.t. $\mr{dim}\, (\NN{\QCM{H'}{\psi^{\sharp}}}) = 1$. 

Henceforth in this section, for convenience, we will denote $\QCM{H}{\psi^{\sharp}}$ by $\ul{M}$ and $\QCM{H'}{\psi^{\sharp}}$ by $\ul{M}'$.

Recall that $\{\ul{w}_n, \mu_n\},\; n \in 0, 1, \ldots, T - 1$, denote the eigen-pairs of $\ul{M}$ ordered as $0 < \mu_0 \leq \mu_1 \leq \cdots \leq \mu_{T - 1}$. 

We next investigate the stability of the QCM under small perturbations. Specifically, we examine how the eigen-modes/energies of the QCM change when a small perturbation is introduced, particularly in the context where the original QCM has full rank (no null space) and perturbations induce a null direction. We use classical matrix perturbation theory to derive first-order approximations and cosine similarity results.

Let $\{\ul{w}'_n,\, \mu'_n\},\; n \in 0, 1, \ldots, T - 1$, denote the eigen-pairs of $\ul{M}'$ ordered as $0 = \mu'_0 < \mu'_1 \leq \cdots \leq \mu'_{T - 1}$. Let us now view the eigen-pairs $\{\ul{w}_n,\, \lambda_n\}$ of $\ul{M}$ as the `perturbed' versions of the eigen-pairs $\{\ul{w}'_n,\, \lambda'_n\}$ of $\ul{M}'$, so that we may apply perturbation theory to $\ul{M}'$. For this purpose, as is customary in perturbation theory, let
\begin{equation}
  \ul{M}
    = \ul{M}' + \epsilon \cdot \delta\ul{M},
\end{equation}
where $\epsilon > 0$ is simply a place-holder. The eigen-pairs of $\ul{M}$ can then be expressed as 
\begin{equation}
  \{\ul{w}_n,\, \lambda_n\}
    = \left\{
        \sum_{k = 0}^{\infty}
        \epsilon^k \ul{w}_n^{(k)},\,
        \sum_{\ell = 0}^{\infty} 
        \epsilon^{\ell} \lambda_n^{(\ell)}
      \right\},
\end{equation}
where $\{w_n^{(0)},\, \lambda_n^{(0)}\} \overset{\Delta}{=} \{w_n',\, \lambda_n'\},\; \forall n \geq 0$. W.l.o.g. we also assume that $\la \ul{w}_n^{(0)} \mid \ul{w}_n^{(k)} \ra = 0,\; \forall\, k \geq 1,\; \forall\, n \geq 0$.

Now we can express the eigen-pair relationship 
\begin{equation}
  \ul{M} \mid \ul{w}_n \ra
    = \lambda_n \mid \ul{w}_n \ra
\end{equation}
as
\begin{equation}
  (\ul{M}' + \epsilon \cdot \delta\ul{M}) \mid
    \sum_{k = 0}^{\infty}
    \epsilon^k \ul{w}_n^{(k)} 
  \ra
    = \sum_{\ell = 0}^{\infty} 
      \epsilon^{\ell} \lambda_n^{(\ell)}
      \mid
      \sum_{k = 0}^{\infty}
      \epsilon^k \ul{w}_n^{(k)} \ra,
\end{equation}
or equivalently, 
\begin{equation}
  \ul{M}' 
  \sum_{k = 0}^{\infty}
  \epsilon^k \mid \ul{w}_n^{(k)} \ra
    + \delta\ul{M}
      \sum_{k = 1}^{\infty}
      \epsilon^k \mid \ul{w}_n^{(k)} \ra 
    = \sum_{k = 0}^{\infty} \epsilon^k
      \sum_{\ell = 0}^k
      \lambda_n^{(k - \ell)} \mid \ul{w}_n^{(\ell)} \ra.
\end{equation}
Gather similar powers of $\epsilon$: for $k \geq 1$, 
\begin{equation}
  (\ul{M}' - \lambda_n^{(0)}) \mid \ul{w}_n^{(k)} \ra
    = \sum_{\ell = 0}^{k - 1}
      \lambda_n^{(k - \ell)} \mid \ul{w}_n^{(\ell)} \ra
        - \delta\ul{M} \mid \ul{w}_n^{(k - 1)} \ra.
\end{equation}
Note that, with $k = 0$, we get $(\ul{M}' - \lambda_n^{(0)}) \mid \ul{w}_n^{(0)} \ra = 0$, which is of course trivially true.

Push the bra $\la \ul{w}_m^{(0)} \mid$: for $k \geq 1$, 
\begin{equation}
  (\lambda_m^{(0)} - \lambda_n^{(0)}) \la \ul{w}_m^{(0)} \mid \ul{w}_n^{(k)} \ra 
    = \sum_{\ell = 0}^{k - 1}
      \lambda_n^{(k - \ell)} \la \ul{w}_m^{(0)} \mid \ul{w}_n^{(\ell)} \ra
        - \la \ul{w}_m^{(0)} \mid \delta\ul{M} \mid \ul{w}_n^{(k - 1)} \ra.
  \label{eq:Pert1}
\end{equation}
Put $m = n$ in \eqref{eq:Pert1} to get the order-$k$ correction for $\lambda_n^{(0)}$:
\begin{equation}
  \lambda_n^{(k)}
    = \la \ul{w}_n^{(0)} \mid \delta\ul{M} \mid \ul{w}_n^{(k - 1)} \ra,\; 
      k \geq 1.
\end{equation}
For $m \neq n$, put $k = 1$ in \eqref{eq:Pert1} to get
\begin{equation}
  \la \ul{w}_m^{(0)} \mid \ul{w}_n^{(1)} \ra
    = \frac{\la \ul{w}_m^{(0)} \mid \delta\ul{M} \mid \ul{w}_n^{(0)} \ra}{\lambda_n^{(0)} - \lambda_m^{(0)}},
\end{equation}
where we assume that $\ul{M}^{(0)}$ is non-degenerate. Since $\{\ul{w}_n^{(0)}\}$ forms an ONB, we can get the order-$1$ correction for $\ul{w}_n^{(0)}$ as
\begin{equation}
  \mid \ul{w}_n^{(1)} \ra
    = \sum_{m = 0}^{\infty} 
      \mid \ul{w}_m^{(0)} \ra 
      \cdot
      \la \ul{w}_m^{(0)} \mid \ul{w}_n^{(1)} \ra
    = \mathop{\sum_{m = 0}^{\infty}}_{m \neq n}
      \mid \ul{w}_m^{(0)} \ra 
      \cdot
      \frac{\la \ul{w}_m^{(0)} \mid \delta\ul{M} \mid \ul{w}_n^{(0)} \ra}{\lambda_n^{(0)} - \lambda_m^{(0)}}.
\end{equation}

Notice that, since $\la \ul{w}_n^{(0)} \mid \ul{w}_n^{(1)} \ra = 0$, we have 
\begin{equation}
  \Vert \ul{w}_n^{(0)} + \ul{w}_n^{(1)} \Vert^2
    = \la \ul{w}_n^{(0)} \mid \ul{w}_n^{(0)} \ra + \la \ul{w}_n^{(1)} \mid \ul{w}_n^{(1)} \ra
    = 1 + \Vert \ul{w}_n^{(1)} \Vert^2,
\end{equation}
because $\ul{w}_n^{(0)}$ is already normalized. Therefore, denoting by $\wh{\ul{w}}_n^{(0)}$ the normalized order-$1$ corrected $\ul{w}_n^{(0)}$, we have  
\begin{equation}
  \wh{\ul{w}}_n^{(0)}
    = \frac{\ul{w}_n^{(0)} + \ul{w}_n^{(1)}}{(1 + \Vert \ul{w}_n^{(1)} \Vert^2)^{1/2}},
\end{equation}
where 
\begin{align}
  \Vert \ul{w}_n^{(1)} \Vert^2
    &= \la \ul{w}_n^{(1)} \mid \ul{w}_n^{(1)} \ra 
       \notag \\   
    &= \sum_{\ell \neq n} \sum_{m \neq n}
       \frac{\la \ul{w}_m^{(0)} \mid \delta\ul{M} \mid \ul{w}_n^{(0)} \ra}{\lambda_n^{(0)} - \lambda_m^{(0)}}
       \cdot
       \frac{\la \ul{w}_{\ell}^{(0)} \mid \delta\ul{M} \mid \ul{w}_n^{(0)} \ra}{\lambda_n^{(0)} - \lambda_{\ell}^{(0)}} 
       \cdot
       \la \ul{w}_{\ell}^{(0)} | \ul{w}_m^{(0)} \ra
       \notag \\
    &= \sum_{m \neq n}
       \left(
         \frac{\la \ul{w}_m^{(0)} \mid \delta\ul{M} \mid \ul{w}_n^{(0)} \ra}{\lambda_n^{(0)} - \lambda_m^{(0)}}
       \right)^2.
\end{align}
Thus, 
\begin{align}
  \la \ul{w}_n^{(0)} \mid \wh{\ul{w}}_n^{(0)} \ra
    &= \frac{\la \ul{w}_n^{(0)} \mid \ul{w}_n^{(0)} \ra + \la \ul{w}_n^{(0)} \mid \ul{w}_n^{(1)} \ra}{(1 + \Vert \ul{w}_n^{(1)} \Vert^2)^{1/2}}
       \notag \\
    &= \frac{1}{(1 + \Vert \ul{w}_n^{(1)} \Vert^2)^{1/2}}
     \geq 
       1 - \frac{1}{2}\, \Vert \ul{w}_n^{(1)} \Vert^2,
\end{align}
where we used the Taylor series expansion
\begin{equation}
  \frac{1}{(1 + x)^{1/2}}
    = 1 - \frac{1}{2}\, x + \frac{1}{2!}\, \frac{3}{4}\, (1 + c)^{-5/2}\, x^2,\;
      x > 0,
\end{equation}
for some $c \in (0, x)$, which in turn implies that 
\begin{equation}
  \frac{1}{(1 + x)^{1/2}}
    \geq 1 - \frac{1}{2}\, x,
\end{equation}
because $(1 + c)^{-5/2} > 0,\; c \in (0, x)$. 

In summary, we have
\begin{equation}
  \la \ul{w}_n^{(0)} \mid \wh{\ul{w}}_n^{(0)} \ra
    \geq 
      1 
        - \frac{1}{2} 
          \sum_{m \neq n}
          \left(
            \frac{\la \ul{w}_m^{(0)} \mid \delta\ul{M} \mid \ul{w}_n^{(0)} \ra}{\lambda_n^{(0)} - \lambda_m^{(0)}}
          \right)^2.
\end{equation}
We know that $\{\ul{w}_0^{(0)},\, \lambda_0^{(0)}\} = \{\ul{w}_0',\, \lambda_0' = 0\}$ is the `smallest' eigen-pair of $\ul{M}'$. So, putting $n = 0$, we have 
\begin{equation}
  \la \ul{w}_0^{(0)} \mid \wh{\ul{w}}_0^{(0)} \ra
    \geq 
      1 
        - \frac{1}{2} 
          \sum_{m \neq 0}
          \left(
            \frac{\la \ul{w}_m^{(0)} \mid \delta\ul{M} \mid \ul{w}_0^{(0)} \ra}{\lambda_m^{(0)}}
          \right)^2.
\end{equation}

To conclude, this section shows that introducing a small perturbation to a full-rank QCM systematically creates a well-defined null space, corresponding to an emergent low-variance direction in the QTN. The perturbation analysis, grounded in classical matrix perturbation theory, quantifies how close the perturbed eigen-mode remains to the original eigen-structure through a first-order approximation. This understanding is critical in practice, as it ensures that small modeling errors or noise do not destabilize the uncertainty quantification pipeline and that the emergence of low-uncertainty directions remains theoretically sound and controllable.


\section{Prompt Templates}
\label{sec:Templates}


\paragraph{Phrase length Model Prompting and Answer Selection} 
The prompt template used for all datasets to generate LLM responses is
\vspace*{-0.1in}
\begin{quote}
\emph{Answer the following question as briefly as possible:} \{question\} \\
\emph{Answer:}
\end{quote}
\vspace*{-0.1in}

\paragraph{Sentence length Model Prompting and Answer Selection} 
The prompt template used for all datasets to generate LLM responses is
\vspace*{-0.1in}
\begin{quote}
\emph{Answer the following question in a single brief but complete sentence:} \{question\} \\
\emph{Answer:}
\end{quote}
\vspace*{-0.1in}


\paragraph{Entailment Estimation/Semantic Clustering} 
To detect entailment we utilized the following prompt template:
\vspace*{-0.1in}
\begin{quote}
\emph{We are evaluating answers to the question:} \{question\}

\emph{Here are two possible answers:} \\
Possible Answer 1: \{text1\} \\
Possible Answer 2: \{text2\}

\emph{Does Possible Answer 1 semantically entail Possible Answer 2?} \\
\emph{Respond with:} entailment, contradiction, or neutral.
\end{quote}


\section{Extended Evaluation of Hallucination Detection Across Diverse Settings}
\label{supp:further_details}


Due to page limitations, we reported the results summary in the main paper. In this section, we present in depth experimental results to validate that  our confabulation detection framework works across diverse datasets. Building on the example analysis in Section~\ref{sec:results}, we extend the evaluation to TriviaQA, NQ, SVAMP, and SQuAD across multiple quantization levels, thereby encompassing a broader spectrum of reasoning complexities and linguistic variations. In addition, we break down results for both phrase-level and sentence-level generations, via prompts discussed in Section \ref{sec:Templates}) to examine the impact of output length on detection robustness. For each dataset, we consistently apply the eight LLM models (Mistral-7B-v0.1, Mistral-7B-instruct-v0.3, Falcon-rw-1b, LLaMA-3.2-1b, LLaMA-2-13b-chat, LLaMA-2-7b-chat, LLaMA-2-13b, and LLaMA-2-7b) and maintain the original entailment-based clustering and uncertainty quantification pipelines without modification.  This controlled experimental design ensures that any observed performance trends can be attributed to dataset characteristics rather than methodological changes. The results demonstrate that our framework generalizes effectively across heterogeneous tasks, reinforcing its applicability to varied real-world deployment scenarios.


\subsection{AUROC Performance Across Various Precision Models and Datasets}
\label{supp:AUROC_all}

To more rigorously validate the robustness of our confabulation detection framework, we extend the evaluation to sentence-level and phrase-level generations under different quantization precision (16-bit, 8-bit, and 4-bit). This setup allows us to examine how model compression impacts uncertainty estimation and confabulation detection performance, when sentence length answers as well as short answers are generated. We consistently apply the set of LLMs and datasets as in the main paper, thereby ensuring that observed variations are attributable solely to quantization effects rather than changes in methodology. Importantly, quantization is a necessary step for deploying large models in resource-constrained environments, yet hallucination behavior under quantized models has been largely overlooked in prior work. By explicitly studying confabulation detection in this setting and validating our approach against the SOTA methods discussed in Section~\ref{sec:experimental_setup}, we address a critical gap in the literature and highlight the practical importance of uncertainty-aware methods for LLMs.

\subsubsection{Sentence length output performance.} 
\label{sec:AUROC_sentence}

\begin{figure}[htpb]
  \centering
  \includegraphics[width = 0.80\columnwidth, keepaspectratio]{%
    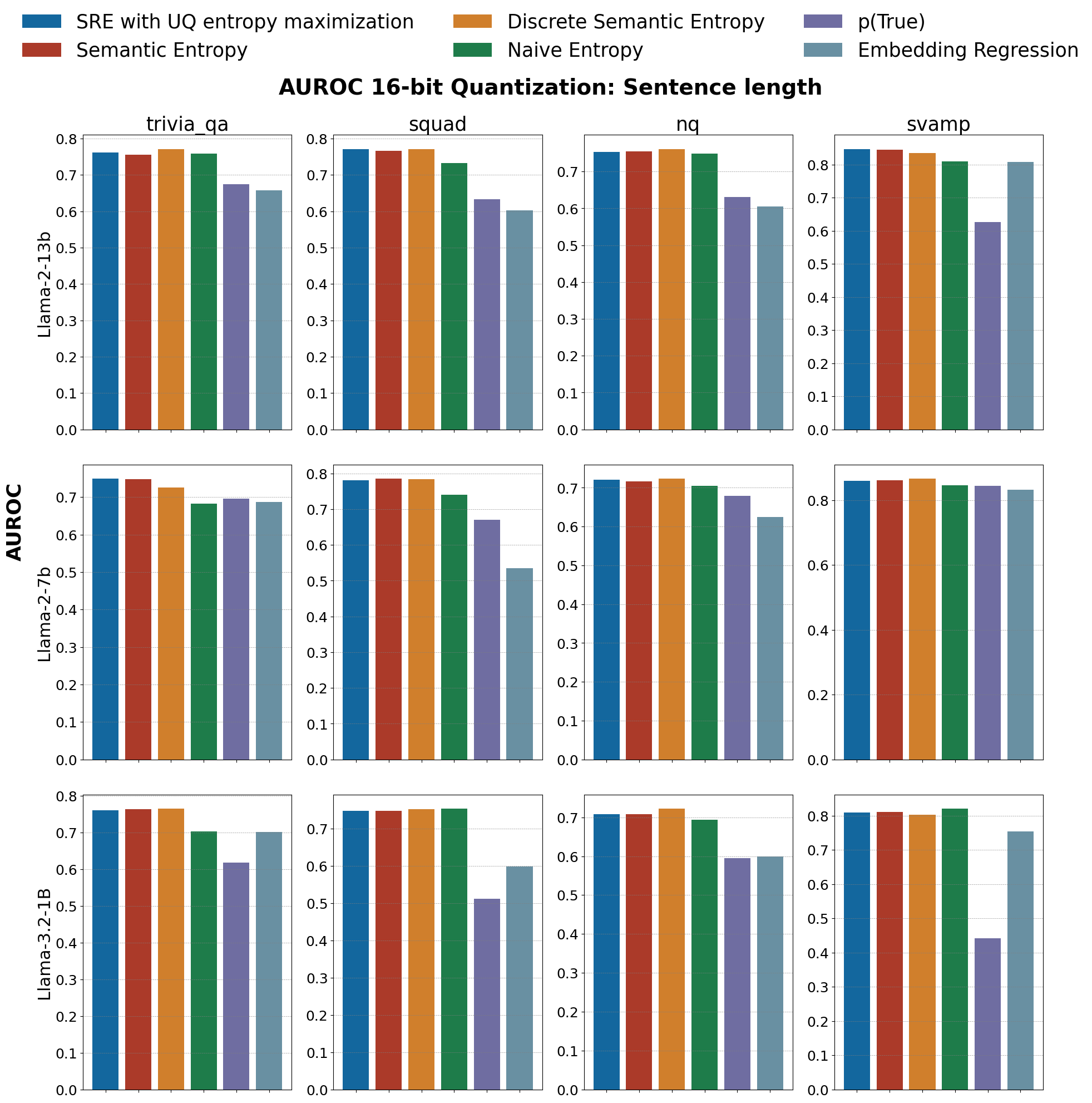}
  \caption{%
  \tb{summarizing 12 experimental} scenarios, AUROC scores for confabulation detection across three LLMs (LLaMA 2 13B, LLaMA 2 7B, LLaMA 3.2 1B) at 16 bit precision and four datasets (TriviaQA, SQuAD, NQ, SVAMP) for sentence length output. The performance of the proposed SRE with UQ is in par or even higher than SOTA methods.}
  \label{fig:auroc_16bit_sentence}
\end{figure}

\begin{figure}[htpb]
  \centering
  \includegraphics[width = 0.80\columnwidth, keepaspectratio]{%
    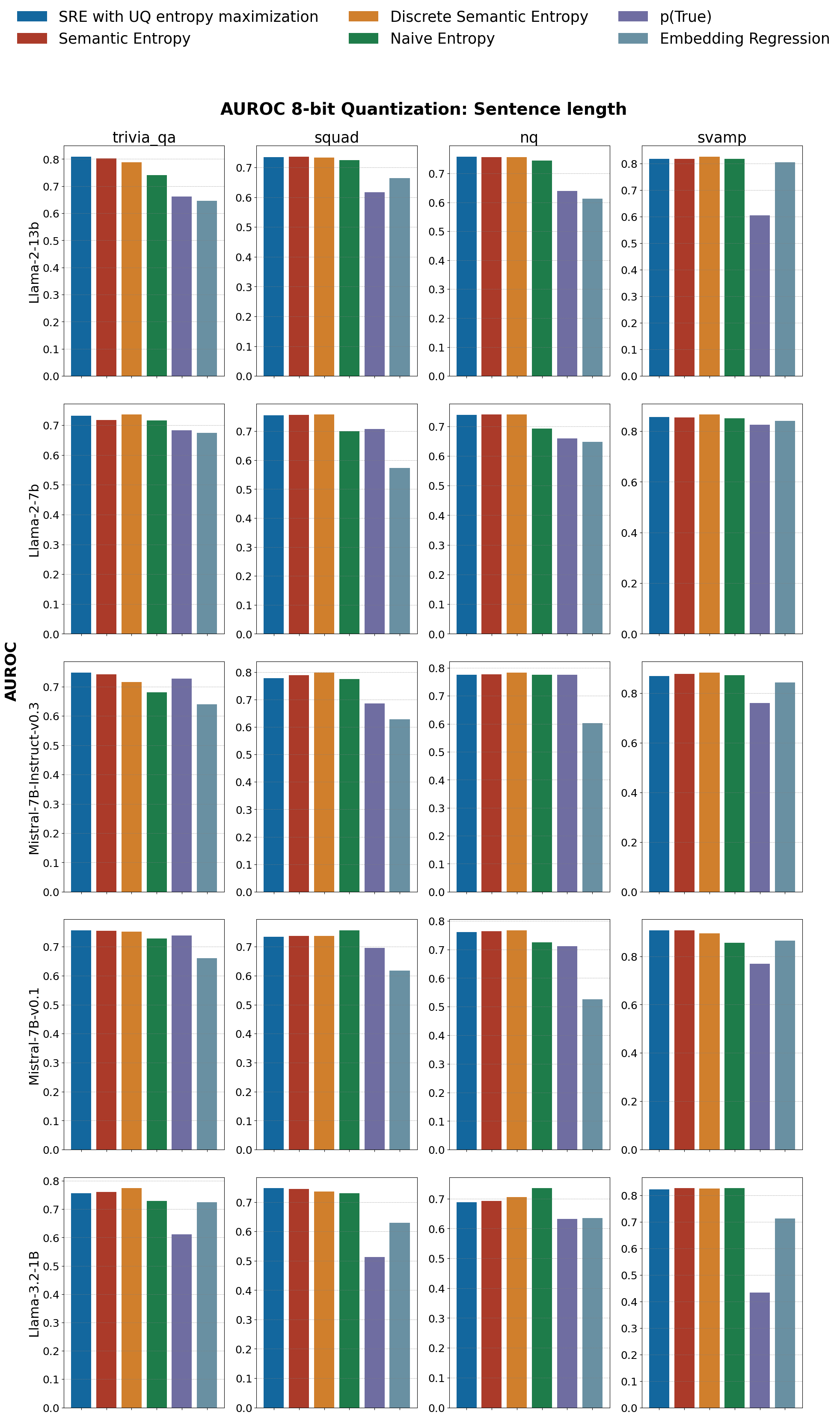}
  \caption{%
  \tb{summarizing 20 experimental} scenarios, AUROC scores for confabulation detection across five LLMs (LLaMA 2 13B, LLaMA 2 7B, LLaMA 3.2 1B, Mistral-7B, Mistral-7B-chat) at 8 bit precision and four datasets (TriviaQA, SQuAD, NQ, SVAMP) for sentence length output. The performance of the proposed SRE with UQ is in par or even higher than SOTA methods.}
  \label{fig:auroc_8bit_sentence}
\end{figure}

\begin{figure}[htpb]
  \centering
  \includegraphics[width = 0.80\columnwidth, keepaspectratio]{%
    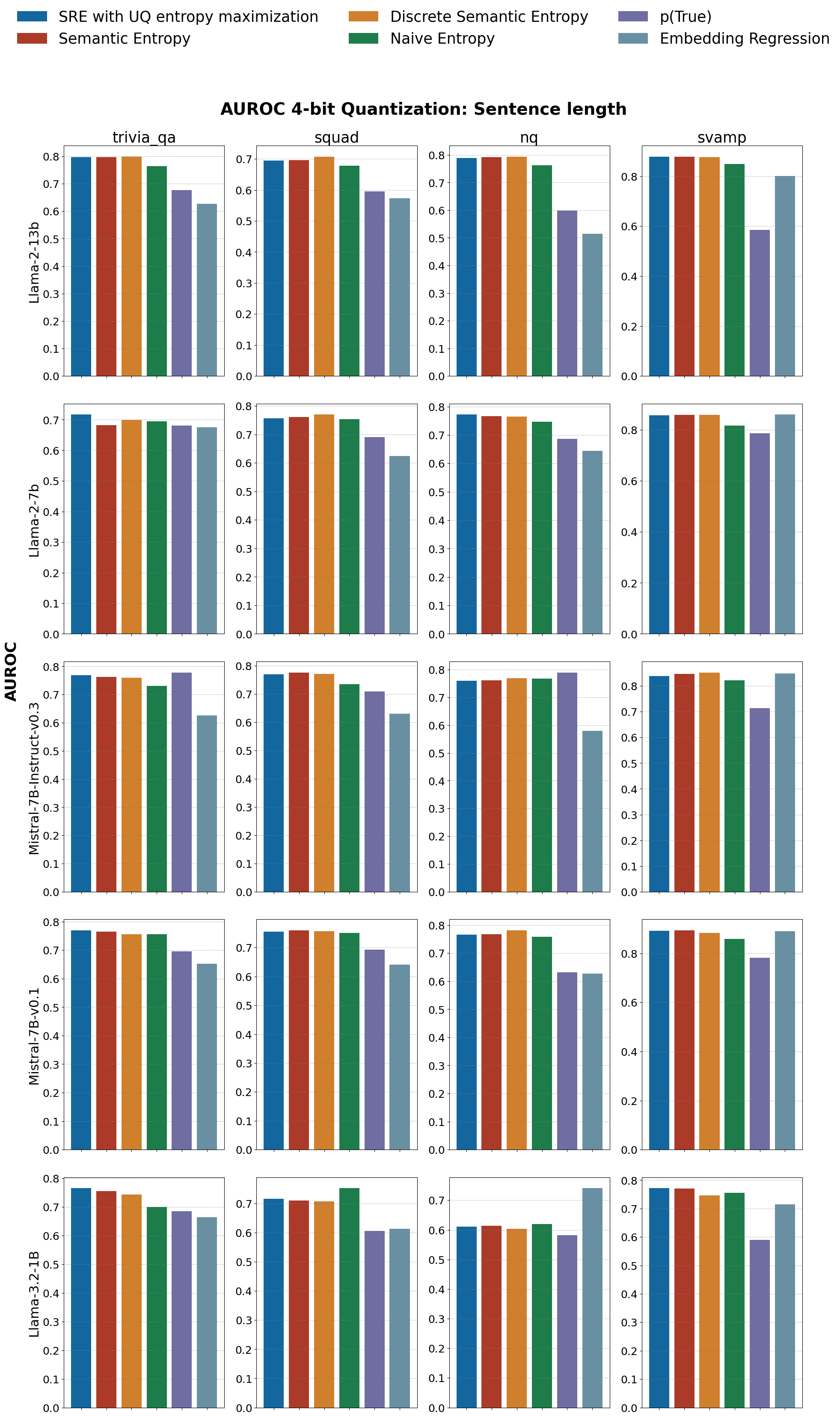}
  \caption{%
  \tb{summarizing 20 experimental} scenarios, AUROC scores for confabulation detection across five LLMs (LLaMA 2 13B, LLaMA 2 7B, LLaMA 3.2 1B, Mistral-7B, Mistral-7B-chat) at 4 bit precision and four datasets (TriviaQA, SQuAD, NQ, SVAMP) for sentence length output. The performance of the proposed SRE with UQ is in par or even higher than SOTA methods.}
  \label{fig:auroc_4bit_sentence}
\end{figure}

Figs.~\ref{fig:auroc_16bit_sentence}, \ref{fig:auroc_8bit_sentence}, and \ref{fig:auroc_4bit_sentence} illustrate the AUROC performance at 16-bit, 8-bit, and 4-bit precision, respectively, for six different hallucination quantification methods across various LLMs and four datasets, using sentence-length outputs. In total, we conducted \tb{52 experiments} spanning all quantization levels, providing a comprehensive basis for sentence length analysis. The experimental results reveal several notable trends and insights:

\vspace{-2mm}
\begin{enumerate}\setlength{\itemsep}{0pt}\setlength{\parskip}{0pt}
\item Methods based on semantic information—such as SRE with UQ, SE, and DSE—consistently outperform baselines based on raw token sequence probabilities ($p(\text{True})$) and embedding regression across all datasets. These results confirm effectiveness of semantic diversity rather than raw output probabilities in detecting confabulations.
\item Across all quantization levels, our framework maintains competitive AUROC performance relative to baseline methods. Importantly, we observe that reductions in model precision (e.g., moving from 16-bit to 4-bit quantization) do not significantly degrade the reliability of our uncertainty-based detection.
\item When model precision is reduced, SOTA methods such as SE, NE have performance degradation, especially in TriviaQA and SQuAD datasets. This highlights the resilience of our approach to model compression, making it suitable for deployment in resource-constrained environments where lightweight LLMs are required.
\item Smaller models such as Llama 3.2-1B show slightly reduced AUROC compared to larger models like LLama-13B, revealing that model size and internal representational power are significant factors in uncertainty calibration.
\item We also observe a minor performance drop for the NQ dataset when using the Llama 3.1B model, which warrants further investigation in future work.
\end{enumerate}

\FloatBarrier 

\subsubsection{Phrase length output performance on instruct models.} 

In this section, we extend the analysis to phrase-level generations, focusing on instruction-tuned LLMs, such as  under multiple quantization precisions (16-bit, 8-bit, and 4-bit). Phrase-level evaluation captures shorter, context-sensitive completions, which are particularly relevant for instruction-following scenarios where answers are often concise. By applying the same uncertainty quantification pipeline across datasets and models, we ensure methodological consistency, allowing observed performance variations to be attributed to quantization and model adaptation effects. While quantization has become standard for enabling deployment of instruction-tuned LLMs in practical applications, its influence on hallucination behavior remains unexplored. Our experiments directly addresses this gap by examining confabulation detection under compressed, instruction-following settings.

\begin{figure}[htpb]
  \centering
  \includegraphics[width = 0.8\columnwidth, keepaspectratio]{%
    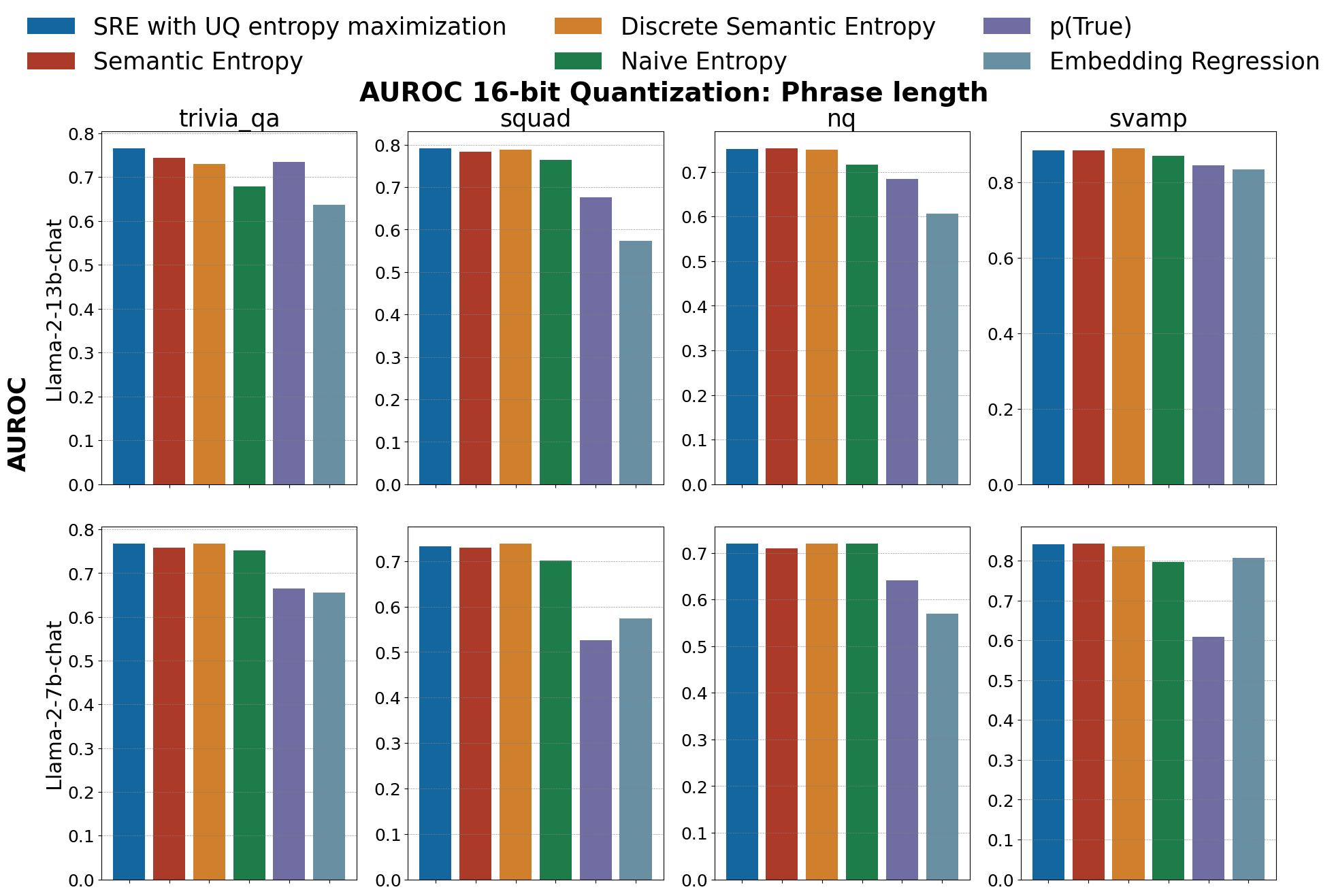}
  \caption{%
  \tb{summarizing 8 experimental} scenarios, AUROC scores for confabulation detection across two LLMs (LLaMA 2 7B chat, LLaMA 2 13B chat) at 16 bit precision and four datasets (TriviaQA, SQuAD, NQ, SVAMP). The performance of the proposed SRE with UQ is in par or even higher than SOTA methods.}
  \label{fig:auroc_16bit_chat}
\end{figure}

\begin{figure}[htpb]
  \centering
  \includegraphics[width = 0.8\columnwidth, keepaspectratio]{%
    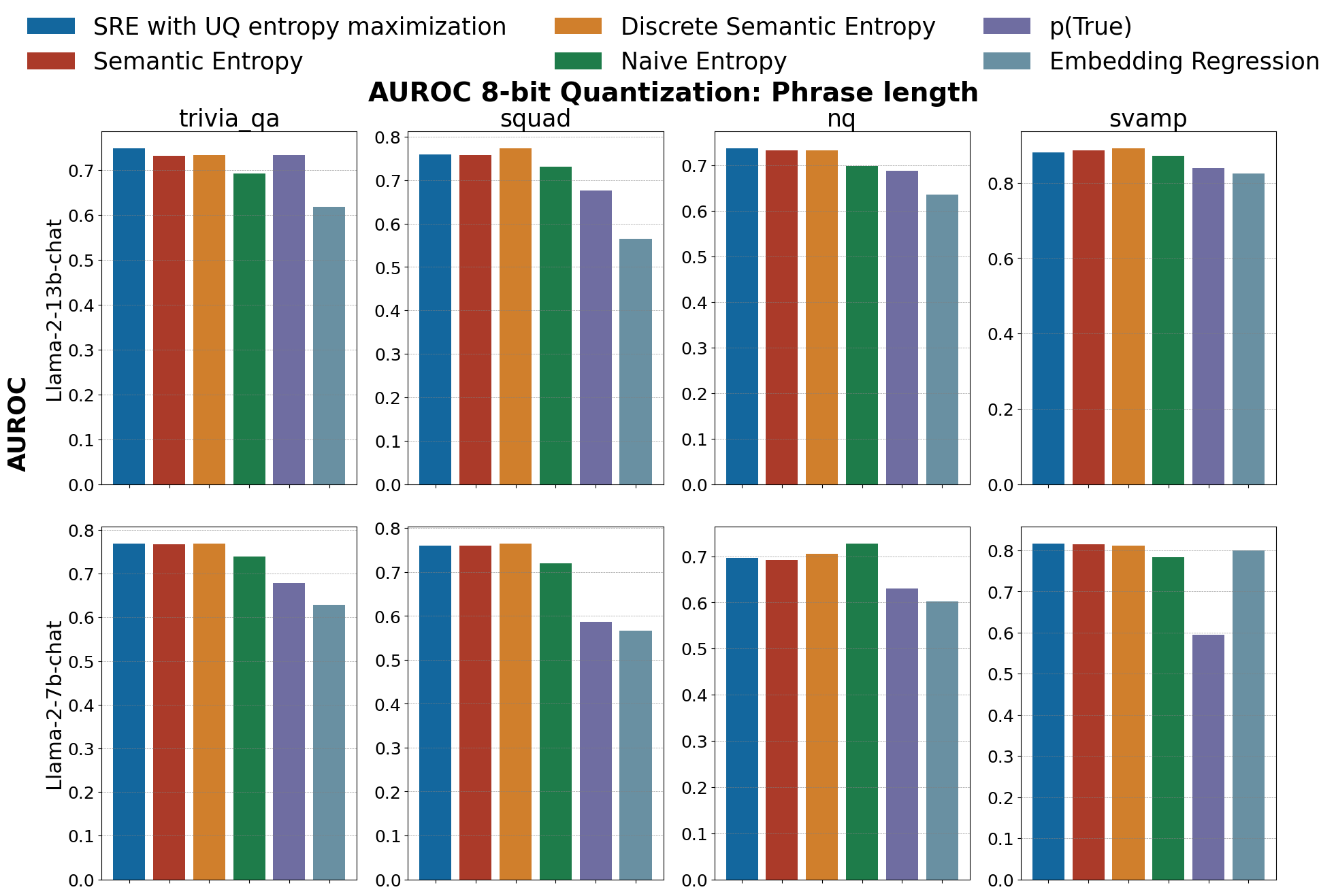}
  \caption{%
  \tb{summarizing 8 experimental} scenarios, AUROC scores for confabulation detection across four LLMs (Mistral-7B, Falcon-1B, LLaMA 3.2B, LLaMA 2 7B 4-bit) and four datasets (TriviaQA, SQuAD, NQ, SVAMP). The performance of the proposed semantic R{\'e}nyi entropy is in par or even higher than SOTA methods.}
  \label{fig:auroc_8bit_chat}
\end{figure}

\begin{figure}[htpb]
  \centering
  \includegraphics[width = 0.8\columnwidth, keepaspectratio]{%
    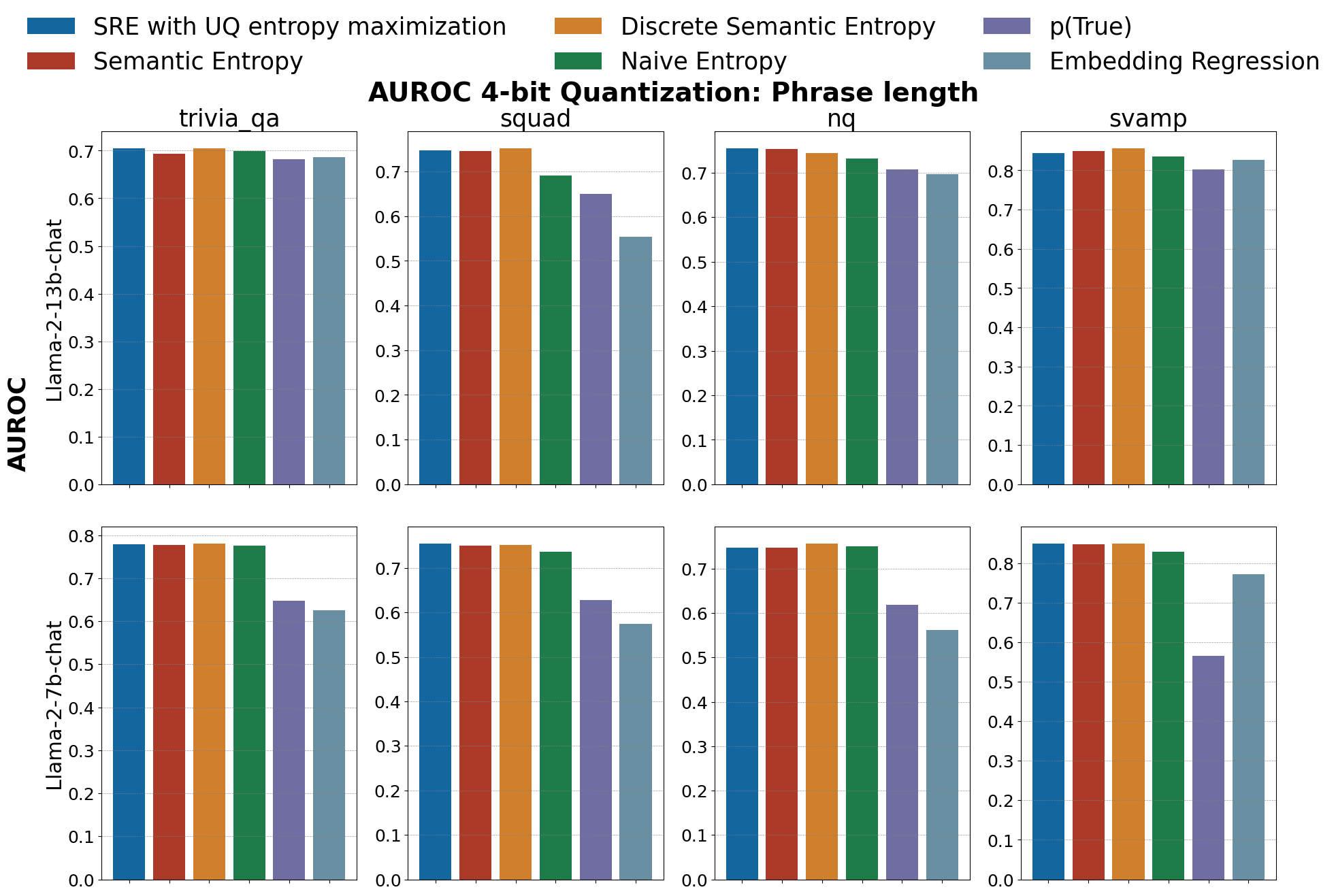}
  \caption{%
  \tb{summarizing 8 experimental} scenarios, AUROC scores for confabulation detection across four LLMs (Mistral-7B, Falcon-1B, LLaMA 3.2B, LLaMA 2 7B 4-bit) and four datasets (TriviaQA, SQuAD, NQ, SVAMP). The performance of the proposed semantic R{\'e}nyi entropy is in par or even higher than SOTA methods.}
  \label{fig:auroc_4bit_chat}
\end{figure}

Figs.~\ref{fig:auroc_16bit_chat}, \ref{fig:auroc_8bit_chat}, and \ref{fig:auroc_4bit_chat} present the AUROC scores for six hallucination quantification approaches across diverse datasets and instruction-tuned LLMs, under 16-bit, 8-bit, and 4-bit precision. Across all quantization levels, we performed a total of \tb{24 experiments}, providing a robust foundation for phrase-level evaluation on instruction fine tuned models. The results reveal several consistent patterns and distinctive insights:

\vspace{-2mm}
\begin{enumerate}\setlength{\itemsep}{0pt}\setlength{\parskip}{0pt}
\item Semantic-based criteria such as SRE with UQ maximization, SE, and DSE—consistently outperform probability-based ($p(\text{True})$) and embedding-based baselines, indicating that semantic variability is a reliable uncertainty signal even for shorter responses.
\item Across all quantization levels, our framework maintains competitive AUROC performance relative to baseline methods. Importantly, we observe that reductions in model precision (e.g., moving from 16-bit to 4-bit quantization) do not significantly degrade the reliability of our uncertainty-based detection
\item By contrast, prior SOTA baselines such as SE and NE show larger declines under reduced precision, particularly in more challenging datasets (e.g., NQ, SQuAD), underscoring the sensitivity of token-level entropy measures to compression.
\item Instruction-tuned models provide better calibrated AUROC overall compared to their base versions, though a consistent gap remains between the smaller LLaMA-2-7B-chat and the larger LLaMA-2-13B-chat, highlighting the role of model scale in uncertainty reliability.
\end{enumerate}

\FloatBarrier

\vspace{-2mm}

\subsubsection{Phrase length output performance on non instruct models.} 

In this section, we analyze phrase-level generations from non-instruction-tuned LLMs under varying quantization levels (16-bit, 8-bit, and 4-bit). Phrase-level completions in base models are especially prone to drifting and incoherent continuations because, unlike instruction-tuned models, they lack explicit task alignment. This makes them a critical yet underexplored regime for studying hallucination behavior—particularly when combined with quantization, which is essential for deploying large models in constrained environments. Prior SOTA methods have largely overlooked this intersection, focusing either on sentence-level or instruction-following setups, leaving open questions about uncertainty reliability in compressed base models. Our study directly addresses this gap by evaluating uncertainty-aware hallucination detection in non-instruction-tuned, phrase-level settings.

\begin{figure}[htpb]
  \centering
  \includegraphics[width = 0.90\columnwidth, keepaspectratio]{%
    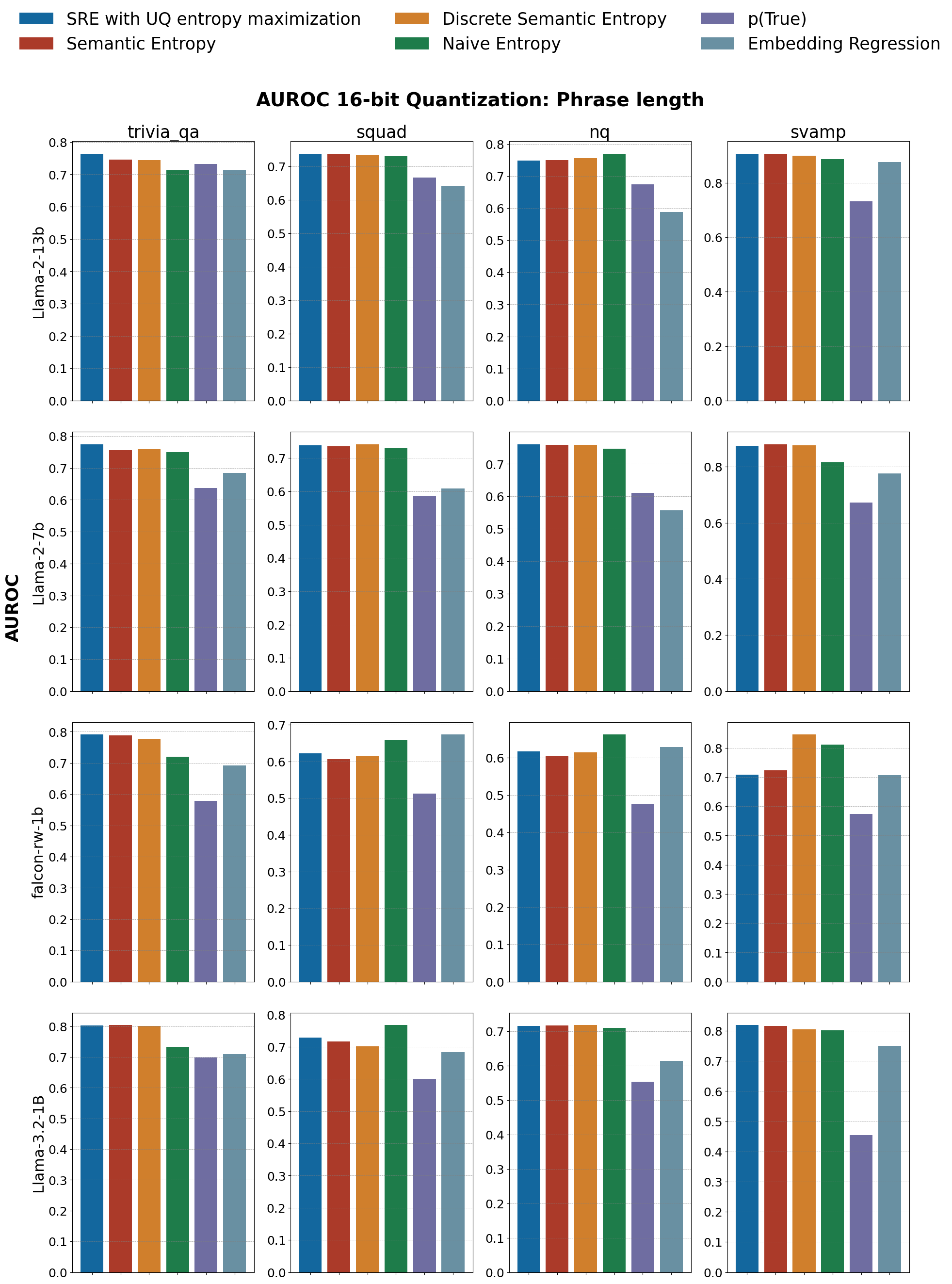}
  \caption{%
  \tb{summarizing 16 experimental} scenarios, AUROC scores for confabulation detection across four LLMs (Falcon-1B, LLaMA 3.2B, LLaMA 2 7B, LLaMA 2 13B) and four datasets (TriviaQA, SQuAD, NQ, SVAMP). The performance of the proposed SRE with UQ is in par or even higher than SOTA methods.}
  \label{fig:auroc_16bit_non_instruct}
\end{figure}

\begin{figure}[htpb]
  \centering
  \includegraphics[width = 0.90\columnwidth, keepaspectratio]{%
    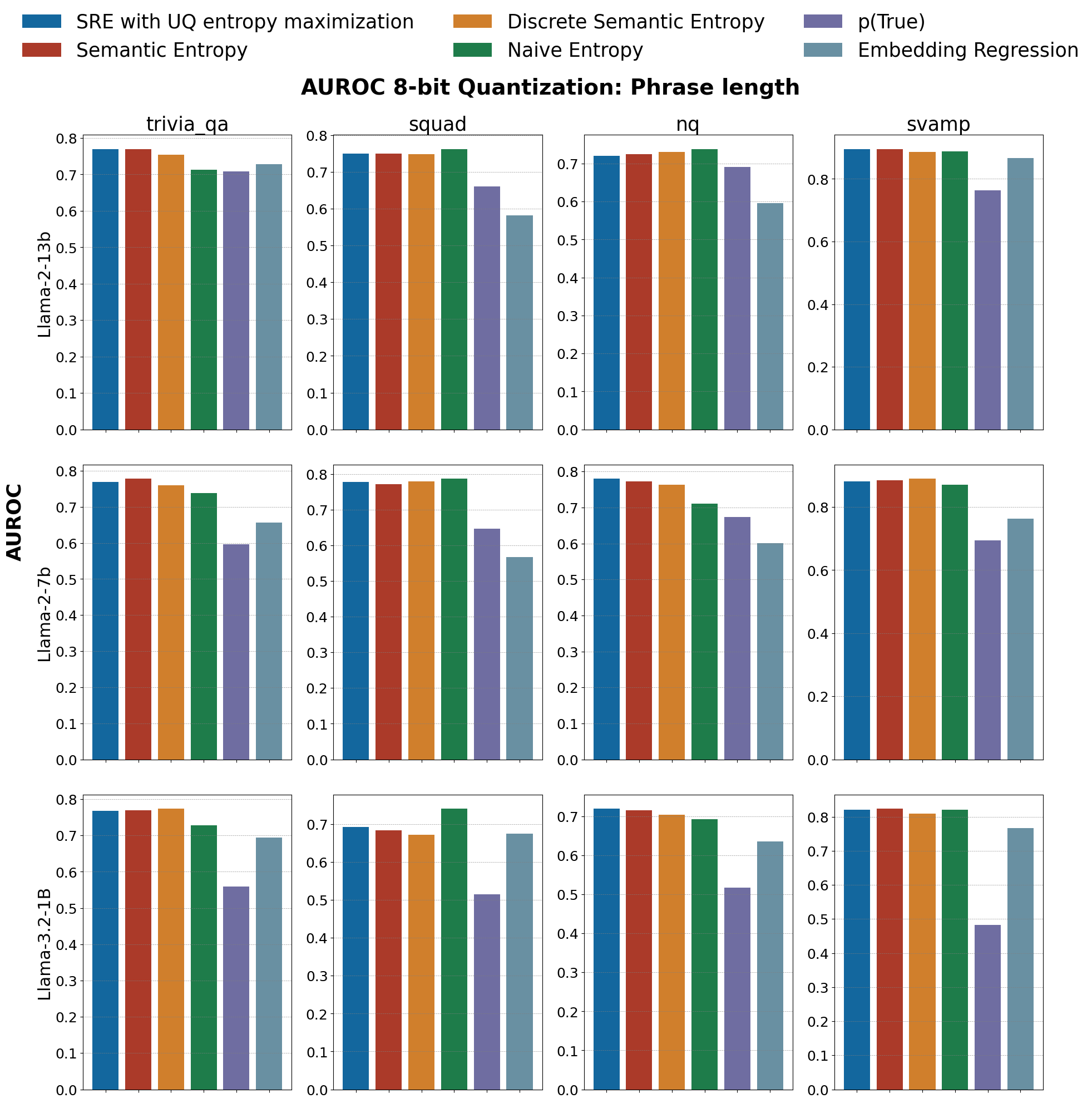}
  \caption{%
  \tb{summarizing 12 experimental} scenarios, AUROC scores for confabulation detection across three LLMs (LLaMA 3.2B, LLaMA 2 7B, LLaMA 2 13B) and four datasets (TriviaQA, SQuAD, NQ, SVAMP). The performance of the proposed SRE with UQ is in par or even higher than SOTA methods.}
  \label{fig:auroc_8bit_non_instruct}
\end{figure}

\begin{figure}[htpb]
  \centering
  \includegraphics[width = 0.90\columnwidth, keepaspectratio]{%
    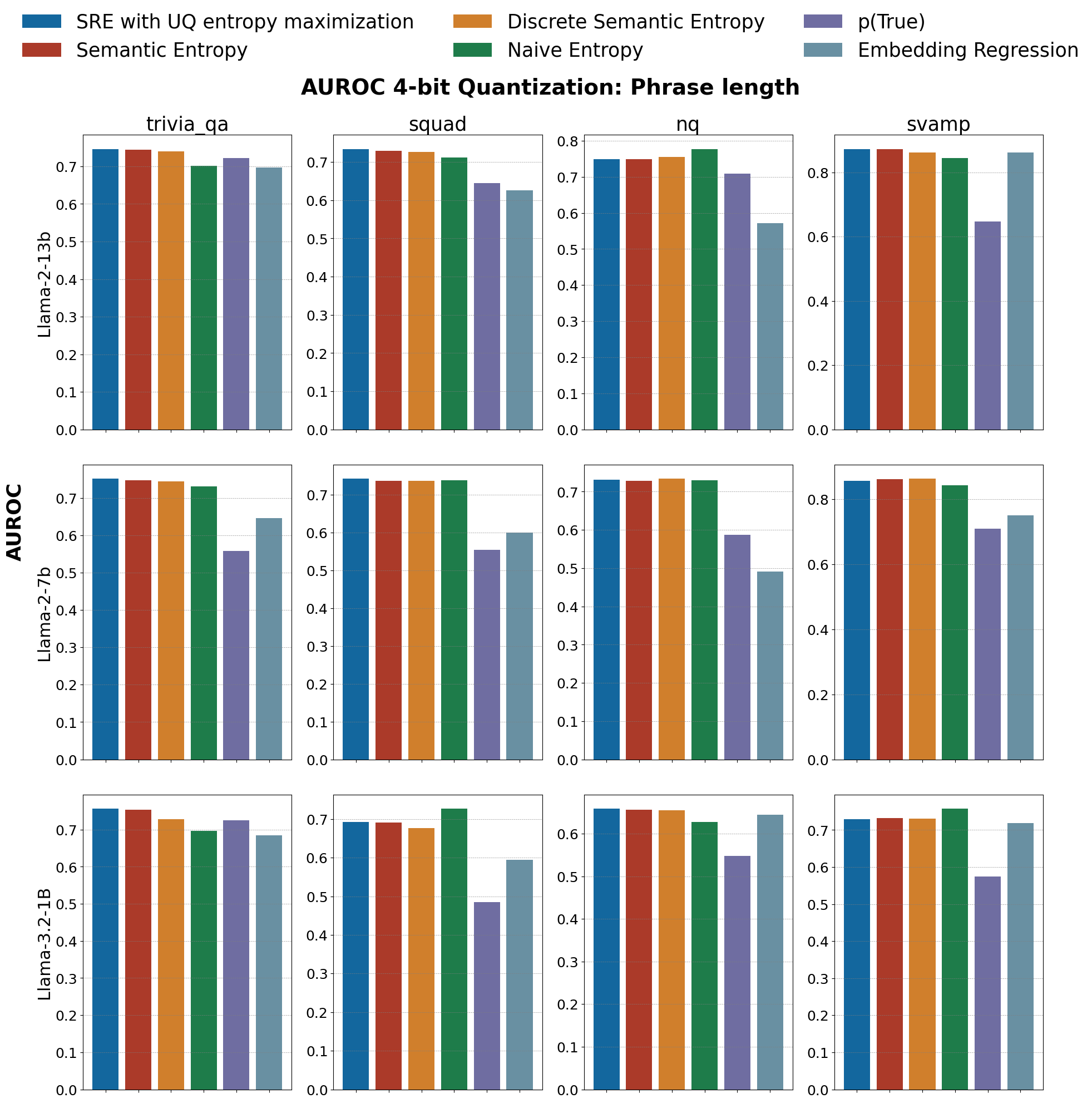}
  \caption{%
  \tb{summarizing 12 experimental} scenarios, AUROC scores for confabulation detection across three LLMs (LLaMA 3.2B, LLaMA 2 7B, LLaMA 2 13B) and four datasets (TriviaQA, SQuAD, NQ, SVAMP). The performance of the proposed SRE with UQ is in par or even higher than SOTA methods.}
  \label{fig:auroc_4bit_non_instruct}
\end{figure}

Figs.~\ref{fig:auroc_16bit_non_instruct}, \ref{fig:auroc_8bit_non_instruct}, and \ref{fig:auroc_4bit_non_instruct} summarize AUROC performance across six uncertainty quantification methods, four datasets, and multiple base LLMs under 16-bit, 8-bit, and 4-bit quantization. In total, we conducted \tb{40 experiments}, enabling a systematic comparison of quantization and uncertainty estimation in this overlooked regime. The results reveal the following consistent findings:

\vspace{-2mm}
\begin{enumerate}\setlength{\itemsep}{0pt}\setlength{\parskip}{0pt}
\item Semantic-based methods dominate: Across datasets and precisions, SRE with UQ, SE, and DSE consistently outperform $p(\text{True})$ and embedding regression, confirming that semantic diversity remains a strong uncertainty signal even in base models with less task alignment.
\item Across all quantization levels, our framework maintains competitive AUROC performance relative to baseline methods. Importantly, we observe that reductions in model precision (e.g., moving from 16-bit to 4-bit quantization) do not significantly degrade the reliability of our uncertainty-based detection
\item Token-level entropy baselines (NE) suffer performance drops—most evident in NQ and SQuAD—highlighting their fragility under quantization and emphasizing the robustness of semantic-based signals.
\item Larger base models (e.g., LLaMA-2-13B) consistently achieve higher AUROC than smaller ones (e.g., LLaMA-3.2-1B, Falcon-RW-1B). This indicates that representational richness is particularly crucial for phrase-level uncertainty calibration in non-instruct settings.
\end{enumerate}


\subsection{RAC Performance Across various precision models and datasets}
\label{supp:AURAC_all}

To more rigorously evaluate the robustness of our confabulation detection framework, we complement the AUROC analysis (Appendix~\ref{supp:AUROC_all}) with RAC and their area-under-curve summary, AURAC, across multiple quantization precisions (16-bit, 8-bit, and 4-bit). While AUROC measures separability under idealized conditions, RAC and AURAC provide a stricter evaluation by quantifying how reliably uncertainty estimates improve accuracy under progressive rejection. Given that quantization is now standard for deploying LLMs in resource-constrained environments, understanding its impact on rejection behavior remains a critical but largely overlooked dimension. By systematically analyzing RACs and AURAC under quantization, we disentangle the effects of compression and model scale, ensuring methodological consistency via identical clustering-based uncertainty pipelines. In this setting, steeper RAC slopes indicate stronger capacity to prioritize reliable generations while filtering confabulations.

Expanding on the AURAC summary results reported in Section~\ref{sec:results_benchmark}, our findings establish that the proposed SRE with UQ not only surpasses prior methods in robustness but also enables fine-grained and safe filtering of LLM outputs—demonstrating both higher accuracy and greater deployability than existing SOTA uncertainty estimators.

\subsubsection{RAC Performance: sentence length}

\begin{figure}[htpb]
  \centering
  \includegraphics[width = 0.90\columnwidth, keepaspectratio]{%
    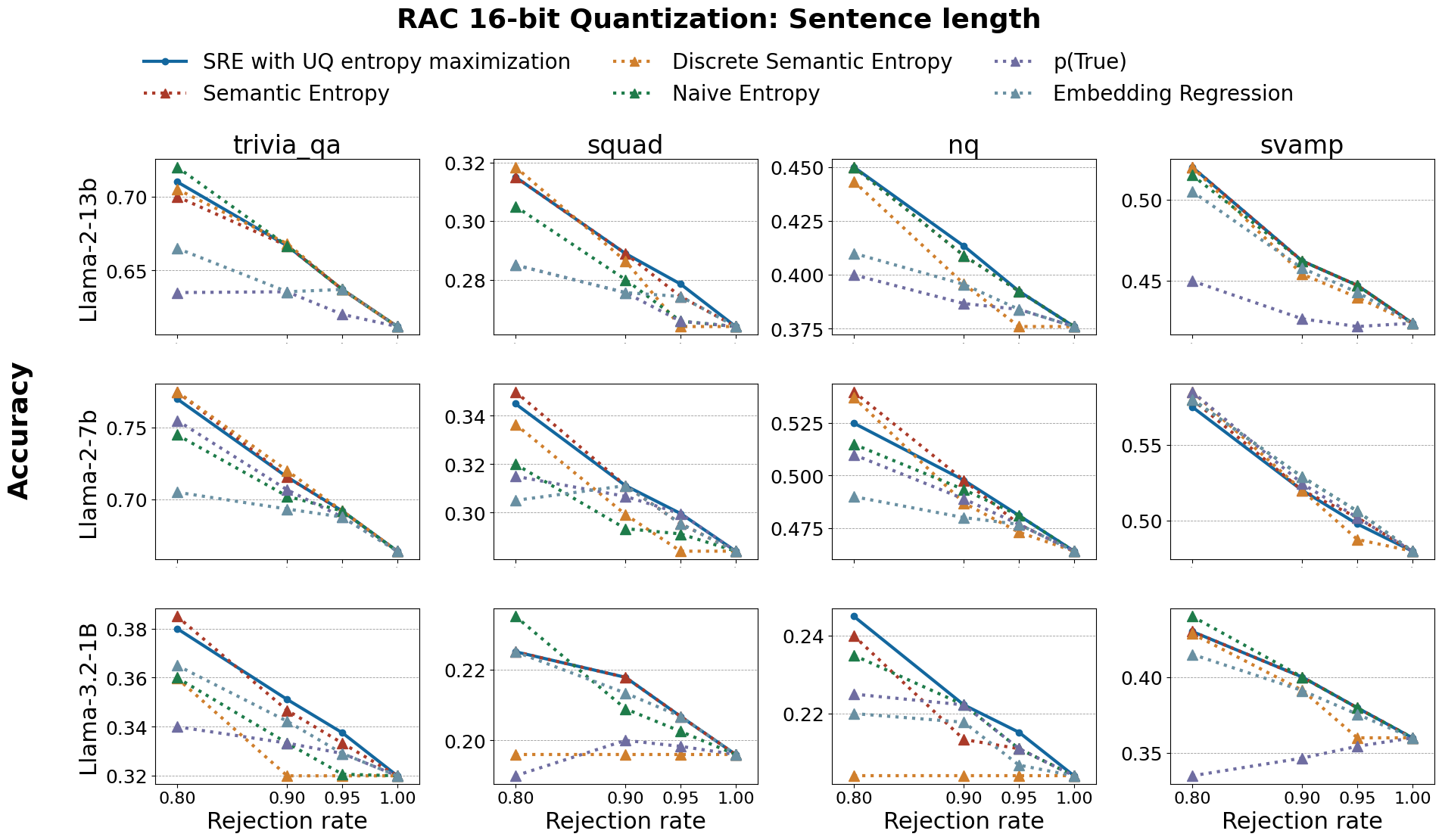}
  \caption{%
  \tb{summarizing 12 experimental} scenarios, RAC scores for confabulation detection across three LLMs (LLaMA 3.2 1B, LLaMA 2 7B, LLaMA 2 13B) at 16 bit precision and four datasets (TriviaQA, SQuAD, NQ, SVAMP). The performance of the proposed SRE with UQ is in par or even higher than SOTA methods.}
  \label{fig:rac_16bit_sentence}
\end{figure}

\begin{figure}[htpb]
  \centering
  \includegraphics[width = 0.90\columnwidth, keepaspectratio]{%
    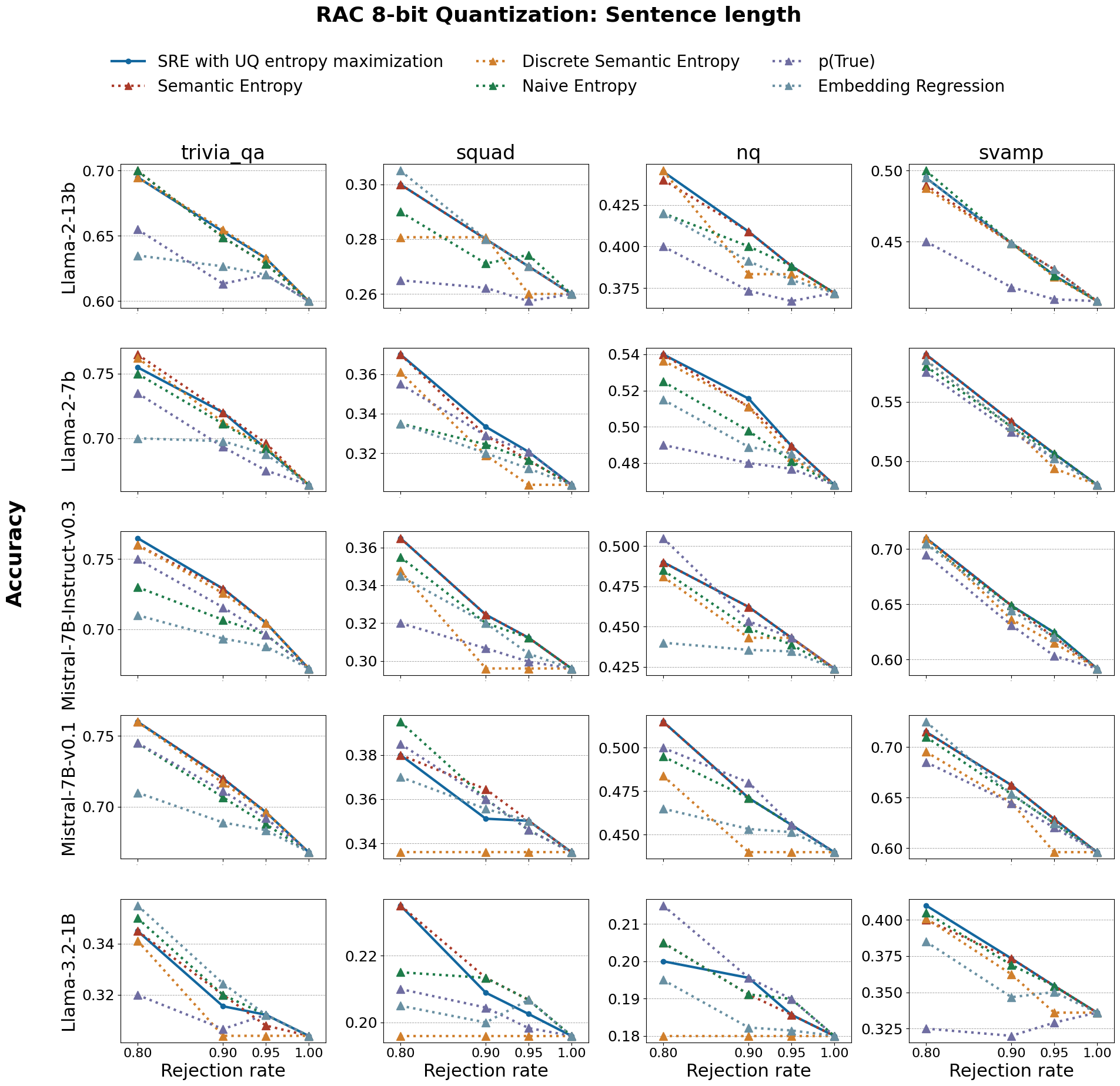}
  \caption{%
  \tb{summarizing 20 experimental} scenarios, RAC scores for confabulation detection across five LLMs (LLaMA 2 13B, LLaMA 2 7B, LLaMA 3.2 1B, Mistral-7B, Mistral-7B-chat) at 8 bit precision and four datasets (TriviaQA, SQuAD, NQ, SVAMP). The performance of the proposed SRE with UQ is in par or even higher than SOTA methods.}
  \label{fig:rac_8bit_sentence}
\end{figure}

\begin{figure}[htpb]
  \centering
  \includegraphics[width = 0.90\columnwidth, keepaspectratio]{%
    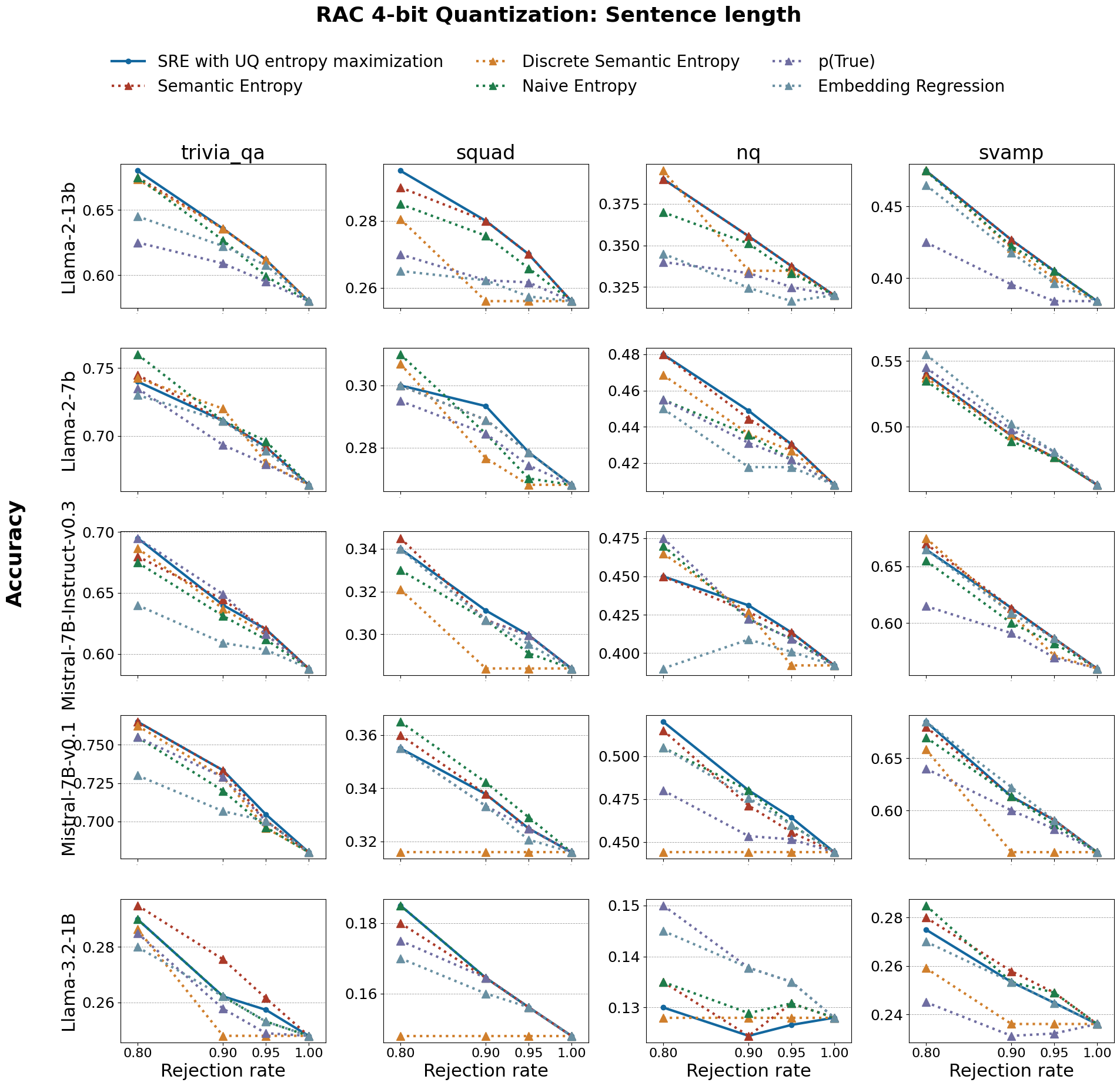}
  \caption{%
  \tb{summarizing 20 experimental} scenarios, RAC scores for confabulation detection across five LLMs (LLaMA 2 13B, LLaMA 2 7B, LLaMA 3.2 1B, Mistral-7B, Mistral-7B-chat) at 8 bit precision and four datasets (TriviaQA, SQuAD, NQ, SVAMP). The performance of the proposed SRE with UQ is in par or even higher than SOTA methods.}
  \label{fig:rac_4bit_sentence}
\end{figure}

Figs.~\ref{fig:rac_16bit_sentence}, \ref{fig:rac_8bit_sentence}, and \ref{fig:rac_4bit_sentence} present RAC curves at 16-bit, 8-bit, and 4-bit precision, respectively, for six hallucination quantification methods across multiple LLMs and four datasets under sentence-length generation. In total, we conducted \tb{52 experiments} spanning all quantization levels, providing a comprehensive basis for sentence length analysis, drawing parallel with its AUROC counterpart evaluation (Appendix \ref{sec:AUROC_sentence}). Several notable patterns emerge:

\vspace{-2mm}
\begin{enumerate}\setlength{\itemsep}{0pt}\setlength{\parskip}{0pt}
\item The proposed SRE with UQ consistently yields the steepest accuracy gains as rejection increases, demonstrating superior ability to prioritize reliable outputs and suppress confabulations. Other semantic-based criteria (SE and Discrete SE) follow closely but remain less effective than SRE with UQ, while probability-based ($p(\text{True})$) and embedding regression baselines lag significantly behind.
\item The advantage of SRE with UQ is particularly pronounced in challenging datasets such as NQ and SQuAD, where its RAC curves show clear separation from both semantic and non-semantic baselines, highlighting robustness under high variability.
\item Moving from 16-bit to 4-bit precision induces only minor changes in the RAC slopes of SRE with UQ, whereas competing methods (e.g., SE and NE) degrade more noticeably. This underscores the resilience of SRE with UQ to quantization.
\item Model size plays a role, with smaller models (e.g., LLaMA-3.2-1B) exhibiting flatter RAC curves than larger counterparts (e.g., LLaMA-2-13B). Nonetheless, SRE with UQ consistently preserves its relative advantage across all scales, confirming its effectiveness even under reduced representational capacity.
\item We also observe a minor performance drop for the NQ dataset when using the Llama 3.1B model, consistent with the results in \ref{sec:AUROC_sentence}, which warrants further investigation in future work.
\end{enumerate}

\subsubsection{RAC Performance: Phrase length on instruct models}

\begin{figure}[htpb]
  \centering
  \includegraphics[width = 0.90\columnwidth, keepaspectratio]{%
    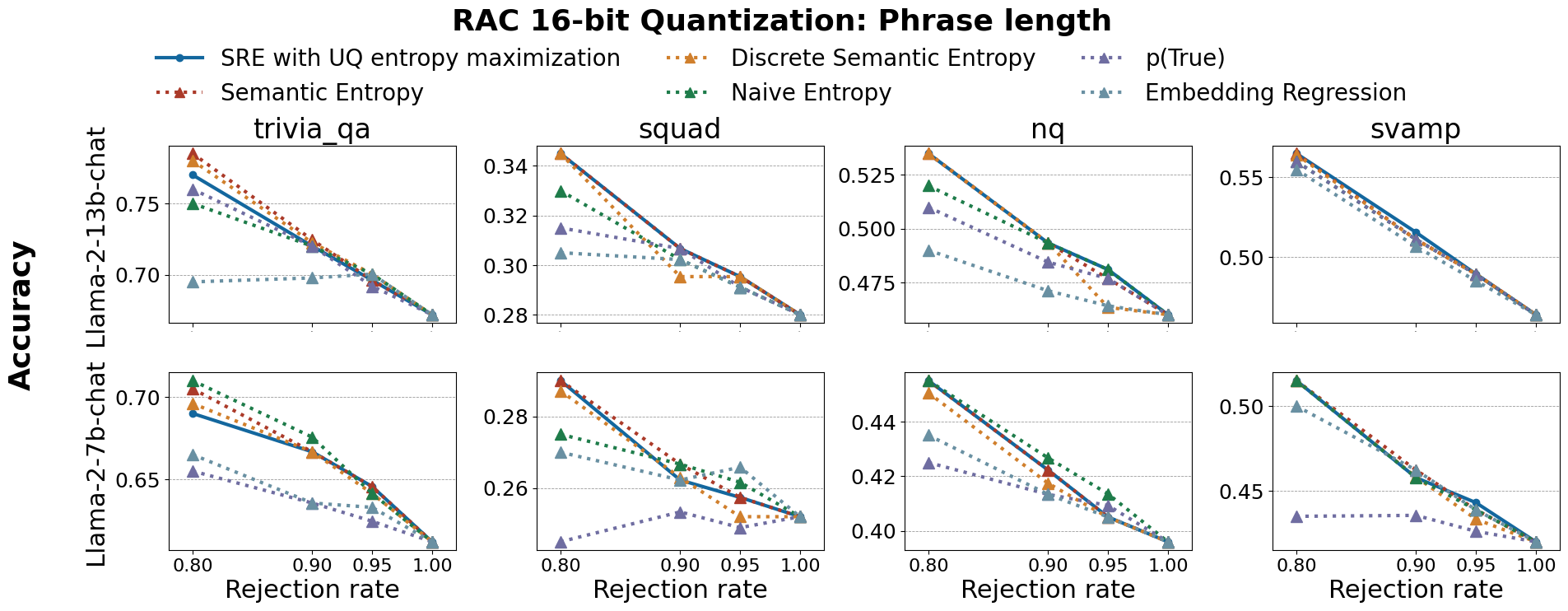}
  \caption{%
  \tb{summarizing 8 experimental} scenarios, RAC scores for confabulation detection across two LLMs (LLaMA 2 13B-chat, LLaMA 2 7B-chat) at 16 bit precision and four datasets (TriviaQA, SQuAD, NQ, SVAMP). The performance of the proposed SRE UQ is in par or even higher than SOTA methods.}
  \label{fig:rac_16bit_chat}
\end{figure}

\begin{figure}[htpb]
  \centering
  \includegraphics[width = 0.90\columnwidth, keepaspectratio]{%
    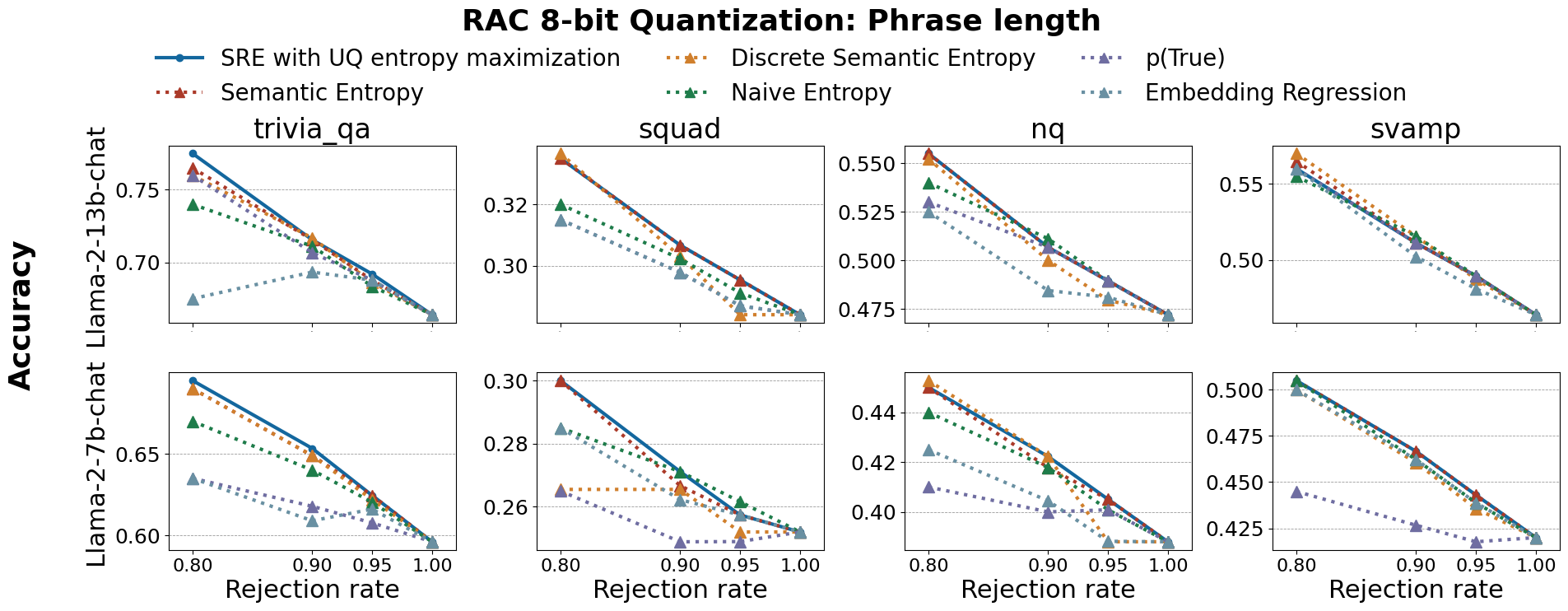}
  \caption{%
  \tb{summarizing 8 experimental} scenarios, RAC scores for confabulation detection across two LLMs (LLaMA 2 13B-chat, LLaMA 2 7B-chat) at 8 bit precision and four datasets (TriviaQA, SQuAD, NQ, SVAMP). The performance of the proposed SRE UQ is in par or even higher than SOTA methods.}
  \label{fig:rac_8bit_chat}
\end{figure}

\begin{figure}[htpb]
  \centering
  \includegraphics[width = 0.90\columnwidth, keepaspectratio]{%
    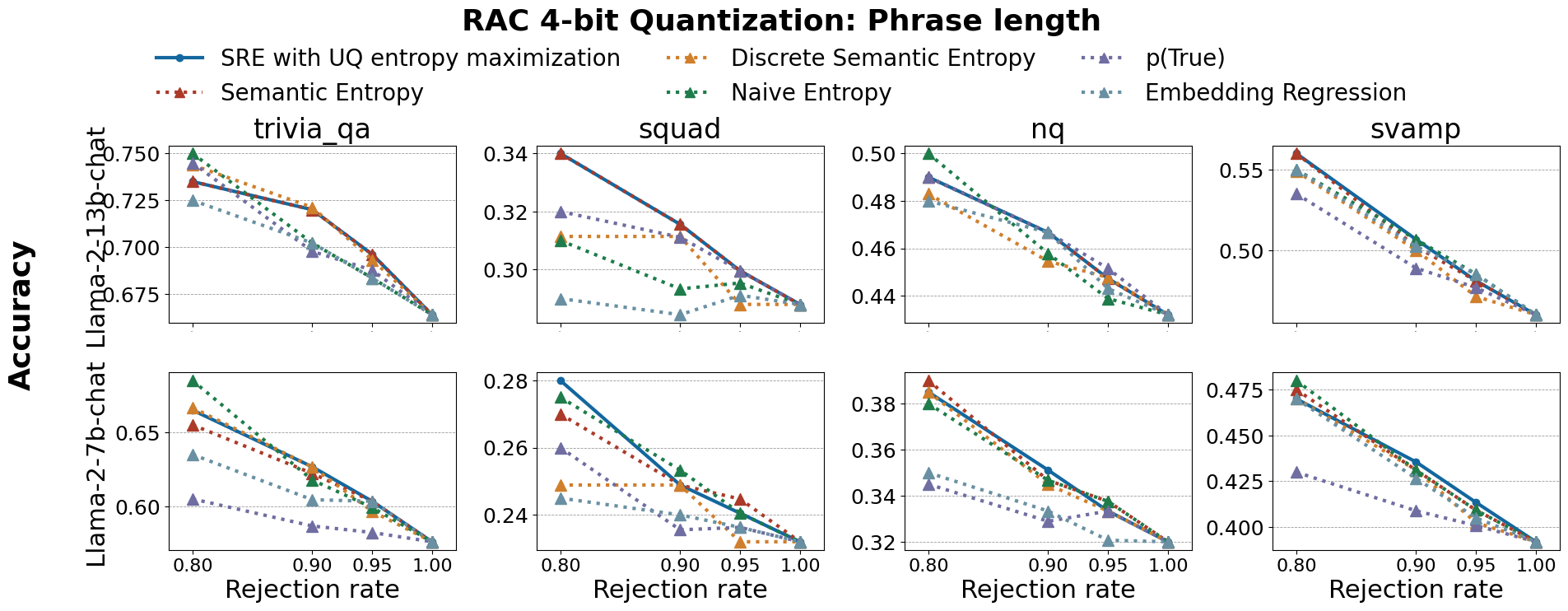}
  \caption{%
  \tb{summarizing 8 experimental} scenarios, RAC scores for confabulation detection across two LLMs (LLaMA 2 13B-chat, LLaMA 2 7B-chat) at 4 bit precision and four datasets (TriviaQA, SQuAD, NQ, SVAMP). The performance of the proposed SRE UQ is in par or even higher than SOTA methods.}
  \label{fig:rac_4bit_chat}
\end{figure}

Figs.~\ref{fig:rac_16bit_chat}, \ref{fig:rac_8bit_chat}, and \ref{fig:rac_4bit_chat} present RAC curves at 16-bit, 8-bit, and 4-bit precision, respectively, for six hallucination quantification methods across multiple LLMs and four datasets under phrase-length generation on instruct fine tuned models. 
Phrase-level completions are particularly relevant for instruction-following use cases, where responses tend to be short and context-sensitive. While prior SOTA hallucination detection methods have largely overlooked this setting, understanding rejection–accuracy behavior under compressed instruction-tuned models is crucial for safe deployment in real-world applications. 

This analysis reveals how quantization interacts with model adaptation to instruction tuning, and whether semantic-based uncertainty measures such as SRE with UQ retain their reliability when generations are concise and more sensitive to uncertainty calibration. In total, we conducted \tb{8 experiments} spanning all quantization levels, providing a comprehensive basis for phrase length analysis, drawing parallel with its AUROC counterpart evaluation (Appendix \ref{sec:AUROC_sentence}). Several notable patterns emerge  upon analysis:

\vspace{-2mm}
\begin{enumerate}\setlength{\itemsep}{0pt}\setlength{\parskip}{0pt}
\item The proposed SRE with UQ consistently yields the steepest RAC slopes across datasets, demonstrating stronger rejection behavior than both probability-based ($p(\text{True})$) and embedding regression baselines.
\item Instruct models reveal that semantic criteria—especially SRE with UQ—retain substantially higher accuracy at rejection rates above 90\%, with the advantage most pronounced on challenging datasets such as TriviaQA and NQ.
\item Across quantization levels, SRE with UQ remains stable from 16-bit to 4-bit precision, whereas baselines degrade more noticeably, underscoring its resilience to model compression.
\item Larger instruct model (e.g., LLaMA-2-13B-chat) amplify the gains of SRE with UQ, while smaller model (e.g., LLaMA-2-7B-chat) show flatter RAC curves; nonetheless, SRE with UQ consistently outperforms alternatives across scales.
\end{enumerate}

\subsubsection{RAC Performance: Phrase length on non instruct models}

\begin{figure}[htpb]
  \centering
  \includegraphics[width = 0.90\columnwidth, keepaspectratio]{%
    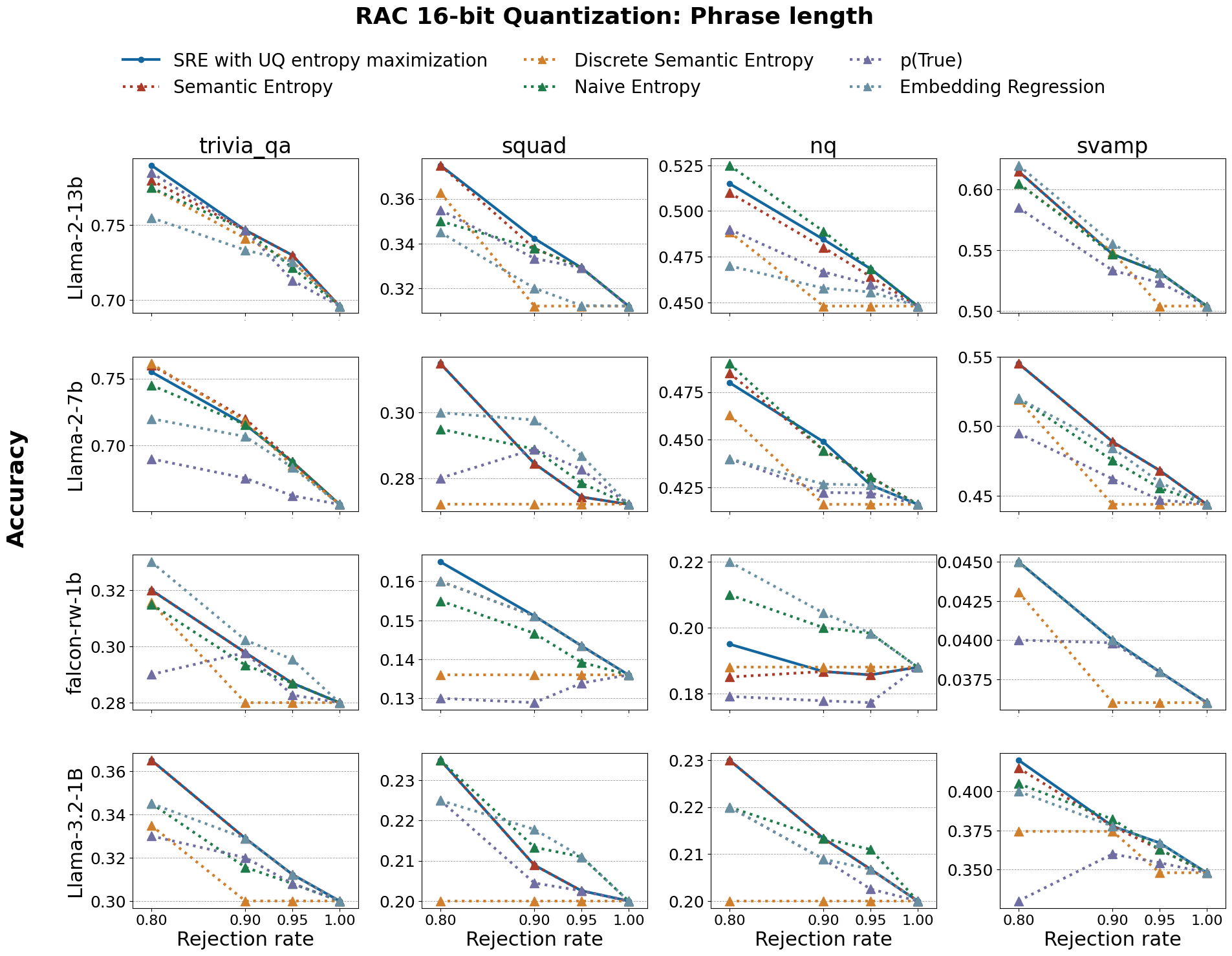}
  \caption{%
  RAC scores for confabulation detection across four LLMs (Mistral-7B, Falcon-1B, LLaMA 3.2B, LLaMA 2 7B 4-bit) and four datasets (TriviaQA, SQuAD, NQ, SVAMP). The performance of the proposed semantic R{\'e}nyi entropy is in par or even higher than SOTA methods.}
  \label{fig:rac_16bit_non_instruct}
\end{figure}

\begin{figure}[htpb]
  \centering
  \includegraphics[width = 0.90\columnwidth, keepaspectratio]{%
    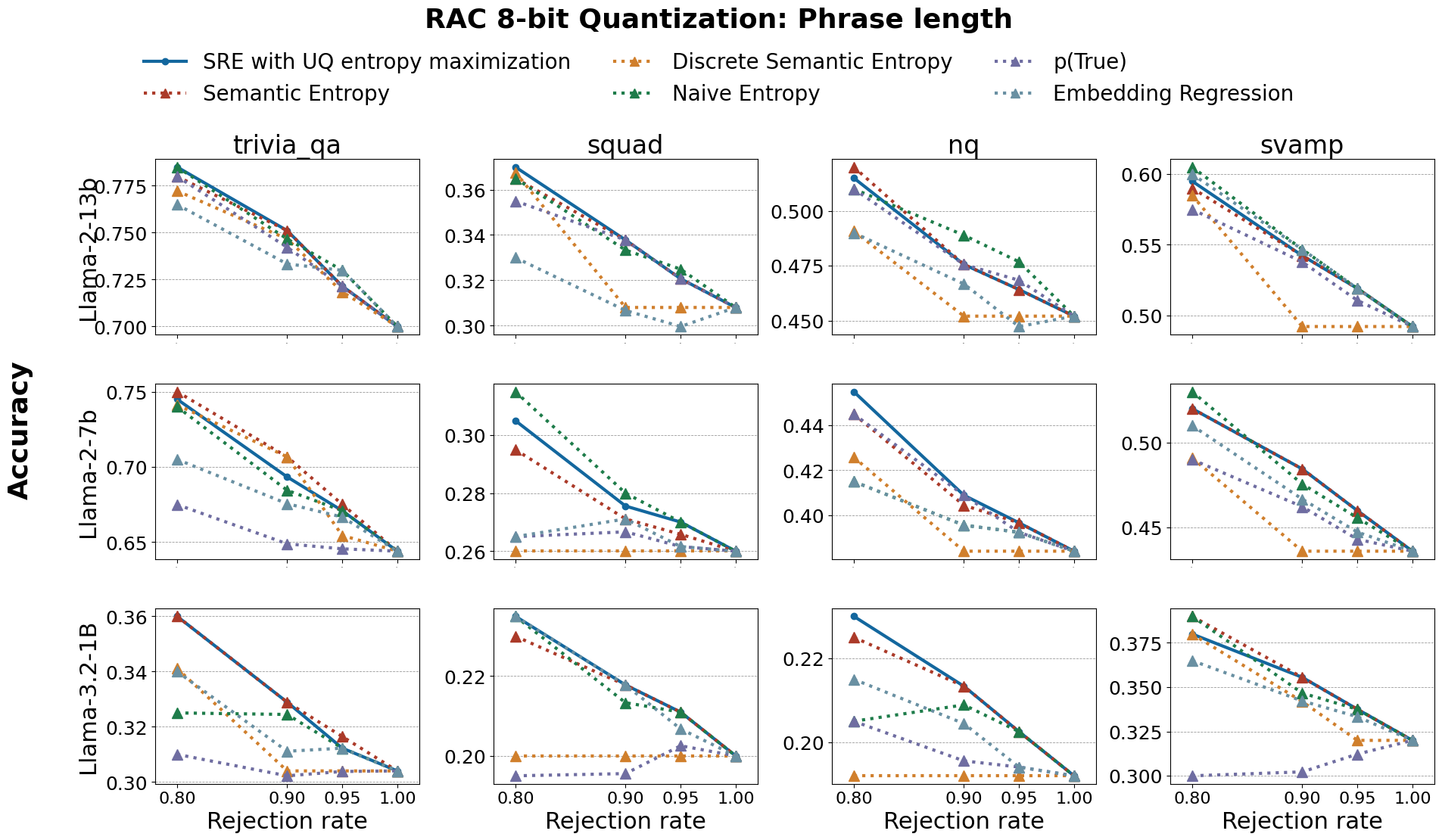}
  \caption{%
  RAC scores for confabulation detection across four LLMs (Mistral-7B, Falcon-1B, LLaMA 3.2B, LLaMA 2 7B 4-bit) and four datasets (TriviaQA, SQuAD, NQ, SVAMP). The performance of the proposed semantic R{\'e}nyi entropy is in par or even higher than SOTA methods.}
  \label{fig:rac_8bit_non_instruct}
\end{figure}

\begin{figure}[htpb]
  \centering
  \includegraphics[width = 0.90\columnwidth, keepaspectratio]{%
    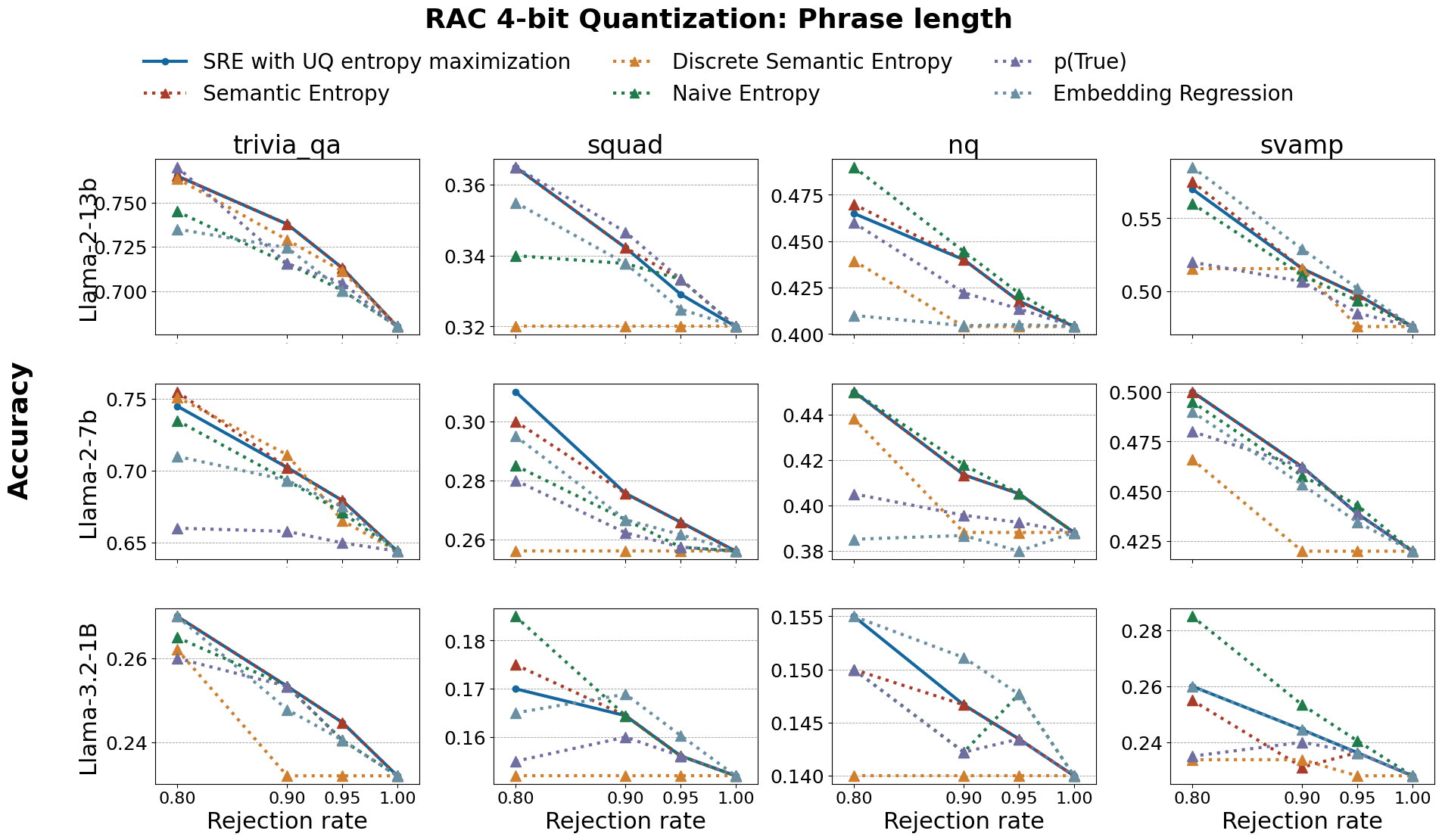}
  \caption{%
  RAC scores for confabulation detection across four LLMs (Mistral-7B, Falcon-1B, LLaMA 3.2B, LLaMA 2 7B 4-bit) and four datasets (TriviaQA, SQuAD, NQ, SVAMP). The performance of the proposed semantic R{\'e}nyi entropy is in par or even higher than SOTA methods.}
  \label{fig:rac_4bit_non_instruct}
\end{figure}

Phrase-level RAC plots for non-instruct LLMs (Figs.~\ref{fig:rac_16bit_non_instruct}, \ref{fig:rac_8bit_non_instruct}, and \ref{fig:rac_4bit_non_instruct}) provide further evidence that semantic-based hallucination detection, and in particular our proposed SRE with UQ, maintains strong robustness under quantization and across diverse datasets. Unlike token-probability and embedding-based baselines, which exhibit flatter or unstable RAC slopes, SRE-UQ consistently yields steeper rejection–accuracy improvements, highlighting its ability to reliably prioritize correct outputs even when compression reduces representational capacity. These findings extend the AUROC trends to phrase-level rejection settings, demonstrating that SRE-UQ is not only effective in distinguishing confabulations but also excels in dynamically filtering them under realistic deployment constraints. Importantly, the proposed SRE with UQ demonstrates consistently stronger rejection–accuracy behavior, highlighting its suitability for lightweight non-instruct models where calibration is more challenging. Several notable patterns emerge upon analysis:

\vspace{-2mm}
\begin{enumerate}\setlength{\itemsep}{0pt}\setlength{\parskip}{0pt}
\item SRE-UQ leads across precisions, across 16-bit, 8-bit, and 4-bit quantization, SRE-UQ produces the steepest RAC curves, confirming that it most effectively raises accuracy as rejection increases. This advantage is especially evident in TriviaQA and NQ, where probability-based baselines stagnate or even flatten.
\item While baselines such as Discrete Semantic Entropy and Naïve Entropy show visible degradation under 4-bit quantization, SRE-UQ’s RAC slopes remain sharp and consistent, underscoring resilience to aggressive model compression.
\item Gains are most pronounced in datasets with higher linguistic variability (e.g., NQ and SVAMP), where SRE-UQ sharply separates from weaker baselines. By contrast, probability-based $p(\text{True})$ often collapses under phrase-level rejection, failing to provide meaningful filtering.
\item Smaller models (e.g., LLaMA-3.2-1B) exhibit generally flatter curves due to limited representational power, but even in these challenging cases, SRE-UQ maintains relative gains, outperforming all other methods consistently.
\end{enumerate}

\subsection{Hyperparameter Selection}
\label{sec:lambda_select_test}

In this section, we examine the effect of hyperparameter tuning for $\lambda$ in the entropy maximization framework introduced in \eqref{eq:entropy_max}. Recall that this formulation adjusts the output TS probabilities by jointly maximizing semantic entropy and penalizing deviations from the original model distribution, weighted by the model’s uncertainty estimates. A critical innovation of our approach lies in integrating UQ directly into the token adjustment mechanism, thereby enhancing the model’s sensitivity to confabulations beyond conventional semantic entropy analysis.

To assess the impact of $\lambda$, we measure AUROC scores for hallucination detection across different values of the hyperparameter. In each case, the R{\'e}nyi’s semantic entropy is computed after probability adjustment, and the results are compared against the baseline R{\'e}nyi’s entropy without uncertainty correction (denoted by the \textcolor{red}{red} line). The $\lambda$ yielding the highest AUROC is in \textcolor{ForestGreen}{green}. This comparison isolates the contribution of UQ-aware entropy maximization to the confabulation detection task.

\begin{figure}[htbp]
  \centering
  \includegraphics[width = 0.95\columnwidth]{%
    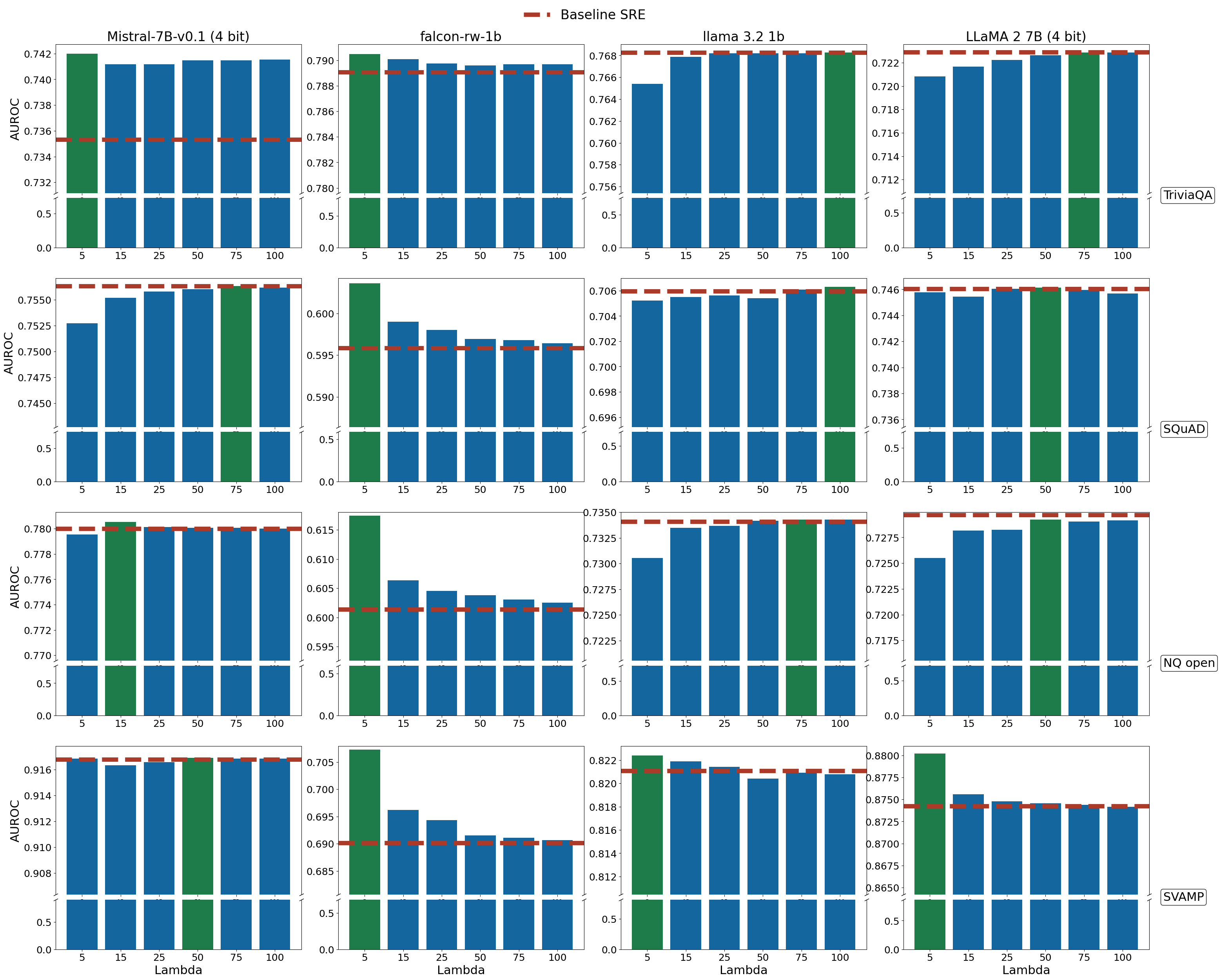}
  \caption{Effect of $\lambda$ hyperparameter selection on AUROC performance. The \textcolor{red}{red} line denotes the baseline SRE without UQ adjustment. Integrating uncertainty via \eqref{eq:entropy_max} consistently enhances hallucination detection capability across all LLM models and datasets. The $\lambda$ yielding the highest AUROC is in \textcolor{ForestGreen}{green}.}
  \label{fig:all_dataset_lambda_search}
\end{figure}

As shown in Fig.~\ref{fig:all_dataset_lambda_search}, incorporating UQ-based probability adjustments yields substantial improvements over the baseline entropy approach across all models tested. 

Several key observations are notable:
\begin{enumerate}
\item Moderate values of $\lambda$ systematically boost AUROC scores, confirming that a calibrated trade-off between entropy maximization and fidelity to the original model predictions is beneficial. 
\item While the precise optimal $\lambda$ may vary slightly across models and datasets, the performance gains are robust across a wide range of $\lambda$ values, underscoring the general effectiveness of the proposed correction. 
\item This experiment validates one of the most crucial contributions of our work: by introducing UQ into semantic entropy calculations, we can significantly sharpen the model’s ability to discriminate confabulations, especially in the presence of noisy or low-confidence predictions.
\end{enumerate}

\section{Additional Details on Algorithmic Structures and Computational Cost}
\label{sec:cost_comp_and_pseudocode}

\subsection{Computational Overhead}
\label{app:cost}

The computational overhead introduced by the QTN-based UQ module is modest and remains well within real-time inference constraints. The method requires storing only $M = 8$ q-bit perturbation features in a $2^8 \times 2^8$ matrix, which is significantly smaller than the embedding caches required by clustering-based semantic entropy methods. Constructing the Hamiltonian associated with the sampled KME incurs a \emph{one-time} pre-processing cost of approximately $45$--$60$\,s on CPU, though this can be substantially reduced through GPU-accelerated linear algebra routines. At inference time, the per-query overhead is minimal: evaluating the perturbation-based uncertainty features adds merely $6$--$10$\,ms on an NVIDIA A6000 GPU, which is negligible compared to the intrinsic latency of LLM decoding. Overall, the QTN perturbation step introduces only a lightweight computational cost while enabling richer uncertainty estimates.

\subsection{Algorithmic Structure and Complexity}
\label{app:alg}

In this subsection, we present pseudocode illustrating key components of our approach, complementing the pipeline diagram provided in Fig.~\ref{fig:semantic_pipeline}. 

This includes pseudocode for (i) Hamiltonian construction from the sampled KME, (ii) extraction of first-order perturbation features, and (iii) the entropy-maximization procedure from Sec.~2.3.
\begin{algorithm}[H]
\caption{Construction of Empirical KME $\widehat{\psi}(x)$}
\label{alg:kme}
\begin{algorithmic}[1]
\STATE \textbf{Input:} Samples $\{x_i\}_{i=1}^N$ from random variable $X$; kernel bandwidth $\sigma$; sampling grid $\{x_j\}_{j=0}^{M-1}$
\STATE \textbf{Output:} Sampled KME vector $\widehat{\boldsymbol{\psi}} = \{\widehat{\psi}_j\}_{j=0}^{M-1}$

\FOR{$j = 0$ to $M-1$}
    \STATE Compute Parzen PDF estimate:
    \[
      \widehat{p}(x_j) = \frac{1}{N} \sum_{i=1}^N \kappa_\sigma(x_j; x_i)
    \]
    where $\kappa_\sigma(x_j;x_i)$ is the Gaussian kernel
    \STATE Set KME value $\widehat{\psi}_j \gets \widehat{p}(x_j)$
\ENDFOR

\STATE \textbf{return} $\widehat{\boldsymbol{\psi}}$
\end{algorithmic}
\end{algorithm}

\begin{algorithm}[H]
\caption{QTN Hamiltonian Construction via Quantum Correlation Matrix}
\label{alg:hamiltonian}
\begin{algorithmic}[1]
\STATE \textbf{Input:} Sampled KME vector $\widehat{\boldsymbol{\psi}}$ of length $M=2^L$; operator basis $\widehat{\mathcal{H}} = \{\widehat{\boldsymbol{H}}_k\}_{k=0}^{T-1}$
\STATE \textbf{Output:} Local Hamiltonian $\widehat{\boldsymbol{H}}$

\STATE Construct the QCM matrix using the inner products, where the $(k,\ell)$ entry is given by :
\[
  \mathbf{M}_{\widehat{\boldsymbol{\psi}}}(k,\ell)
  = \langle \widehat{\boldsymbol{H}}_k\widehat{\boldsymbol{\psi}}, \widehat{\boldsymbol{H}}_\ell \,\widehat{\boldsymbol{\psi}} \rangle
\]

\STATE Compute null space of QCM:
\[
 \mathbf{w} = \{w_k\}_{k=0}^{T-1} \in \mathrm{Null}(\mathbf{M}_{\widehat{\boldsymbol{\psi}}})
\]

\STATE Form Hamiltonian as linear combination:
\[
 \widehat{\boldsymbol{H}} = \sum_{k=0}^{T-1} w_k \widehat{\boldsymbol{H}}_k
\]

\STATE \textbf{return} $\widehat{\boldsymbol{H}}$
\end{algorithmic}
\end{algorithm}


\begin{algorithm}[H]
\caption{First-Order Perturbation Corrections and TS UQ}
\label{alg:perturbation}
\begin{algorithmic}[1]
\STATE \textbf{Input:} Base Hamiltonian 
\[
   \widehat{\boldsymbol{H}} = \!\sum_{k=0}^{T-1} w_k\, \widehat{\boldsymbol{H}}_k,
\]
where $\{w_k\}$ are the learned weights multiplying the Hermitian operator basis; perturbation vector $\Delta \mathbf{w}$ specifying perturbation of these weights.
\STATE \textbf{Output:} First-order corrected eigen-modes $\{\psi_m^{(1)}\}$ and the resulting UQ feature vector.

\STATE Construct the perturbed Hamiltonian:
\[
   \Delta \boldsymbol{H} = \sum_{k=0}^{T-1} \Delta w_k\, \widehat{\boldsymbol{H}}_k.
\]
This corresponds to perturbing the coefficients (weights) associated with the Pauli-like operators forming the QTN Hamiltonian.

\STATE Compute eigen-decomposition of the unperturbed Hamiltonian:
\[
  \widehat{\boldsymbol{H}}\, \psi_m = E_m\, \psi_m,
  \qquad m = 0,\dots,M-1.
\]

\FOR{$m = 0$ to $M-1$}
   \STATE Compute first-order energy correction:
   \[
     E_m^{(1)} 
       = \langle \psi_m,\; \Delta \boldsymbol{H}\, \psi_m \rangle.
   \]

   \STATE Compute first-order eigenvector correction:
   \[
     \psi_m^{(1)}
       = \sum_{n \neq m}
         \frac{
            \langle \psi_n,\; \Delta \boldsymbol{H}\, \psi_m \rangle
         }{E_m - E_n}\;
         \psi_n.
   \]
\ENDFOR

\STATE We utilize Eq.~\ref{eq:UQ_equation} and $\{\psi_m^{(1)}\}$  to compute UQ feature vectors in the KME amplitude domain, yielding the local UQ scores that appear in Eq.~\ref{eq:UQ_equation2}.

\STATE \textbf{return} $\{\psi_m^{(1)}\}$ \& ${\mathrm{UQ}(p_s^{(r)})}$.
\end{algorithmic}
\end{algorithm}


\begin{algorithm}[H]
\caption{Calibrated Entropy-Maximization Update for TS Probabilities}
\label{alg:entropy_max}
\begin{algorithmic}[1]

\STATE \textbf{Input:} 
\begin{itemize}[noitemsep,topsep=0pt,leftmargin=2em]
    \item Original TS probability $p_s^{(r)}$
    \item Uncertainty score $\mathrm{UQ}(p_s^{(r)})$ computed via Eq.~\ref{eq:UQ_equation2}
    \item Regularization parameter $\lambda$
\end{itemize}

\STATE \textbf{Output:} Adjusted probability $p_s^{(r)^*}$

\vspace{0.5em}
\STATE Define objective:
\[
  \mathcal{L}(\widehat{p}) 
     = -\log\!\left(\widehat{p}^2 + (1-\widehat{p})^2\right)
       - \frac{\lambda}{\mathrm{UQ}(p_s^{(r)})} 
         \cdot \mathrm{KL}\!\left(\widehat{p} \,\Vert\, p_s^{(r)}\right)
\]
where $\mathrm{KL}(\widehat{p}\Vert p)=\widehat{p}\log\!\frac{\widehat{p}}{p} + (1-\widehat{p})\log\!\frac{1-\widehat{p}}{1-p}$.

\STATE Initialize: $\widehat{p}^{(0)} \gets p_s^{(r)}$

\vspace{0.5em}
\FOR{$t = 1$ to $T_{\max}$}
   \STATE Compute gradient $\nabla \mathcal{L}(\widehat{p}^{(t-1)})$
   \STATE Update via projected gradient ascent
   \STATE Stop early if $|\widehat{p}^{(t)} - \widehat{p}^{(t-1)}| < \delta$
\ENDFOR

\vspace{0.5em}
\STATE Set $p_s^{(r)^*} \gets \widehat{p}^{(T)}$

\STATE \textbf{return} $p_s^{(r)^*}$

\end{algorithmic}
\end{algorithm}

\end{document}